\documentclass{article}

\usepackage{microtype}
\usepackage{graphicx}
\usepackage{enumitem}
\usepackage{subcaption}
\usepackage{lineno, blindtext}
\usepackage{booktabs} % for professional tables
\usepackage[export]{adjustbox}
\usepackage{placeins}
\usepackage{pifont}
\usepackage{tcolorbox}
\usepackage{enumitem}
\usepackage{comment}
\usepackage{xcolor}

\setlist{nosep}

\newcommand{\cmark}{\ding{52}}%
\newcommand{\xmark}{\ding{56}}%
\usepackage{hyperref}

\usepackage{algorithmic}

\usepackage[accepted]{icml2023}

\usepackage{amsmath}
\usepackage{amssymb}
\usepackage{mathtools}
\usepackage{amsthm}

\usepackage[capitalize,noabbrev]{cleveref}

\theoremstyle{plain}

\theoremstyle{definition}

\theoremstyle{remark}

\newif\ifshowcomments

\newfloat{algorithmfloat}{htbp}{loa}
\floatname{algorithmfloat}{Algorithm}

\usepackage[textsize=tiny]{todonotes}

\icmltitlerunning{A Three-regime Model of Network Pruning}

\begin{document}

\twocolumn[
\icmltitle{A Three-regime Model of Network Pruning}



\icmlsetsymbol{equal}{*}

\begin{icmlauthorlist}
\icmlauthor{Yefan Zhou}{icsi,ucb}
\icmlauthor{Yaoqing Yang}{dart}
\icmlauthor{Arin Chang}{ucb}
\icmlauthor{Michael W. Mahoney}{icsi,ucb,lbnl}
%\icmlauthor{Michael W. Mahoney}{sch}
%\icmlauthor{Firstname5 Lastname5}{yyy}
%\icmlauthor{Firstname7 Lastname7}{comp}
%\icmlauthor{Firstname8 Lastname8}{sch}
%\icmlauthor{Firstname8 Lastname8}{yyy,comp}
\end{icmlauthorlist}

\icmlaffiliation{icsi}{International Computer Science Institute, CA, USA}
\icmlaffiliation{ucb}{University of California, Berkeley, CA, USA}
\icmlaffiliation{dart}{Dartmouth College, NH, USA}
\icmlaffiliation{lbnl}{Lawrence Berkeley National Laboratory, CA, USA}

\icmlcorrespondingauthor{Yefan Zhou}{yefan0726@berkeley.edu}

\icmlkeywords{Machine Learning, ICML}

\vskip 0.3in
]



\printAffiliationsAndNotice{}  % leave blank if no need to mention equal contribution

\begin{abstract}
Recent work has highlighted the complex influence training hyperparameters, e.g., the number of training epochs, can have on the prunability of machine learning models. 
Perhaps surprisingly, a systematic approach to predict precisely how adjusting a specific hyperparameter will affect prunability remains elusive. 
To address this gap, we introduce a phenomenological model grounded in the statistical mechanics of learning. 
Our approach uses \emph{temperature-like} and \emph{load-like} parameters to model the impact of neural network (NN) training hyperparameters on pruning performance.
A key empirical result we identify is a sharp transition phenomenon: depending on the value of a load-like parameter in the pruned model, increasing the value of a temperature-like parameter in the pre-pruned model may either enhance or impair subsequent pruning performance. 
Based on this transition, we build a three-regime model by taxonomizing the \emph{global structure} of the pruned NN loss landscape. 
Our model reveals that the dichotomous effect of high temperature is associated with transitions between distinct types of global structures in the post-pruned model.
Based on our results, we present three case-studies:
1) determining whether to increase or decrease a hyperparameter for improved pruning; 
2) selecting the best model to prune from a family of models; and 
3) tuning the hyperparameter of the Sharpness Aware Minimization method for better pruning~performance.
\end{abstract}
\vspace{-0.8cm}
\section{Introduction}

\begin{figure*}[t]
\centering

\subfloat[]{
\adjustbox{valign=c}{       
\begin{tabular}{c|c|c|c}
\toprule
     &   \includegraphics[width=0.14\linewidth]{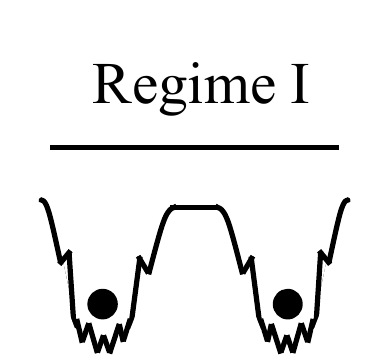}             &   \includegraphics[width=0.14\linewidth]{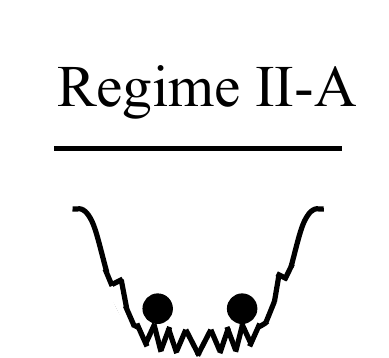}          &   \includegraphics[width=0.14\linewidth]{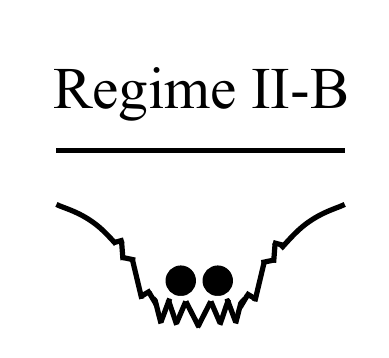}        \\ \hline

\rule{0pt}{30pt} Connectivity  &    \xmark        &     \cmark  &     \cmark  \\ \hline

\rule{0pt}{30pt} Similarity  &     \xmark        &     \xmark  &     \cmark  
\end{tabular}
}
\label{fig:table}
}
\subfloat[]{           
     \includegraphics[width=0.33\linewidth,valign=c]{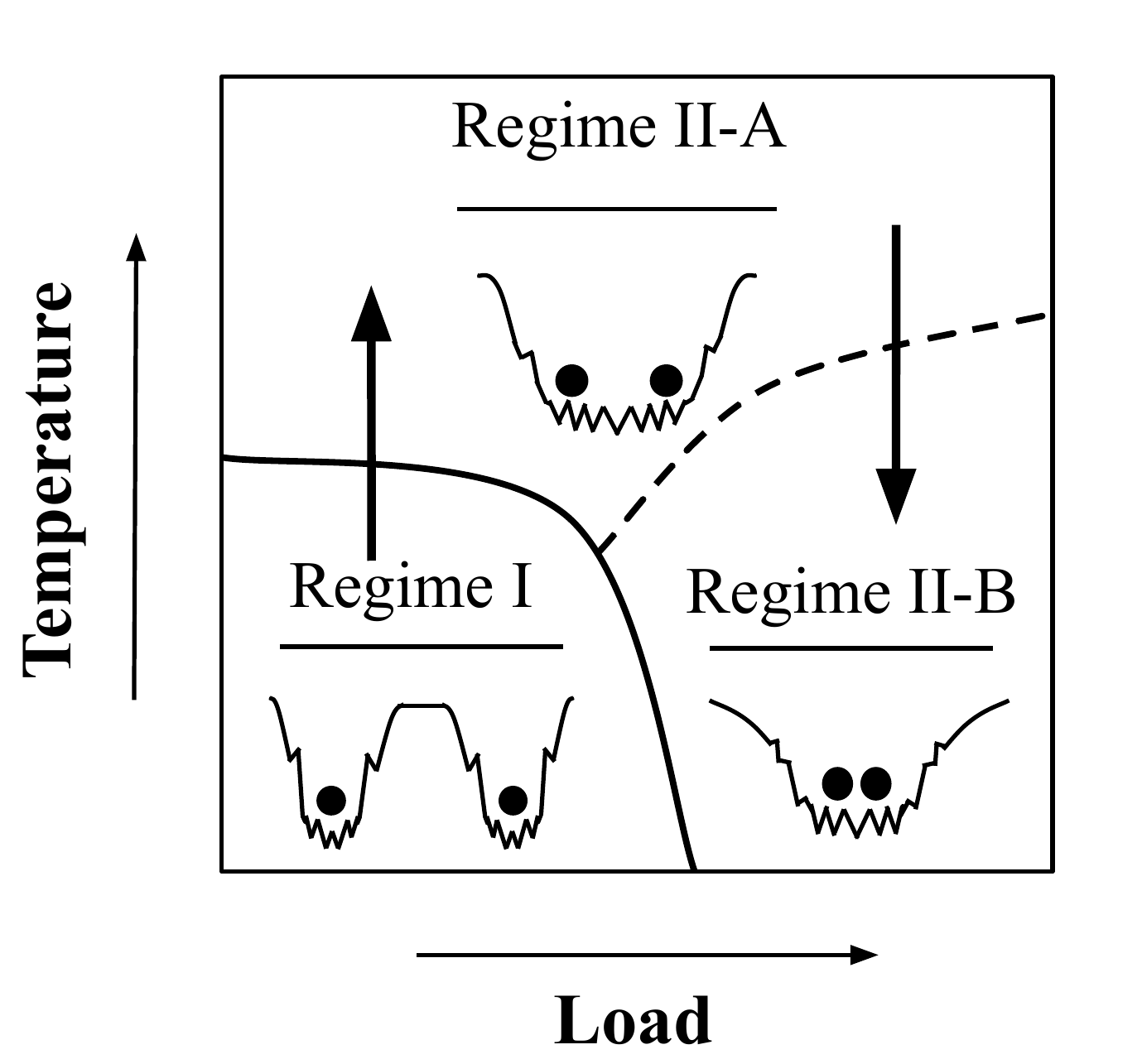}
    \label{fig:three_regime}
}
\vspace{-1mm}
\caption{The three regimes of pruning obtained by varying temperature-like parameters (in the dense pre-pruned model) and load-like parameters (in the sparse post-pruned model): loss landscape connectivity metrics such as LMC identify Regime I versus Regime II; and loss landscape similarity metrics of outputs between models in well-connected regimes then identify Regimes II-A and II-B. 
The regimes are thus Regime I (poorly-connected loss landscapes); Regime II-A (well-connected but relatively dissimilar model outputs); and Regime II-B (well-connected and relatively similar outputs). 
For a given load goal (density of the pruned model), we focus on the favorable transitions from Regime I to Regime II-A (obtained by increasing the temperature) and from Regime II-A to Regime II-B (obtained by decreasing the temperature), as indicated by the arrows.}
\label{fig:caricature}
\end{figure*}

A recently-popular approach to compressing large neural networks (NNs) is to perform pruning, i.e., to remove unnecessary weights from a trained model.
The resulting ``sparser'' NNs often have improved memory and inference efficiencies, compared to the original ``denser'' NNs. 

A common approach to pruning~\citep{obd1989, han2015learning, molchanov2019pruning} involves adopting a three-stage train-prune-retrain pipeline: 
1) train a large or over-parameterized dense model to some sort of convergence; 
2) prune the dense model to obtain a sparse model; and then 
3) retrain the sparse model to recover its performance.
A considerable amount of work has focused on improving the sub-network performance in the (second) pruning stage~\citep{blalock2020state} and the (third) retraining stage~\citep{Renda2020Comparing, le2021network}.
However, there remains little guidance for the (first) stage of dense model training, i.e., how to improve the prunability of the original large model~\cite{rosenfeld2021predictability}.
Recent work has shown that tuning optimization-related hyperparameters, such as the number of training epochs~\citep{li2020train, liu2021sparse, Shen_2022_CVPR} and the batch size~\citep{barsbey2021heavy}, can potentially benefit specific pruning methods. 
Despite these findings, a principled approach to predict when and how adjusting a given hyperparameter during the stage of dense model training will impact subsequent pruning performance remains to be developed.

In this paper, we develop a simple operational model for NN pruning, focusing on the optimal selection of hyperparameters, such as the number of training epochs, in the (first) training stage.
Our model is inspired by recent work in the statistical mechanics of learning~\citep{yang2021taxonomizing,martin2017rethinking,martin2018implicit_JMLRversion}.
In the statistical mechanics approach to learning, multiple qualitatively different ``phases'' of behavior can arise in black-box Deep NNs.
This concept is explicitly illustrated by the Very Simple Deep Learning (VSDL) model, proposed in~\citet {martin2017rethinking}.
Within the VSDL model, these phases and the sharp transitions between them can be identified on a two-dimensional ``phase'' diagram, where the $x$ and $y$ axes represent load-like and temperature-like parameters, respectively. 
Load-like parameters characterize the relationship between the quantity and/or quality of data, relative to model size.
This was represented as label noise in training data by \citet{martin2017rethinking} and as width scaling by \citet{yang2021taxonomizing}.
Temperature-like parameters, on the other hand, characterize the magnitude of noise introduced during stochastic training.
This can be represented in terms of common hyperparameters for training, including the number of epochs and batch size in \citet{martin2017rethinking,yang2021taxonomizing}.
Using the notions of load-like and temperature-like parameters, \citet{yang2021taxonomizing} provided a comprehensive taxonomy of NN loss landscapes, employing easily computed metrics to identify, predict, and distinguish qualitatively different phases of model training.

{\bf Our three-regime VSDL model for pruning.}
We construct a novel model for NN pruning that is based on the VSDL model.
In our model, the load-like parameter is represented as a distinct hyperparameter used in the pruning process, namely the density of the pruned models; and the temperature-like parameter is represented as common hyperparameters for the dense model training (e.g., training epochs, batch size), similar to \citet{martin2017rethinking, yang2021taxonomizing}.

Our empirical results validate the effectiveness of the proposed model, demonstrating that adjusting the load and temperature parameters can lead to relatively-sharp transitions in model performance and that making decisions based on this leads to improved test error for pruned models. 
Moreover, our work confirms the metrics previously used to develop the taxonomy of loss landscapes~\citep{yang2021taxonomizing}, showcasing their applicability to the very different problem of model pruning.
Ultimately, our results contribute to the more efficient, metric-informed selection of temperature-like hyperparameters (depending on the load), offering practical approaches for applying VSDL models.

In more detail, we consider two metrics to measure NN loss landscapes, namely the linear mode connectivity (LMC)~\citep{garipov2018loss, draxler2018essentially,frankle2020linear} and the centered kernel alignment (CKA) similarity~\citep{kornblith2019similarity}.
LMC quantifies how well different local minima are connected to each other in the loss landscape, and thus it captures the \emph{connectivity} between trained models.
CKA similarity, on the other hand, is used to capture the \emph{similarity} between the outputs of models. 
\citet{yang2021taxonomizing} used these two metrics to measure the \emph{global structure} of loss landscapes. 
Here, we apply these insights to the model pruning problem.
As indicated in Figure \ref{fig:caricature}, our results empirically produce a three-regime taxonomy based on the similarity and connectivity of pruned models' loss landscape, effectively diagnosing and explaining the pruning performance as load-like and temperature-like control parameters are varied.

Our main contributions are as follows:
\begin{itemize} 
    \item 
    We construct a VSDL model to study the different types of model training in order to improve pruning performance. 
    Our model is taxonomized into three regimes based on the connectivity and similarity of the pruned model loss landscape. 
    This is graphically represented in Figure~\ref{fig:caricature}.
    \item 
    Our three-regime model effectively identifies and predicts a dichotomous phenomenon: depending on the value of the target load parameter and the measured values of the LMC and CKA, one can obtain improved pruned models either by increasing or decreasing the temperature parameter. 
    Our three-regime model demonstrates that the phenomenon is well correlated with the transition among the three regimes, as shown in Figure~\ref{fig:three_regime}.
    \item 
    Our new insights on three-regime pruning lead to new practical approaches to improving pruning, which we present as three case-studies:
    1) given initial load and temperature hyperparameters, we can predict the correct direction to adjust the temperature in order to achieve optimal pruning performance;
    2) for a given target model load, we can design a new model selection method based on connectivity and similarity that can predict the optimal temperature parameter without the need for an exhaustive grid search; and
    3) we find that training models with the Sharpness Aware Minimization (SAM) method results in improved pruning when the temperature-like parameter of SAM is optimally tuned using our three-regime model.
\end{itemize}

Our code is open-sourced.\footnote{\href{https://github.com/YefanZhou/ThreeRegimePruning}{https://github.com/YefanZhou/ThreeRegimePruning}}
We provide a comprehensive overview of related work as well as provide further discussion of two aspects of our approach that differ from how \citet{yang2021taxonomizing} applied the VSDL model in Appendix~\ref{sxn:related-work}.

\section{Background and Setup} 
\label{def:basic}

In this section, we provide background and a basic setup for our main results. 

\subsection{Load and temperature}
\label{def:load-temp}

Load and temperature parameters enter naturally when considering the statistical mechanics approach to learning and generalization, both historically~\citep{SST92,WRB93,DKST96,EB01_BOOK}, where they enter directly, as well as more recently~\citep{martin2017rethinking,martin2018implicit_JMLRversion}, where they enter indirectly via other control parameters.

Informally, a \emph{load-like parameter}, which we will denote generically by $L$, refers to some quantity related to the quantity/quality of the data, relative to the size of the model. 
Here, we mainly study the models after pruning, and we vary the model density $L_\text{density}$ to change the load. 
In this case, decreasing $L_\text{density}$ corresponds to having more data, relative to the model size.
We also study alternative load parameters, e.g.,  model width scaling $L_\text{width}$ or depth $L_\text{depth}$ of the pruned model, given a fixed model density.

Similarly, a \emph{temperature-like parameter}, which we will denote generically by $T$, refers to some quantity related to the empirical noise/stochasticity introduced in the learning process. 
We mainly adjust the temperature by varying the number $T_\text{epoch}$ of training epochs (in the first stage of the train-prune-retrain pipeline) or the  batch size $T_\text{batch}$. 
The neighborhood size $\rho$, a hyperparameter in SAM (that indicates the magnitude of the adversarial perturbations over the model parameters before each gradient update of SGD optimizer), can also be viewed as a temperature-like parameter (as we show in Section \ref{sec:app3}). 
In practice, an increase in temperature corresponds to fewer training epochs, smaller batch size, and larger neighborhood~size.

\subsection{Preliminaries}  
\label{def:preli}
Consider a NN $f(\mathbf{x}; \Theta): \mathbb{R}^{d_{\text {in }}} \rightarrow \mathbb{R}^{d_{\text {out }}}$ with input $\mathbf{x}$ and parameters $\Theta \in \mathbb{R}^m $. We train the NN using minibatch SGD, with batch size $T_\text{batch}$, and total number of epochs $T_\text{epoch}$. 
For randomly initialized parameters, $\Theta_0$, SGD randomness, $\xi$, the process of training with specific temperature-like parameters is denoted by $\operatorname{Train}\left(\Theta_{0}, \xi, T \right)$, $T = \{T_\text{epoch}, T_\text{batch}, ...\}$. We mainly study NNs in classification tasks, using $\operatorname{error}_{\text {train}}(\Theta)$ and $\operatorname{error}_{\text {test}}(\Theta)$ to represent the error on the training and test sets, respectively.

\subsection{Network pruning}
\label{def:network-prune}

We mainly investigate unstructured pruning, which removes weights without considering model structure.

\textbf{Model density.} 
We define the model density, denoted as $L_\text{density} = \frac{ \left|\mathcal{M}\right|}{m}$, as the fraction of the number of parameters preserved to the total parameter count, where $\mathcal{M} \in \{0, 1\}^N$ is the binary pruning mask informing the position of remaining weights. A pruned model is denoted as $\Theta \odot \mathcal{M} $, where $\odot$ indicates the element-wise product.
The process of pruning a model with specific width/depth into a target density is denoted by $\operatorname{Prune} (\Theta, L)$, $L = \{L_\text{density}, L_\text{width}, L_\text{depth},...\}$.

\textbf{Pruning strategies.}
Magnitude pruning (MP) utilizes the absolute value of each parameter as a measure of importance to determine the mask $\mathcal{M}$. Unstructured MP retains the top-$L_\text{density}$ percent of important parameters, with two variants based on the layer-wise distribution of density: 1) \textsc{UniformMP}, which prunes each layer uniformly, and 2) \textsc{GlobalMP}, which imposes a global threshold on parameter magnitudes of every layer for achieving the desired global density target.

\begin{figure*}[!thb]\centering
  \begin{subfigure}{0.24\linewidth}
       \centering
       \includegraphics[width=\linewidth]{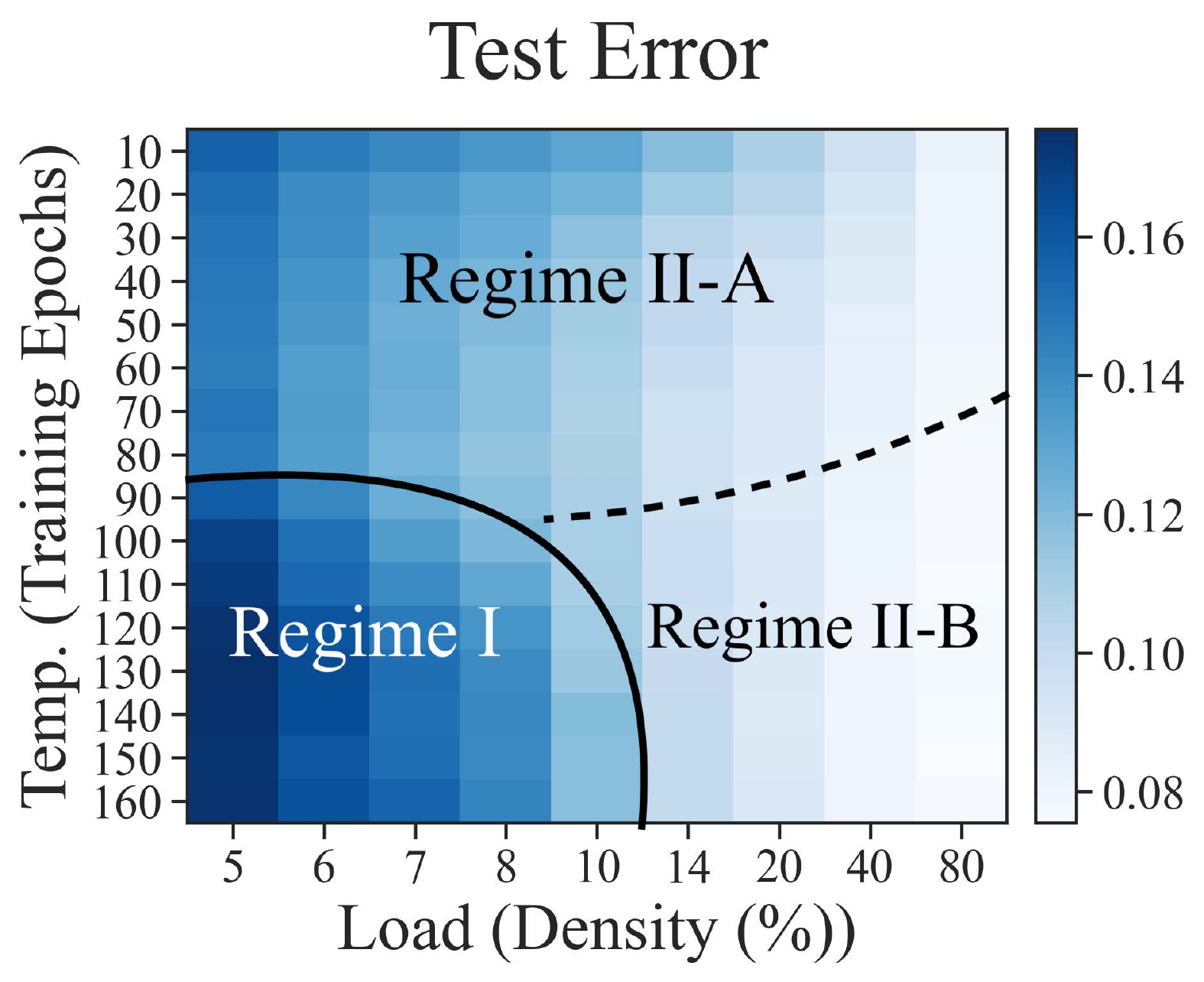}
       \caption{Test error}
       \label{fig:earlystop_testerror}
   \end{subfigure}
    \begin{subfigure}{0.24\linewidth}
       \centering
       \includegraphics[width=\linewidth]{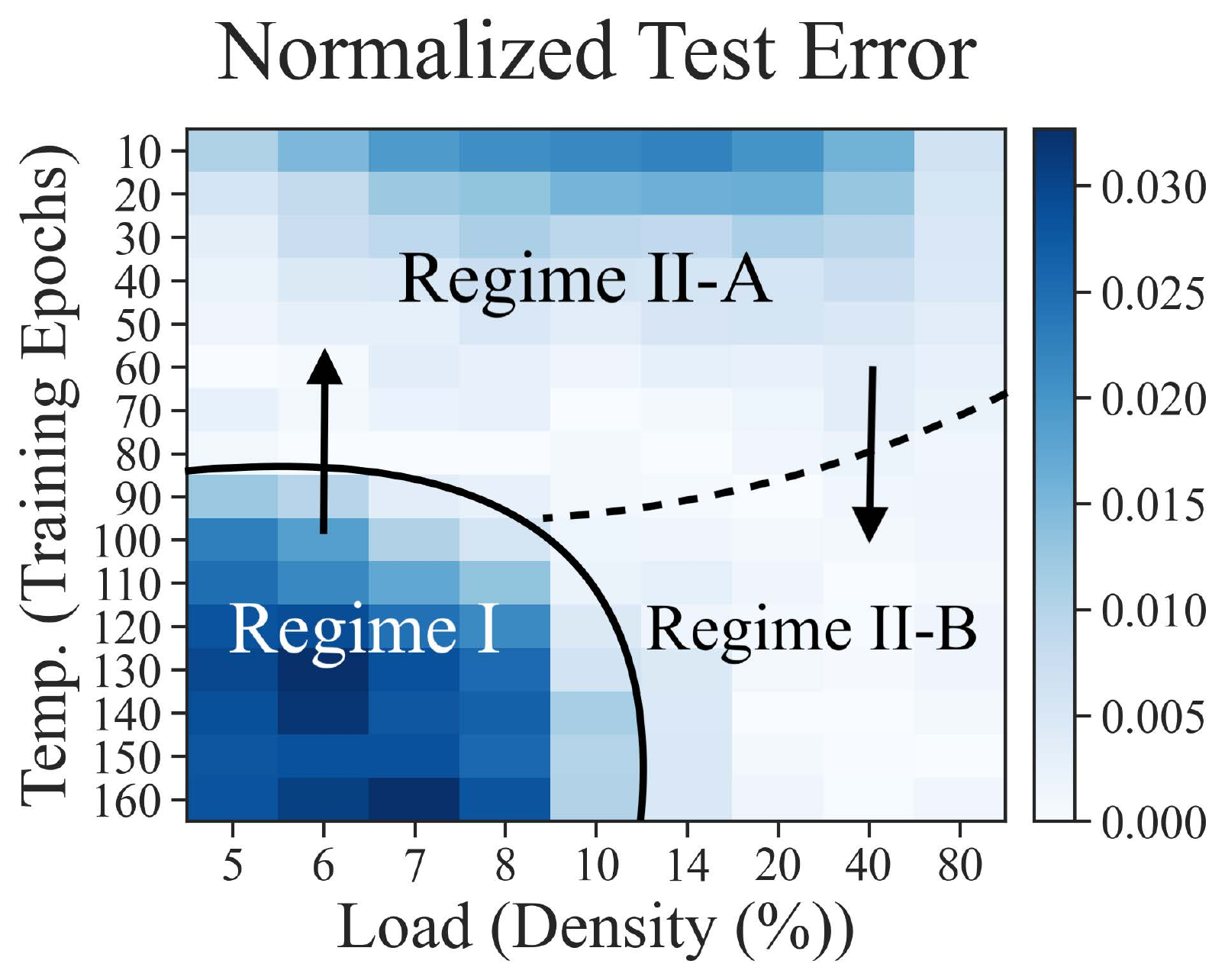}
       \caption{Normalized Test error
       }
       \label{fig:earlystop_norm_testerror}
   \end{subfigure} 
   \begin{subfigure}{0.24\linewidth}
       \centering
       \includegraphics[width=\linewidth]{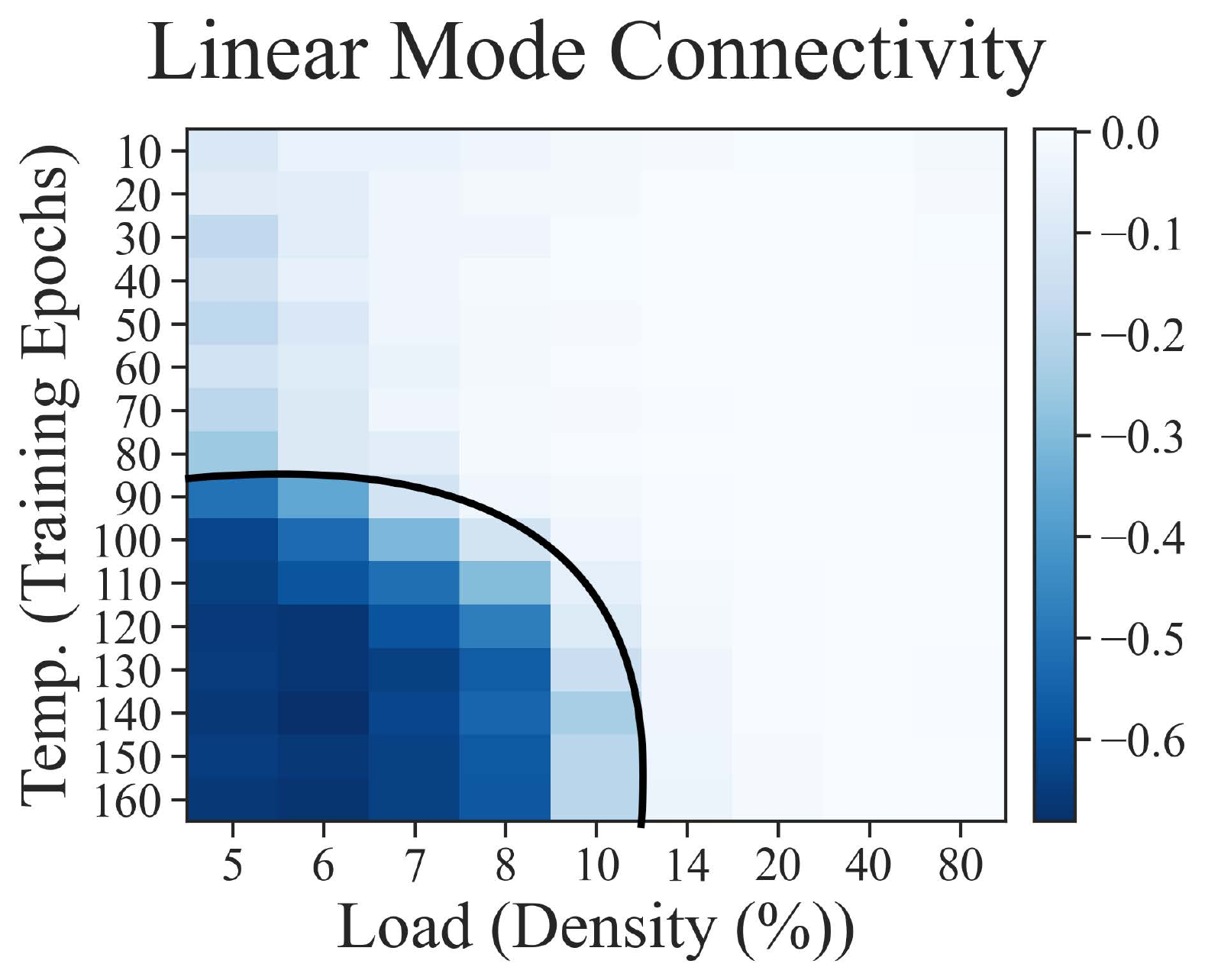}
       \caption{LMC 
       }
       \label{fig:earlystop_linear_mc}
   \end{subfigure}
   \begin{subfigure}{0.24\linewidth}
       \centering
       \includegraphics[width=\linewidth]{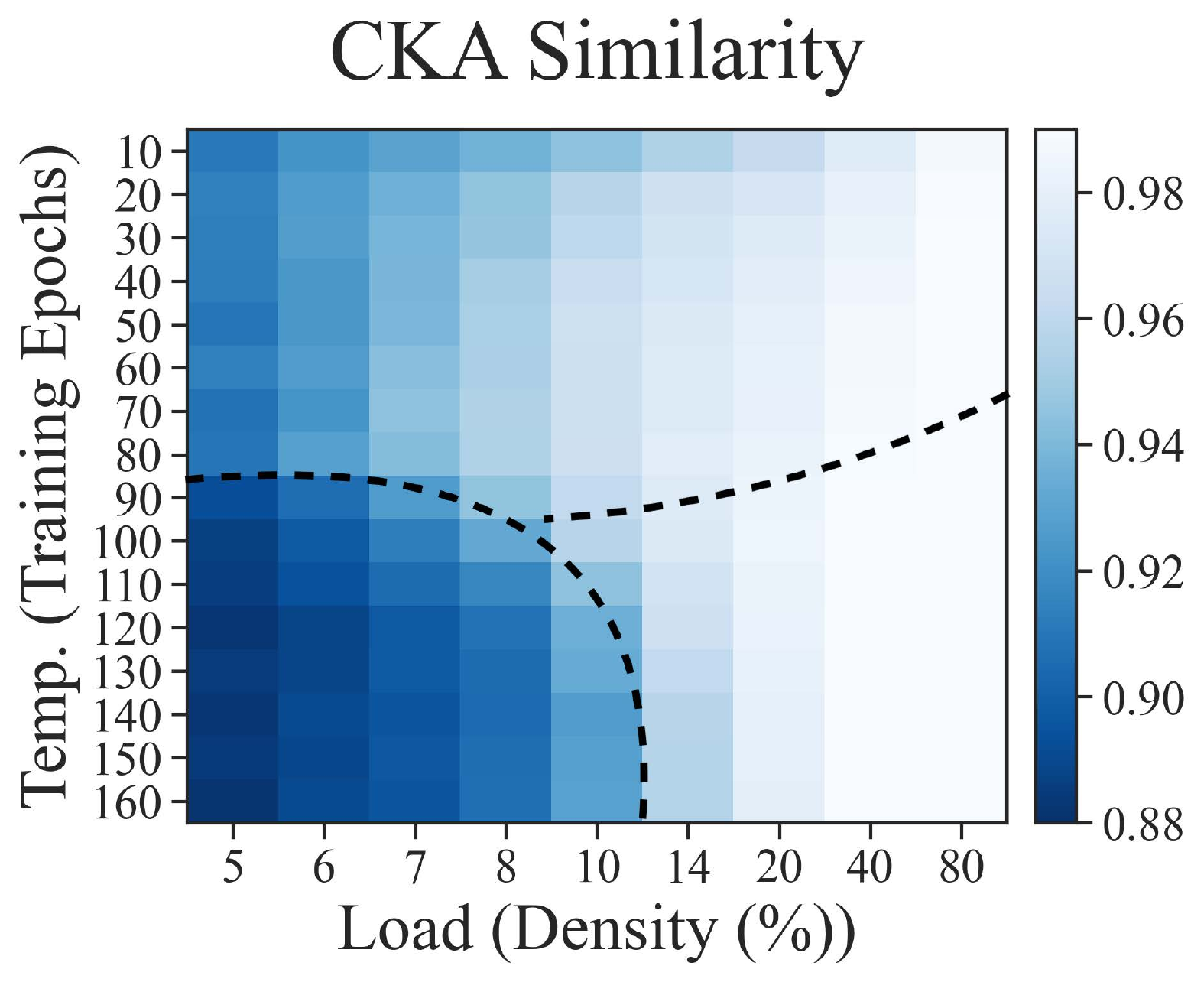}
       \caption{CKA Similarity 
       }
       \label{fig:earlystop_cka_simi}
   \end{subfigure}
    \vspace{-1mm}
    \caption{Partitioning the 2D model density (load) -- training epoch (temperature) diagram into three regimes. 
    Models are trained with PreResNet-20 on CIFAR-10.  
    The $y$-axis denotes a temperature-like parameter, indicated by a range of training epochs preceding the pruning process, while the $x$-axis represents a load-like parameter, expressed through diverse model densities applied to the model.
    (a) Final test error of the models after pruning and retraining. (b) Normalized test error is obtained by subtracting the optimal (lowest) test error from each column of the diagram in (a). The black arrows indicate two favorable transitions to lower test error regimes given a fixed model density.
    (c) LMC forms a sharper boundary that distinguishes Regime I from Regime II.
    (d) CKA shows a smooth transition that categorizes Regimes II-A and II-B.}
    \label{fig:earlystop-regime}
\vspace{-3mm}
\end{figure*}

\subsection{Loss landscape metrics}
\label{def:lossland-metric}
Recent work \citep{yang2021taxonomizing} proposed an extensive taxonomy of the loss landscape of realistic NNs, finding (among other things) that  LMC and CKA similarity can be used to determine the phase in which a trained model lies. 
In this study, we leverage these metrics to pinpoint the particular regime within our three-regime model to which a pruned NN belongs.

\noindent 
\textbf{LMC.}
\citet{frankle2020linear} shows that linear low-loss paths can be found between two networks if they originate from shared trained initialization, motivating us to take the linear variant of the mode connectivity used in \citet{yang2021taxonomizing}.
Given two separate sets of weights $\Theta, \Theta^{\prime}$, we parameterize the linear path $\gamma(t), t \in[0,1]$ connecting them with $
\gamma(t) =t \Theta +(1-t) \Theta^{\prime}
$.
Then, the LMC is defined as 
\begin{equation}
\begin{split}
\label{eq:mode_conn}
    \texttt{lmc}\left(\Theta, \Theta^{\prime}\right)=&\frac{1}{2}\left(
    \operatorname{error}_{\text {train }} (\Theta)
    +
    \operatorname{error}_{\text {train }}(\Theta^{\prime})
     \right) \\
     &- 
     \operatorname{error}_{\text {train }}\left(\gamma\left(t^*\right)\right),
\end{split}
\end{equation}
where $\gamma(t)= t \Theta + (1-t) \Theta^{\prime}$ and $t^*$ maximizes $t \mapsto\left|\frac{1}{2}\left(\operatorname{error}_{\text {train }}(\Theta)+\operatorname{error}_{\text {train }}\left(\Theta^{\prime}\right)\right)-\operatorname{error}_{\text {train }}\left(\gamma(t)\right)\right|$, $t \in [0,1]$.

We consider two cases for LMC.
If $\texttt{lmc} (\Theta, \Theta^{\prime}) \approx 0$, then this implies a curve of low training error connecting $\Theta, \Theta^{\prime}$, rendering the loss landscape well-connected.
If $\texttt{lmc}(\Theta, \Theta^{\prime}) < 0$, then this means that there is a ``barrier'' of high training error between $\Theta$ and $\Theta^{\prime}$.
In this case, we say that the loss landscape is poorly connected (or that the LMC is poor).\footnote{One difference from~\citet{yang2021taxonomizing} is that we study the ``linear version'' of mode connectivity~\cite{frankle2020linear}, which provides faster computation and is easier for downstream tasks, as discussed in Section~\ref{sec:app}. We also note that~\citet{yang2021taxonomizing} discuss cases when the mode connectivity is larger than 0, which happens rarely when we use the linear version of mode connectivity.} 

\noindent 
\textbf{CKA similarity.}
CKA similarity measures the representational similarity of two parameter configurations $\Theta$, $\Theta^{\prime}$ in the output space, as measured by the centered kernel alignment (CKA) metric. 
Following~\citet{yang2021taxonomizing}, we measure the distance between two models using their predictions, instead of weights, to avoid having low similarity resulting from different weights, even though the predictive functions are similar.
Let $\{\mathbf{x}_1, \ldots, \mathbf{x}_s\}$ denote a set of $s$ randomly sampled datapoints, and
$F_\Theta=\left[\begin{array}{lll}
f\left(\mathbf{x}_1;\Theta\right) & \cdots & f\left(\mathbf{x}_s;\Theta\right)
\end{array}\right]^{\top} \in \mathbb{R}^{m \times d_{\text {out }}}$ 
denote the concatenation of the outputs of the network.
In this case, the CKA similarity between the two weights $\Theta$ and $\Theta'$ is given by 
\vspace{-0.5mm}
\begin{equation}
\label{eq:cka}
\operatorname{cka}\left(\Theta, \Theta^{\prime}\right)=\frac{\operatorname{Cov}\left(F_\Theta, F_{\Theta^{\prime}}\right)}{\sqrt{\operatorname{Cov}\left(F_\Theta, F_\Theta\right) \operatorname{Cov}\left(F_{\Theta^{\prime}}, F_{\Theta^{\prime}}\right)}} ,
\end{equation}
\vspace{-0.5mm}
where $\operatorname{Cov}(X, Y)=(m-1)^{-2} \operatorname{tr}\left(X X^{\top} H_m Y Y^{\top} H_m\right)$, for $X, Y \in \mathbb{R}^{s \times d}$, and $H_m=I_m-m^{-1} \mathbf{1 1}^{\top}$ is a centering~matrix.

Note that measuring the LMC and the CKA similarity defined above requires two distinct weights $\Theta$ and $\Theta^{\prime}$ trained on the same sparse architecture. 
To obtain the two weights, we first prune a trained full dense model, and we then retrain two copies of the pruned sparse model with different SGD noise for $\alpha$ epochs. 
Formally, these two weights are defined as $\Theta = \operatorname{Train}\left(\Theta \odot \mathcal{M}, \xi^1, T_\text{epoch} \right)$ and $\Theta^{\prime} = \operatorname{Train}\left(\Theta \odot \mathcal{M}, \xi^2, T_\text{epoch} \right)$, $T_\text{epoch} = \alpha$.

\section{Validity of the Three-regime Model}
\label{sxn:empirical_validity}

In this section, we report our empirical results, illustrating the validity of the three-regime model we introduce.

\vspace{-1mm}
\subsection{Experimental setup}
\label{sec:setup}

We generate thousands of pruned models by systematically varying the load and temperature parameters.
A variant of residual networks with 20 layers (PreResNet-20) was trained on CIFAR-10 with various temperature hyperparameters and pruned to different load (density) levels.

We tune the following hyperparameters to vary the magnitude of load parameters and temperature parameters:
\begin{itemize}
    \item 
    Model density (Load): 
    we consider the model pruned to 9 densities: \{5, 6, 7, 8, 10, 14, 20, 40, 80\}\%. 
    \item 
    Number of training epochs (Temperature): 
    we consider training to numbers of epochs that are multiples of 10 \{10, 20, 30, ... , 160\}. Training for a total of 160 epochs corresponds to no early stopping. 
    \item 
    Batch size (Temperature):
    we consider training the model with varying batch sizes \{16, 21, 27, 32, 38, 44, 52, 64, 92, 128, 180, 256, 512\} while keeping the same amount of training iterations. 
    
\end{itemize}

We study the \textsc{UniformMP} as the basic pruning method. More details on the experimental setup can be found in \cref{sec:imple-detail-taxonomy}.

\vspace{-1mm}
\subsection{Regimes of loss landscape}
\label{sec:phase-plots}

We summarize the results in Figure \ref{fig:earlystop-regime}. 
Each pixel within the sub-figures represents a unique experimental configuration, where a model is trained for a specific number of epochs ($y$-axis) and then pruned to a particular density ($x$-axis).
Figure \ref{fig:earlystop_testerror} displays the test error of each model after pruning and retraining, while Figure \ref{fig:earlystop_norm_testerror} showcases the test error normalized using a scheme applied to each model density (column). 
This normalization process involves subtracting the test error by the optimal early-stopped test error for each density-related column.
Additionally, Figure \ref{fig:earlystop_linear_mc} and \ref{fig:earlystop_cka_simi} present the LMC and CKA of the pruned models, respectively.
The normalization scheme applied to Figure \ref{fig:earlystop_testerror} results in Figure \ref{fig:earlystop_norm_testerror}, enabling the comparison of performance differences across various regimes at each density level. 
Notably, Figure \ref{fig:earlystop_norm_testerror} demonstrates that Regime II-A typically surpasses Regime I in performance but falls short when compared to Regime~II-B.

We observe that the connectivity and similarity metrics can be used to find the transitions observed across the model density (load) -- training epochs (temperature) phase space, forming a three-regime classification that classifies the loss landscape, as shown in Figure \ref{fig:caricature}.
\begin{itemize}
  \item 
  \textbf{LMC distinguishes models with poorly connected loss landscape, which we categorize as Regime I}. 
  The first transition is displayed in Figure \ref{fig:earlystop_linear_mc}. 
  The white region represents the near-zero LMC, which implies a flat curve in the loss landscape between two local minima; and the dark blue region represents the negative LMC, which implies a high loss barrier moving from one local minimum to another. 
  The transition in LMC forms a curve separating Regime I from Regime II.
  
  \item 
  \textbf{CKA similarity further categorizes models with well-connected loss landscapes into models with similar/dissimilar outputs.} 
  In Figure \ref{fig:earlystop_cka_simi}, CKA divides the region where LMC is near-zero in Figure \ref{fig:earlystop_linear_mc} into Regime II-A and Regime II-B, the latter including models with more similarity.\footnote{The transition between Regime II-A and Regime II-B is much smoother, and it need not be viewed as a finite-sized approximation to a ``phase transition,'' in contrast to the much sharper transition between Regime I from Regime II that LMC identifies.} 
  Regime I observes a much smaller CKA similarity than the other two regimes, which coincides with negative LMC. 
  That the CKA similarity diagram exhibits a smooth transition between regimes, rather than a sharp transition from a negative value to zero, as is observed in LMC, is consistent with results in \citet{yang2021taxonomizing}.
\item 
\textbf{To enumerate the three regimes}:
\begin{itemize}
    \item \textbf{Regime I:} Poorly connected loss landscape, less similar output representations.
    \item \textbf{Regime II-A:} Well-connected loss landscape, less similar output representations.
    \item \textbf{Regime II-B:} Well-connected loss landscape, relatively large output similarity.
\end{itemize}
\end{itemize}

Based on the three-regime taxonomy, we assert the following as the central claim of this work. 

\begin{tcolorbox}
\textbf{Main claim.} 
Given a model, we can use the loss landscape metrics to inform the optimal temperature to apply in the training stage at each possible level of its load (model density) after pruning.
\end{tcolorbox}

To elaborate further, we mainly use the connectivity and similarity of the loss landscape, measured by LMC and CKA similarity respectively.
Low-density models with poor connectivity benefit from increased temperature, facilitating a transition from Regime I to Regime II-A. Conversely, high-density models with good connectivity benefit from decreased temperature, enabling a transition from Regime II-A to Regime II-B to improve similarity.

\subsection{Corroborating results} 
\label{sec:corrob-results}

We have additional results that modify the basic setup of Section \ref{sec:phase-plots}.
These are described in more detail in Appendix~\ref{sec:corrob-taxonomy}.
In short, we have studied batch size as an alternative temperature-like parameter, network architectures (DenseNet-40, VGG-19), datasets (CIFAR-100, SVHN), different pruning strategies (\textsc{GlobalMP}), different optimizers (Adam) and different task (machine translation).
In each case, we obtain results that are qualitatively similar to the results described in Section~\ref{sec:phase-plots}, thereby corroborating our main claim more generally.

\begin{table*}[ht]
\centering
\caption{The range of hyperparameters in the experiment setup: a dense model is trained with initial temperature $T$ and subsequently pruned to target level of load $L$.}
\label{tab:load-temp}
\begin{tabular}{@{}c|c|c|c@{}}
\toprule
\multicolumn{2}{c}{Initial temperature to adjust} & \multicolumn{2}{c}{Target level of load} \\ \cmidrule(r){1-2} \cmidrule(l){3-4}
Training Epochs (\small{$T_{\text{epoch}}$})  & Batch Size (\small{$T_{\text{batch}}$})  & Density (\small{$L_{\text{density}}$}) & Width Scalings (\small{$L_{\text{width}}$})  \\ \midrule
\small{160}   &  \small{128}  & \small{$L_{\text{density}}$} $\subseteq$ \{5, 7, 10, 14, 20, 40\}\% & \small{$L_{\text{width}}$} $\subseteq$ \{0.5, 1, 2, 4, 6, 8\}  \\ \bottomrule
\end{tabular}
\end{table*}

\begin{figure*}[th!]
\begin{centering}
    \resizebox{\textwidth}{!}{
	\begin{tabular}{c|c|c|c}
	\toprule
    \multicolumn{2}{c|}{Varying $L_\text{density}$} & \multicolumn{2}{c}{Varying $L_\text{width}$} \\
    \toprule 
    $T_\text{epoch} \times L_\text{density} $  &  $T_\text{batch} \times L_\text{density}$ &   $T_\text{epoch} \times L_\text{width}$  & $T_\text{batch} \times L_\text{width}$ \\
    \midrule

    \includegraphics[width=0.24\linewidth,keepaspectratio]{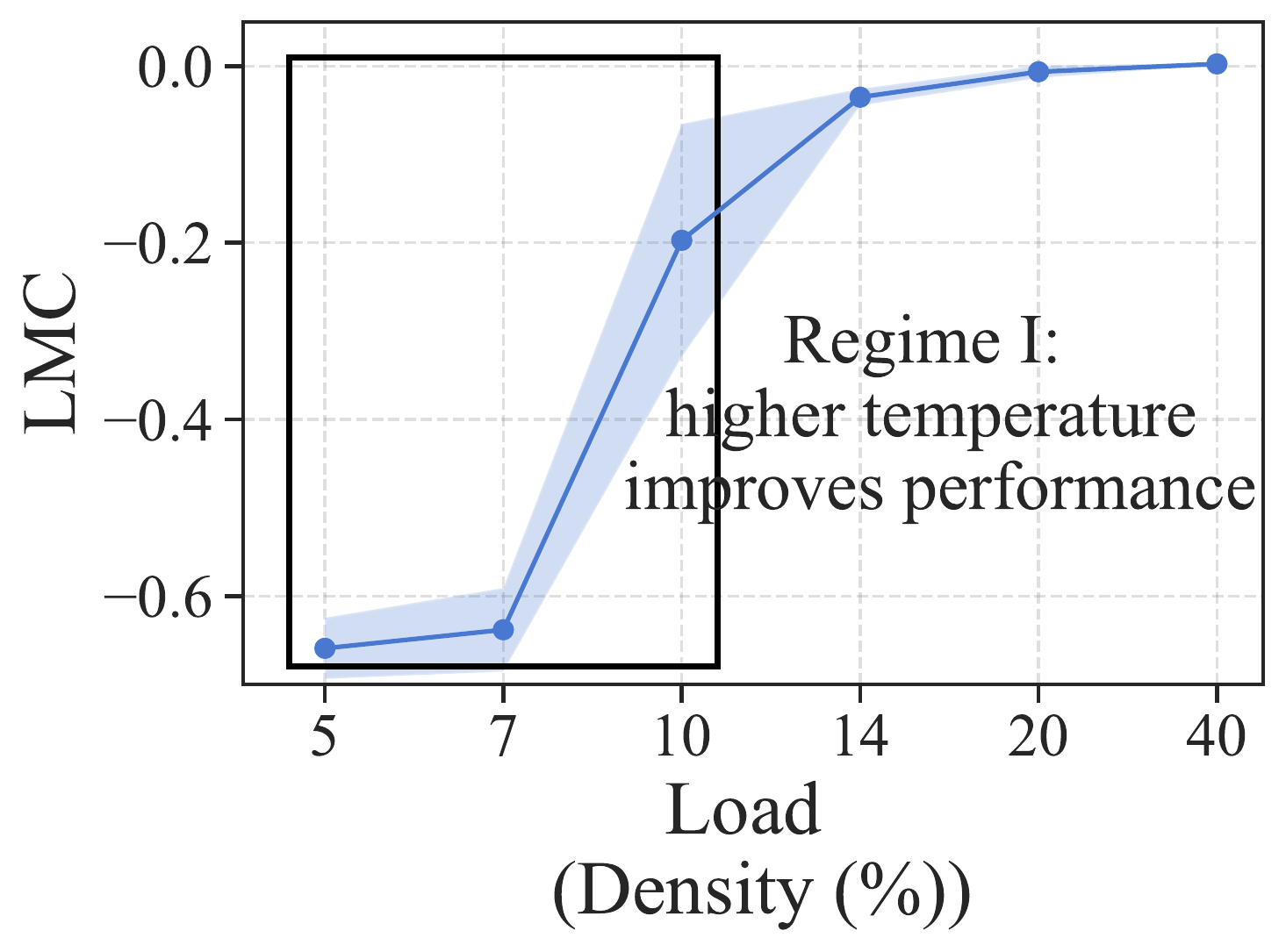} &\includegraphics[width=0.24\linewidth,keepaspectratio]{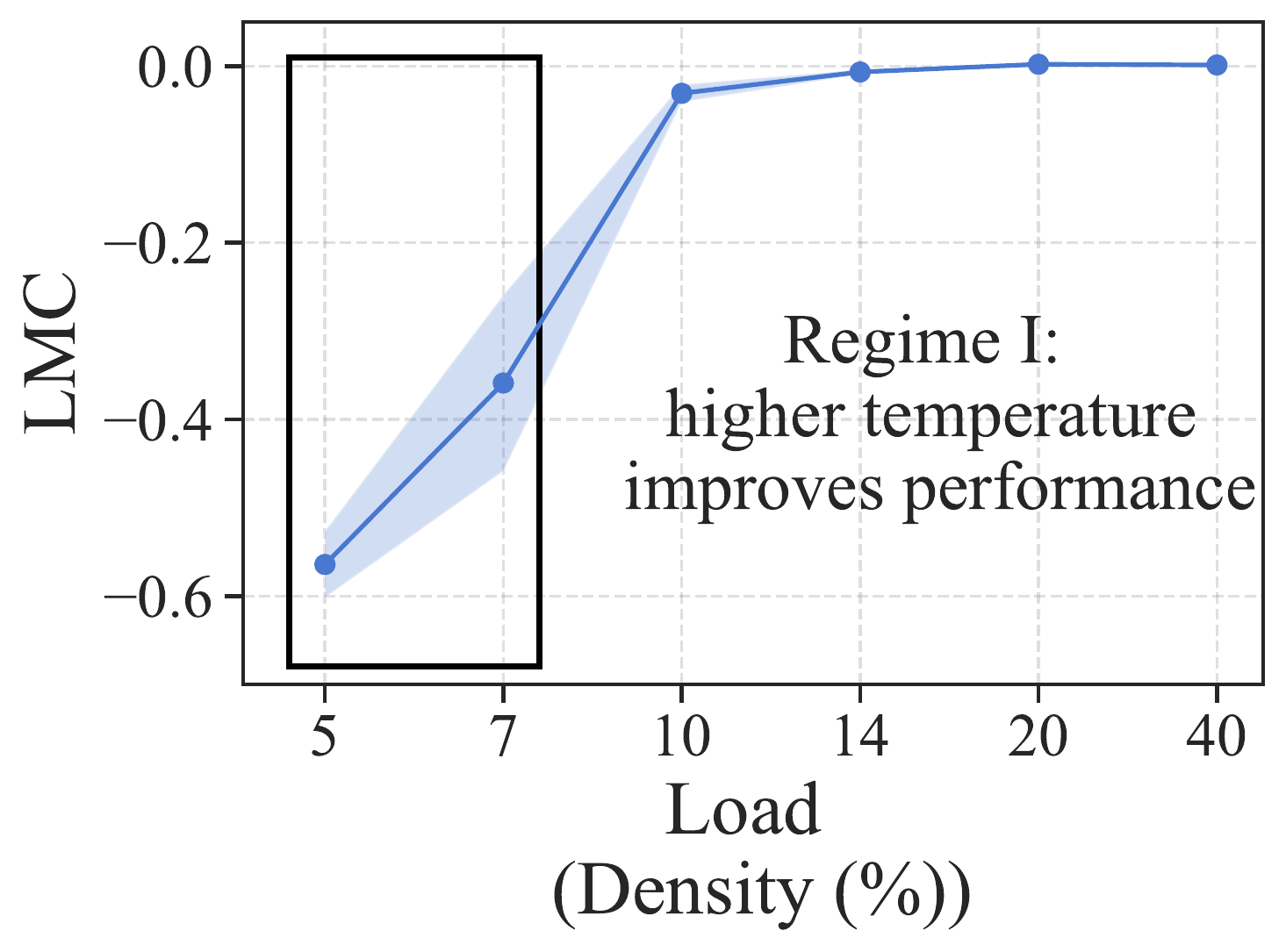} &
	\includegraphics[width=0.24\linewidth,keepaspectratio]{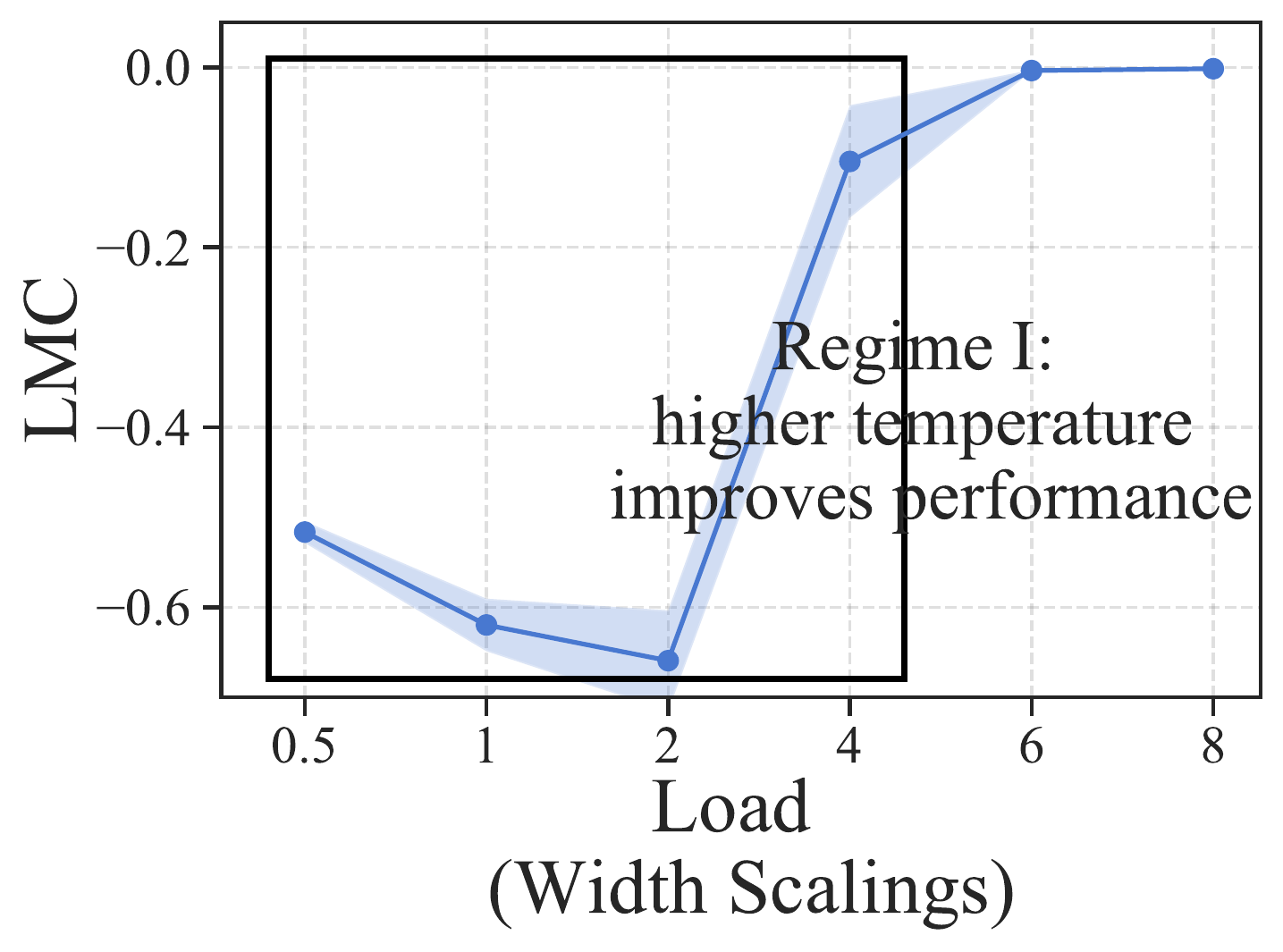} &
	\includegraphics[width=0.24\linewidth,keepaspectratio]{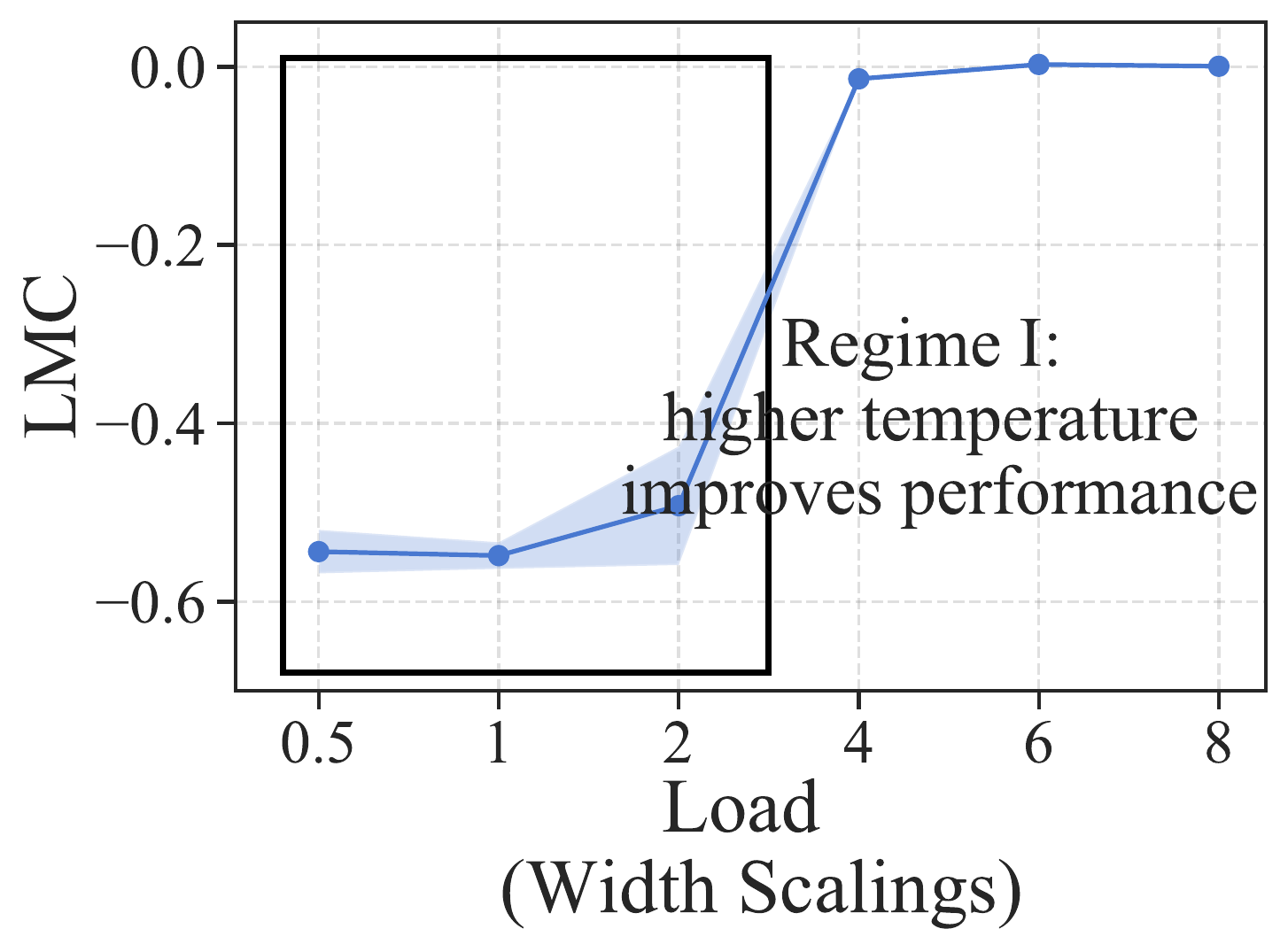} \\
 
	\includegraphics[width=0.24\linewidth,keepaspectratio]{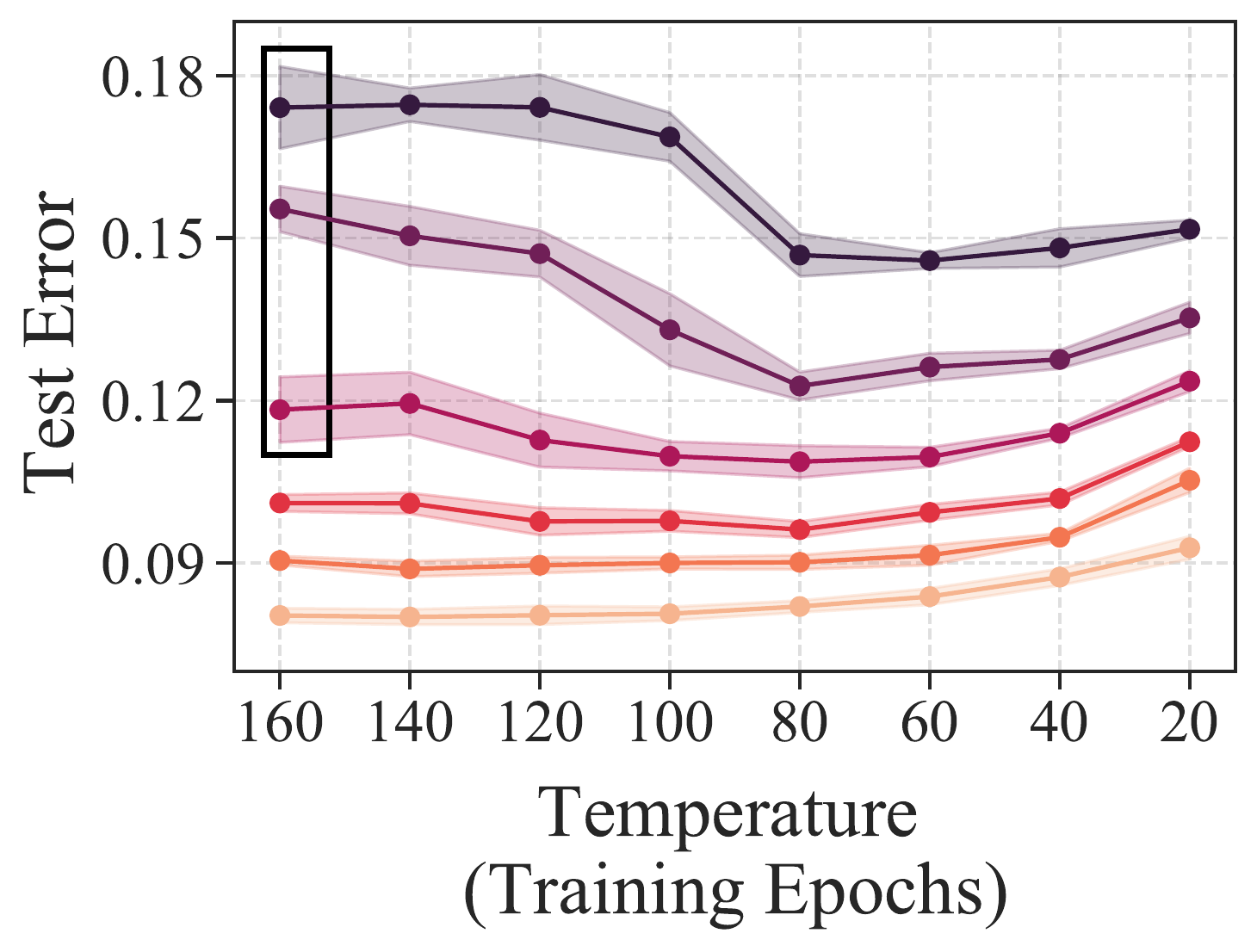}
  &\includegraphics[width=0.24\linewidth,keepaspectratio]{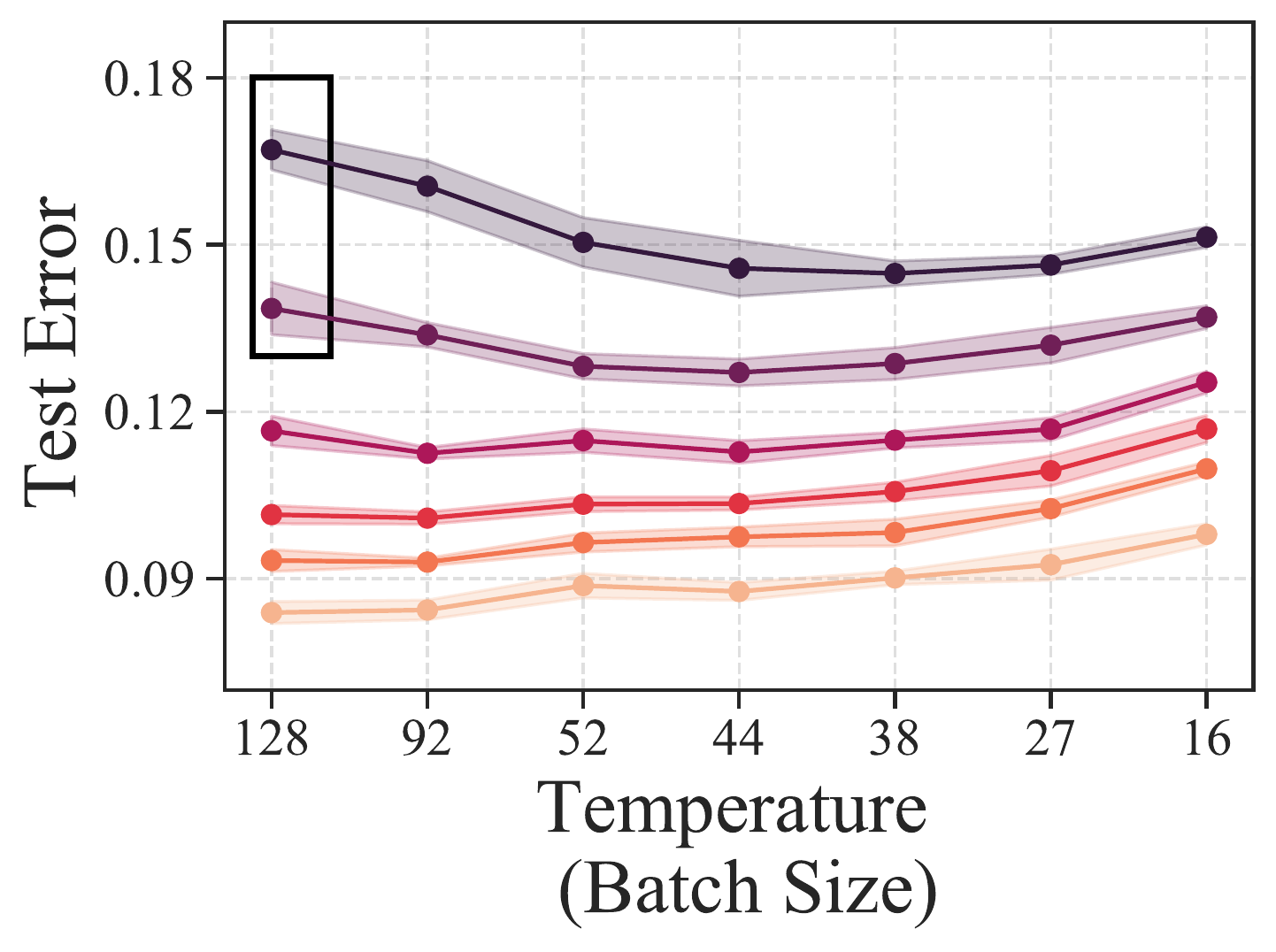} &
	\includegraphics[width=0.24\linewidth,keepaspectratio]{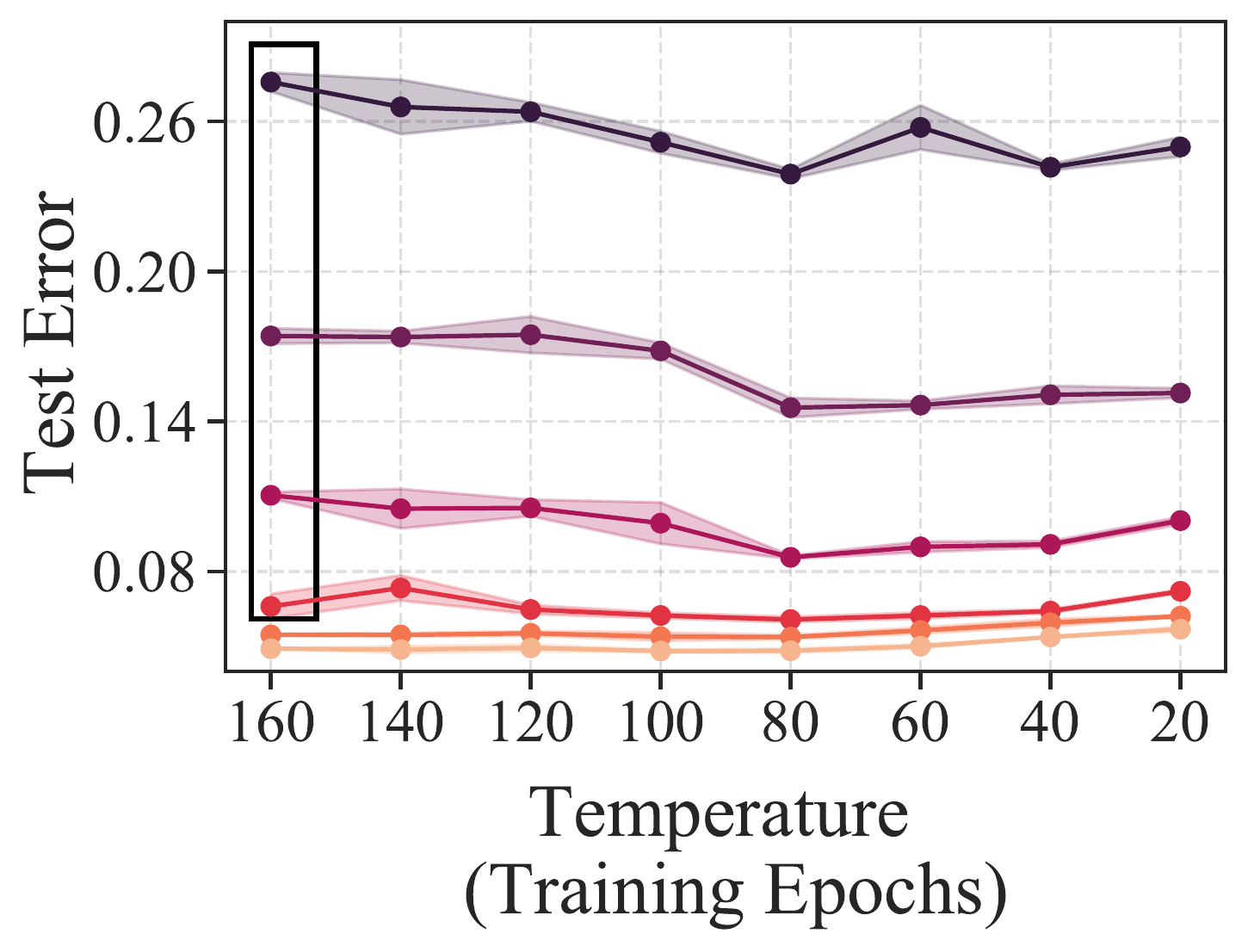} &
	\includegraphics[width=0.24\linewidth,keepaspectratio]{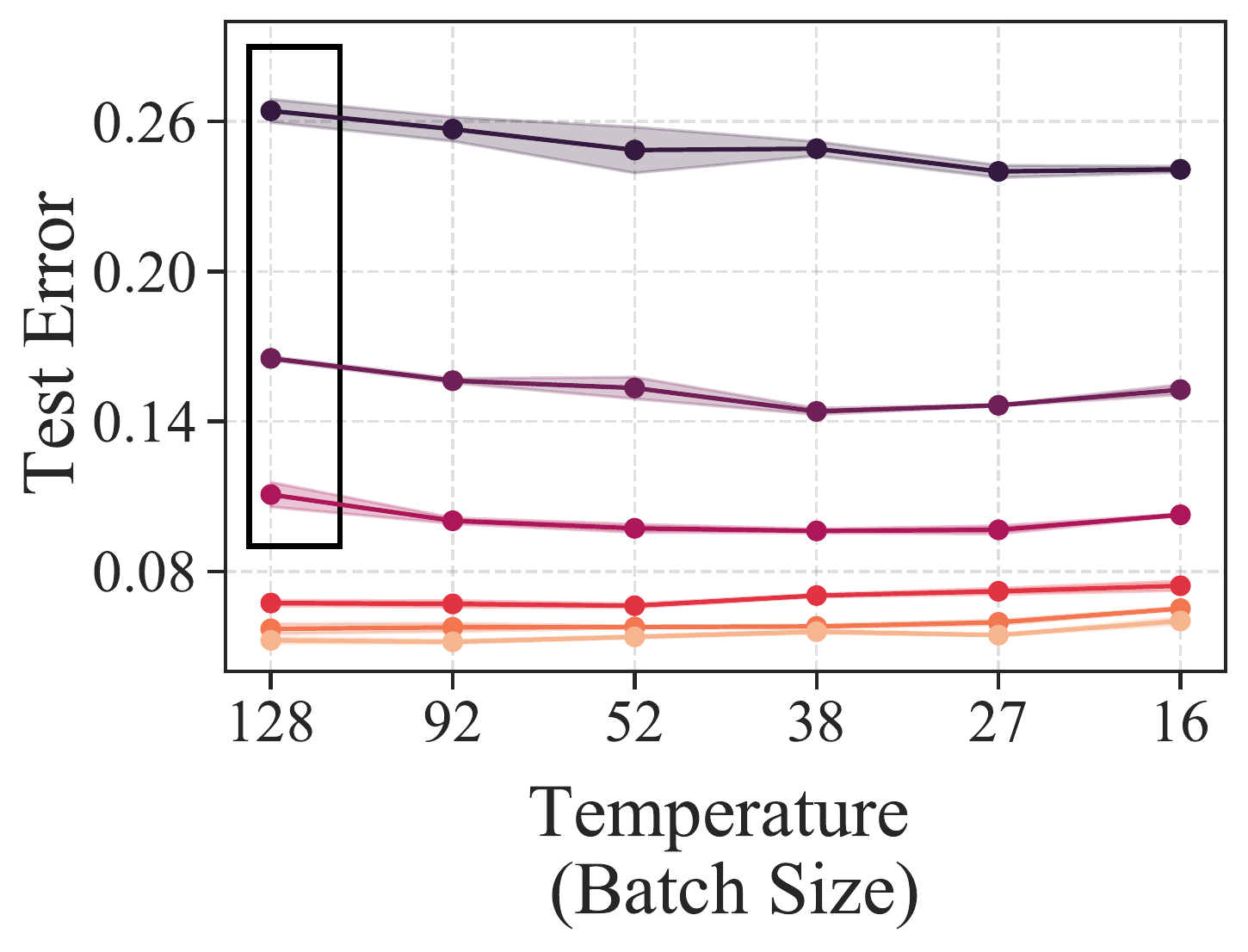} \\
	
    \multicolumn{2}{c|}{\includegraphics[width=0.50\linewidth,keepaspectratio]{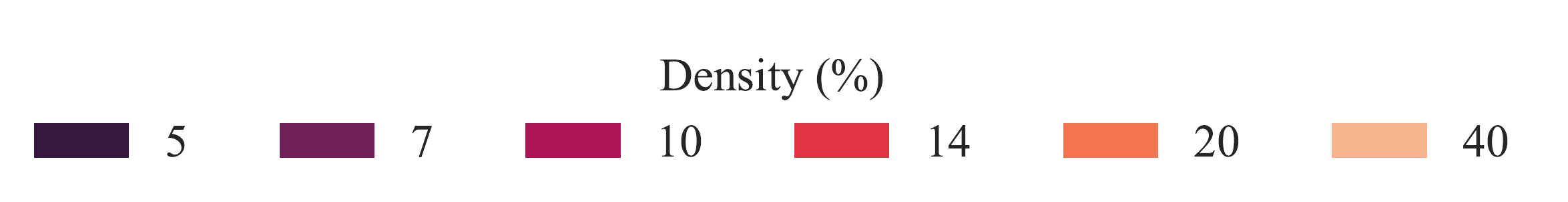}} &
    \multicolumn{2}{c}{\includegraphics[width=0.50\linewidth,keepaspectratio]{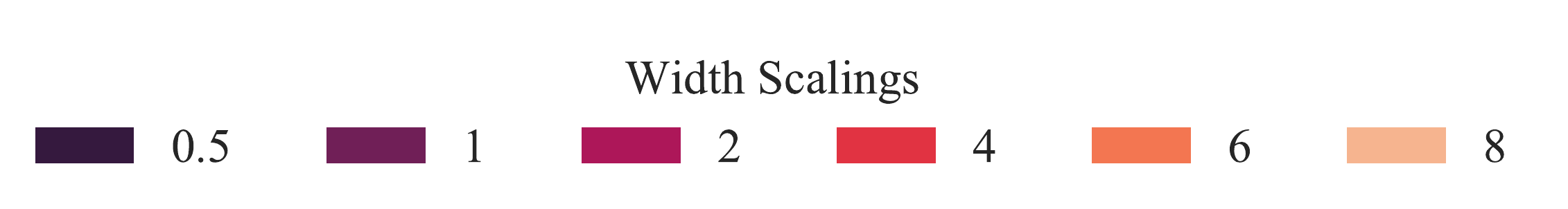}}
	\end{tabular}
 }         \vspace{-5mm}
	\caption{Using LMC to determine the right direction to adjust the temperature: models with negative LMC are located in Regime I (annotated by the black box), and their test error can be reduced by increasing temperature. Otherwise, models with close-to-zero LMC benefit from decreasing temperature. Note that a smaller training epoch or a smaller batch size corresponds to a higher temperature.
 } 
	\label{fig:density-width-vary}
\vspace{-2mm}
\end{centering}
\end{figure*}

\vspace{-1mm}
\section{Application of the Three-regime Model}
\label{sec:app}

In this section, we discuss several applications of our three-regime model.
\vspace{-1mm}
\subsection{Determining how to adjust temperature parameters}
\label{sec:app1}

Here, we develop a scheme to tune temperature parameters based on the three-regime model. 
In particular, we are interested in the following question:

\textbf{Problem statement.}
\emph{Given a model trained with a particular temperature-like parameter and subsequently pruned to a target level of load, can we determine how to adjust the magnitude of temperature to improve performance?}

\begin{algorithm}[tb]
   \caption{Temperature Tuning}
\begin{algorithmic}
   \STATE {\bfseries Input:} dense model initialization $\Theta_0$, initial temperature $T_0$ and increment $\Delta T$, target load $L$, 
   training procedure $\operatorname{Train}$ (Section \ref{def:preli}), pruning procedure $\operatorname{Prune}$ (Section \ref{def:network-prune}), LMC threshold $\epsilon$
   \STATE {\bfseries Output:}  Tuned temperature $T^{\prime}$
   \vskip 0.5em
   \STATE Train with temperature:  $\Theta$ = $\operatorname{Train}$ ($\Theta_0$, $T$)
   \STATE Prune to target load:  $\Theta \odot \mathcal{M}$ = $\operatorname{Prune}$ ($\Theta$, $L$) 
   \STATE Compute LMC on $\Theta \odot \mathcal{M}$ via  \cref{eq:mode_conn}
   \IF{$\text{LMC} < \epsilon$}
    \STATE $\text{Regime I} = true$
    \STATE $T^{\prime}$ = $T_0$ + $\Delta T$
   \ELSE
    \STATE $\text{Regime I} = false$
    \STATE $T^{\prime}$ = $T_0$
   \ENDIF
\end{algorithmic}
\label{alg:tmp-tune}
\end{algorithm}

\textbf{Experiment details.} 
As a concrete setup, we explore two hyperparameters to represent the load-like parameter $L$ of the pruned models: model density $L_\text{density}$ and width scaling $L_\text{width}$. It is worth noting that the unstructured pruning method cannot alter the model width; thus, to vary $L_\text{width}$ of the pruned model, we train multiple dense models with varying widths and prune them to identical densities.

Additionally, we consider two hyperparameters to represent the temperature-like parameter $T$ to be adjusted in the first stage of dense model training: training epoch $T_\text{epoch}$ and batch size $T_\text{batch}$. 
The ranges for these values are specified in Table \ref{tab:load-temp}. 
We use commonly used temperature values for our experiments: for CIFAR-10, a batch size of 128 and training epochs of 160 are commonly used. For instance, see Table 1 in \citet{frankle2020linear, liu2021sparse}.
Thus, it is reasonable to assume that a practitioner might start with these initial temperature values similar to our study.
We also note that when varying one specific parameter to control load, other control parameters are kept constant. 
For example, when varying $L_\text{density}$, $L_\text{width}$ is fixed at 1, and while varying $L_\text{width}$, $L_\text{density}$ is fixed at 5\%.
This is also applied when adjusting the temperature.
The experiment is conducted using PreResNet-20 on the CIFAR-10 dataset, with the models being pruned using the \textsc{UniformMP}.

\textbf{Conventional wisdom.} The conventional approach suggests that dense models should be trained to completion before being pruned~\citep{obd1989, han2015learning}.

\textbf{Proposed method.} 
We observe that the conventional wisdom of training to completion remains effective for pruned models belonging to Regime II, but loses its effectiveness for models in Regime I, where early stopping proves to be more useful. Note that recent studies also show improved pruning using early stopping~\citep{li2020train, liu2021sparse, Shen_2022_CVPR}, while our method here gives a more comprehensive way to view this problem.
Consequently, we propose a temperature-tuning method outlined in Algorithm~\ref{alg:tmp-tune}. This algorithm determines whether training to completion (smaller temperature) or doing early stopping (larger temperature) is necessary to enhance the performance of the pruned model.
Our approach assesses the LMC of the pruned models to identify whether they fall within the regime where conventional wisdom is less effective.
If the current model belongs to Regime II (LMC $\in [ \epsilon, 0]$), then conventional wisdom is effective, and we do not increase the temperature.
If the current model belongs to Regime I (LMC $< \epsilon$), conventional wisdom fails, and we increase the temperature.
We select a threshold value $\epsilon = -0.05$, and discuss the selection of the threshold in \cref{sec:thre-explain}.

\textbf{Results.} 
The results are presented in Figure \ref{fig:density-width-vary}. 
In this figure, the label $T \times L$ signifies that the target model load for pruning is $L$, and the initial temperature setting is $T$ (which is the temperature parameter that we need to adjust).
The four columns represent four distinct cases arising from the use of two specific hyperparameters to characterize $L$ and two other hyperparameters to characterize $T$, namely training epochs, batch size, density, and model width, as detailed in Table \ref{tab:load-temp}.

On the first row of Figure \ref{fig:density-width-vary}, we showcase the LMC results evaluated on a pruned model that has been trained with temperature $T$ and pruned to load $L$. Our method (Algorithm~\ref{alg:tmp-tune}) then identifies the regime to which the pruned model belongs and determines if increasing $T$ is beneficial for different $L$ values.
On the second row, we present the test error resulting from increasing the temperature for all load values, which is used to verify whether Algorithm \ref{alg:tmp-tune} selects the appropriate $L$ for increasing $T$ by examining if the test error decreases when $T$ is increased.

From Figure~\ref{fig:density-width-vary}, we see that Algorithm \ref{alg:tmp-tune} effectively identifies that multiple pruning settings in Table \ref{tab:load-temp} necessitate adjustments with higher temperatures. For instance, see the case $T_\text{epoch} \times L_\text{density}$. 
The LMC results in the first row indicate that three out of six $L_\text{density}$, specifically \{5, 7, 10\}\%, have an LMC $< \epsilon$ (highlighted by the black box), signifying that these settings (pruned models) belong to Regime I and require tuning with higher temperatures.
Indeed, the test error results in the second row demonstrate that increasing the temperature on these settings can reduce the test error.
On the other hand, the remaining $L_\text{density}$ \{14, 20, 40\}\% benefit from decreasing temperature.
We can find similar results on the other three columns, i.e., depending on the LMC value, one can determine whether to increase temperature or not.
In Appendix~\ref{sec:hyper-tune}, we study the third load parameter, model depth $L_\text{depth}$, and we get consistent results.

By utilizing the loss landscape measure LMC, our method efficiently determines the correct regime for hyperparameter tuning.
Thus, our method provides an efficient approach to predict the more efficient direction of tuning hyperparameters based on a single initial ($T$, $L$) pruning setting, thereby eliminating the need for costly grid searches. In contrast to the grid search, which typically necessitates at least two dense model training runs with different hyperparameters, our method accomplishes this with a single dense model training run.
Consequently, our method exhibits twice the efficiency of the grid search approach. 
Furthermore, our method works on a wide range of load-like parameters, making it applicable to different target model sizes.

\subsection{Selecting the best model without grid search} \label{sec:app2}

Here, we develop a model selection method based on the three-regime model. 
In particular, we are interested in the following question.

\textbf{Problem statement.}
\emph{Given a set of models trained with diverse magnitudes of temperature-like parameters (training epochs or batch sizes) and a target model density, which model should we prune to obtain optimal pruning performance?}

\textbf{Experiment details.} 
We study PreResNet-20 on CIFAR-10. 
We prune the model to 9 different model densities, as detailed in Section \ref{sec:setup}.
For each target model density, we select a model to prune from two sets: 
1) 16 different training epochs ranging from 10 to 160 and a fixed batch size; and 
2) 13 different batch sizes ranging from 16 to 512 and a fixed training epoch.

\begin{algorithm}[tb]
   \caption{Model Selection via LMC and test error}
   \begin{algorithmic}
   \STATE {\bfseries Input:} a set of trained dense models \{$\Theta_i\}_{i=1}^{n}$, and their test errors \{$\operatorname{error}_\text{test}(\Theta_i)\}_{i=1}^{n}$, 
   LMC threshold $\epsilon$, training procedure $\operatorname{Train}$ (Section \ref{def:preli}), pruning procedure $\operatorname{Prune}$ (Section \ref{def:network-prune}), target load $L$, retraining epochs $\alpha$
   \STATE {\bfseries Output:} dense model $\Theta_{i^{*}}$ to prune
   \vskip 0.5em
   \STATE $i^{*} = \mathop{\arg\min_i}\{ \operatorname{error}_\text{test}(\Theta_i) \}_{i=1}^{n}$
   \STATE $\Theta_{i^{*}} \odot \mathcal{M} = \operatorname{Prune}(\Theta_{i^{*}}, L)$
   \STATE Compute LMC on $\Theta_{i^{*}} \odot \mathcal{M}$ via \cref{eq:mode_conn}
   \IF{$\text{LMC} < \epsilon$}
    \STATE $\Theta_{i} \odot \mathcal{M}_{i}  = \operatorname{Train}(\operatorname{Prune}(\Theta_{i}, L) , \alpha)~\text{for}~i \in [1, n]$
    \STATE $i^{*} = \mathop{\arg\min}_i \{ \operatorname{error}_\text{test} (\Theta_{i} \odot \mathcal{M}_{i} ) \}_{i=1}^n$
   \ENDIF
\end{algorithmic}
\label{alg:model-sele-lmc}
\end{algorithm}

\textbf{Conventional wisdom.}
The conventional approach \cite{obd1989, han2015learning} to selecting a dense model to prune is based on the test error of the trained dense model.

\textbf{Proposed method.}
We find that the conventional wisdom is effective when the pruned model belongs to Regime II while it does not work when it belongs to Regime I.
Therefore, we propose an LMC-based model selection method (Algorithm \ref{alg:model-sele-lmc}), which uses the LMC metric to detect the wrong prediction made by conventional wisdom. 
Our approach uses the LMC metric to evaluate whether the dense model chosen by conventional means falls within the undesirable regime. If the model belongs to Regime II (LMC $\in [\epsilon, 0]$), conventional wisdom is effective, and the selection process concludes. Conversely, if the model falls under Regime I (LMC $< \epsilon$), conventional wisdom fails, and we evaluate alternative candidate models. We set LMC threshold $\epsilon = -0.05$, same as in \cref{sec:app1}. We set the retraining epochs $\alpha = 2$, which is much smaller than the 160 retraining epochs used by the grid search method.
We also consider the realistic case that the evaluation has no access to test data and propose an LMC-and-CKA-based method Algorithm \ref{alg:model-sele-lmc-cka}, which only requires access to training data.

\begin{figure}[!htb]
   \centering
  \begin{subfigure}{0.48\linewidth}
       \centering
       \includegraphics[width=\linewidth]{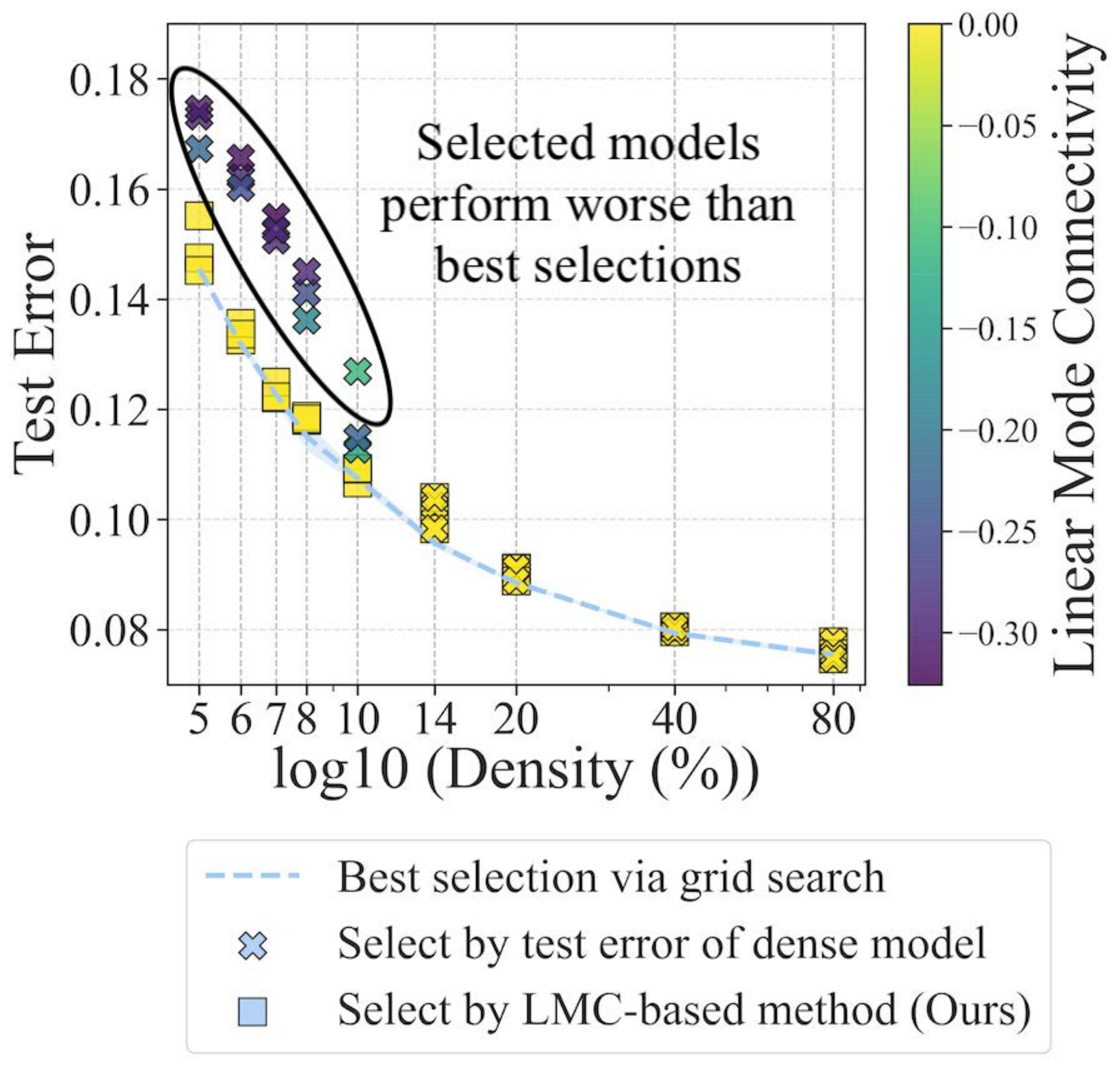}
       \caption{Selecting the best model from those trained with various training epochs.}
       \label{fig:prediction-via-metrics-epoch-fast}
   \end{subfigure}
   \hfill
    \begin{subfigure}{0.48\linewidth}
       \centering
       \includegraphics[width=\linewidth]{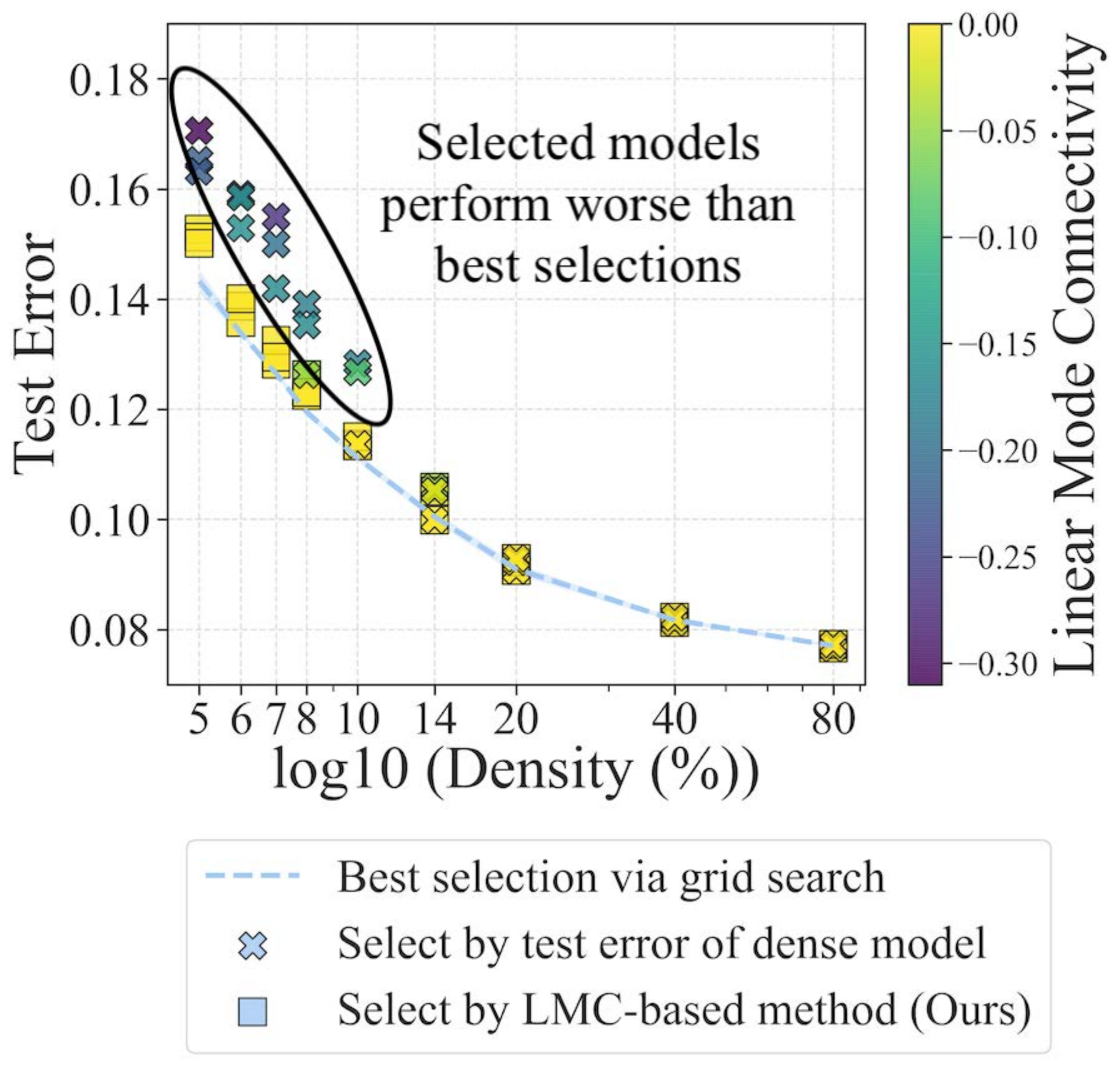}
       \caption{Selecting the best model from those trained with various batch sizes.}
       \label{fig:prediction-via-metrics-batch-size-fast}
   \end{subfigure} 
    \caption{Selecting temperature using the LMC-based method (squares) leads to a smaller test error than selecting temperature using the test error of the unpruned dense model (crosses). 
    The performance of LMC-based selection is close to the best test error found by grid search (dashed lines).
    (Left) Selecting the best training epoch. (Right) Selecting the best batch size. 
    Models that perform significantly worse than grid search tend to have worse LMC, shown by the dark color of markers.
    }
    \label{fig:prediction-via-metrics-fast}
\end{figure}

\textbf{Results.} 
We compare three model selection approaches, i.e., selecting by test error of dense model (conventional wisdom), selecting jointly by LMC and test error (ours, Algorithm \ref{alg:model-sele-lmc}), and best selections obtained by grid search that conducts the full train-prune-retrain procedure for all models in the set. 
The results of comparing the three model selection methods are presented in Figure \ref{fig:prediction-via-metrics-fast}.
The $x$ axis represents the model density, $y$ axis represents the final test error of the pruned models after full time of retraining. The color of the markers represents the value of LMC measured on the pruned models.
Figure~\ref{fig:prediction-via-metrics-epoch-fast} shows the results of selecting the best training epoch, while Figure~\ref{fig:prediction-via-metrics-batch-size-fast} shows the results of selecting the best batch size.

In both figures, we observe the following. 
1) The baseline method (select by test error of the dense model) demonstrates distinct performance in different regimes: it performs well in the large-density regime but poorly in the low-density regime, which is precisely characterized by the LMC measure (clear color transition from yellow to dark blue).
2) Our proposed LMC-based method achieves comparable performance in terms of final test error compared to grid search while significantly reducing the computational requirements. This efficiency is attributed to utilizing the LMC measure to determine whether the pruning configuration belongs to a regime where the baseline approach is effective. Specifically, in the high-density regime (\{14, 20, 40, 80\}\%), the LMC measure informs us to adopt the baseline solution without evaluating other candidates, thus avoiding unnecessary computations. Conversely, in the low-density regime (\{5, 6, 7, 8\}\%), we bypass the poor baseline solution identified by a low LMC value and perform computations (retraining 2 epochs for each candidate) to identify superior models with a high LMC value. The additional results of studying the test data-free case using the LMC-and-CKA-based method are provided in Appendix \ref{sec:sup-model-select-method}.

\FloatBarrier
\subsection{Tuning the temperature of the SAM method}  \label{sec:app3}

Here, we use the three-regime model to diagnose a dichotomous effect of training dense models with the optimizer SAM~\citep{foret2021sharpnessaware}, and we propose a hyperparameter tuning scheme to mitigate the negative effect.

\textbf{Problem statement.}
\emph{Given a target model density, how can we tune the hyperparameter of SAM optimizer to train the dense model for optimal pruning performance?}

\begin{figure}[ht!]
\begin{centering}
       \begin{tabular}{cc}
       \includegraphics[width=0.46\columnwidth,keepaspectratio]{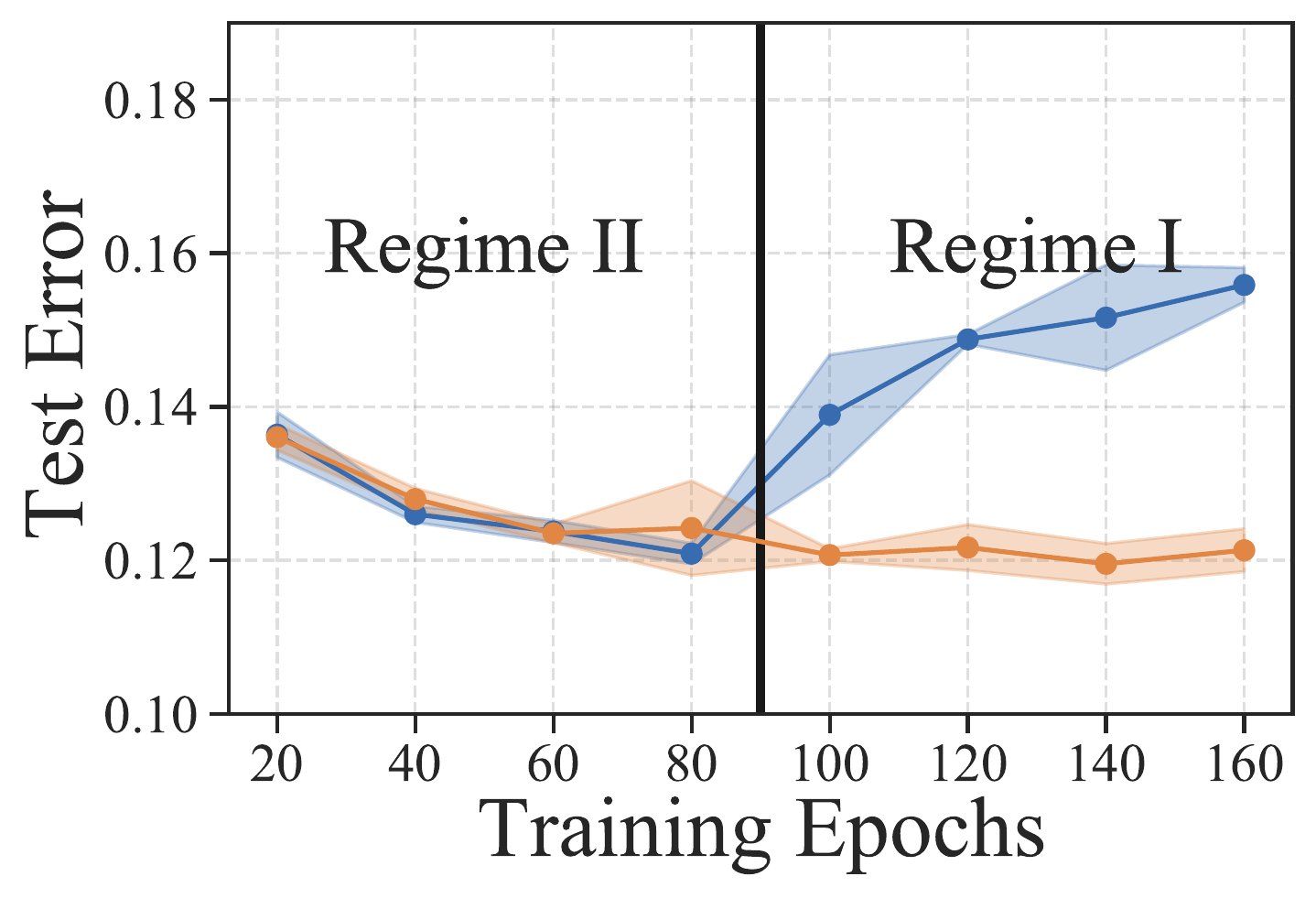}  &
	 \includegraphics[width=0.46\columnwidth,keepaspectratio]{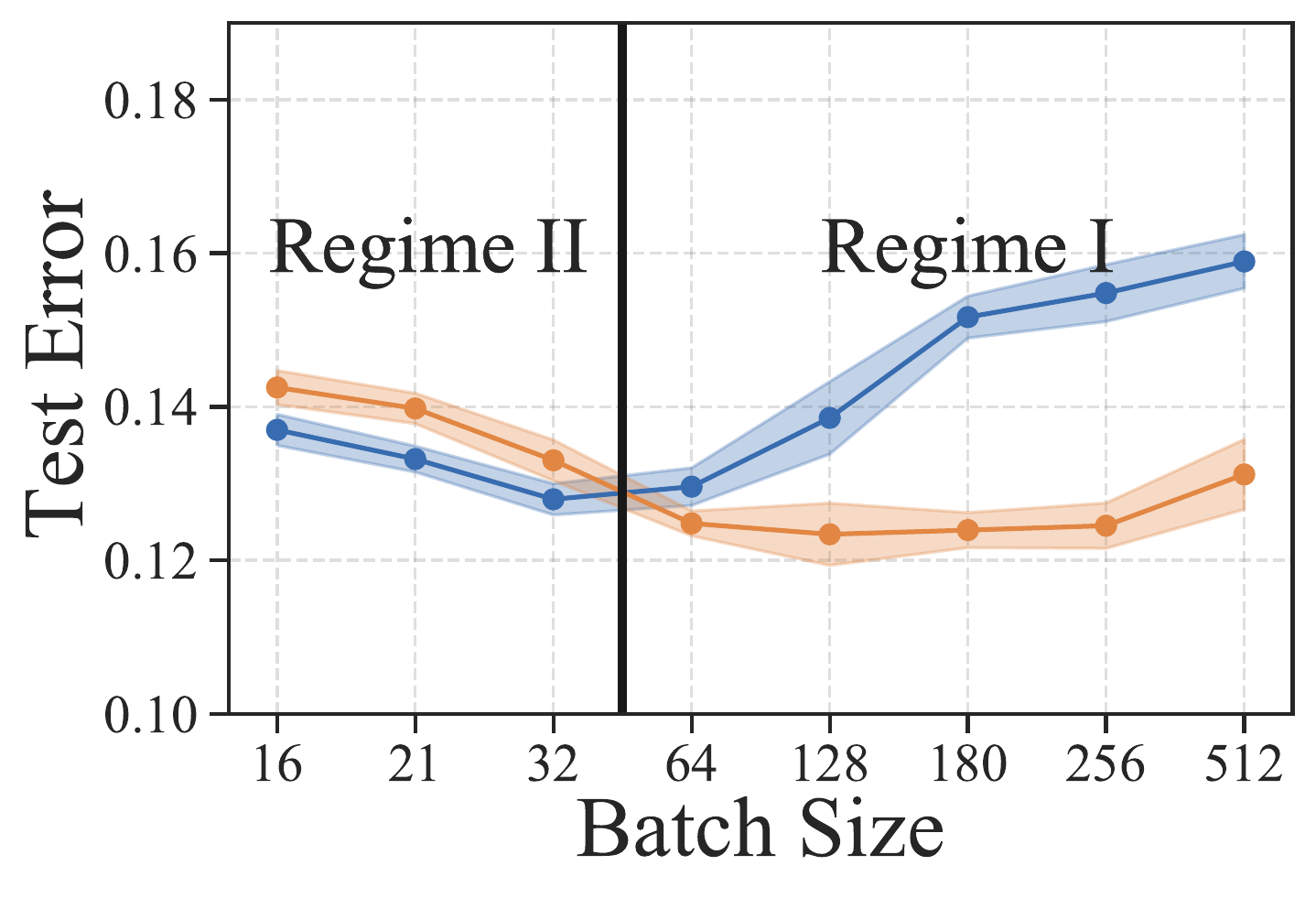} 
  
	\\
	\includegraphics[width=0.46\columnwidth,keepaspectratio]{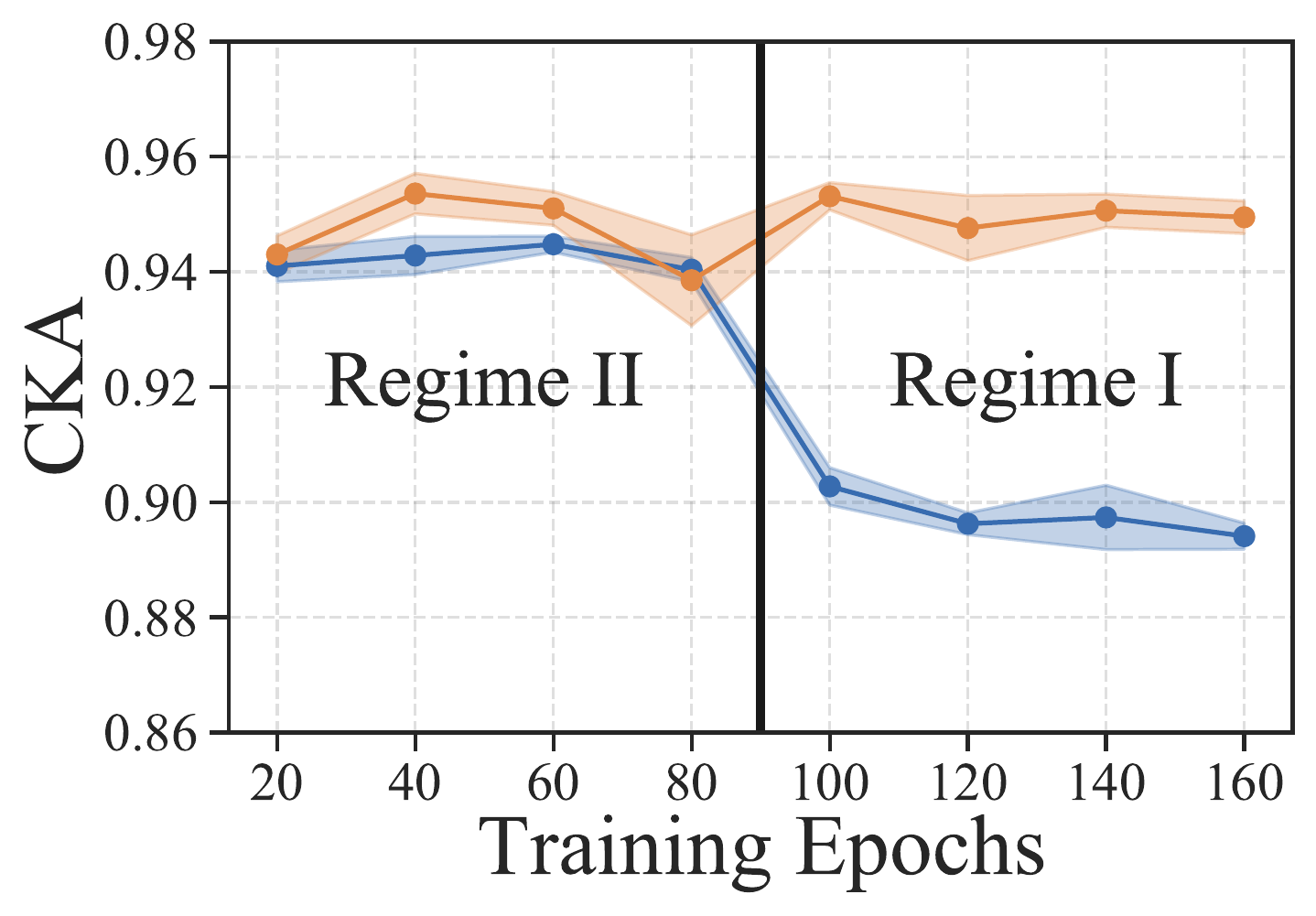} & 
	 \includegraphics[width=0.46\columnwidth,keepaspectratio]{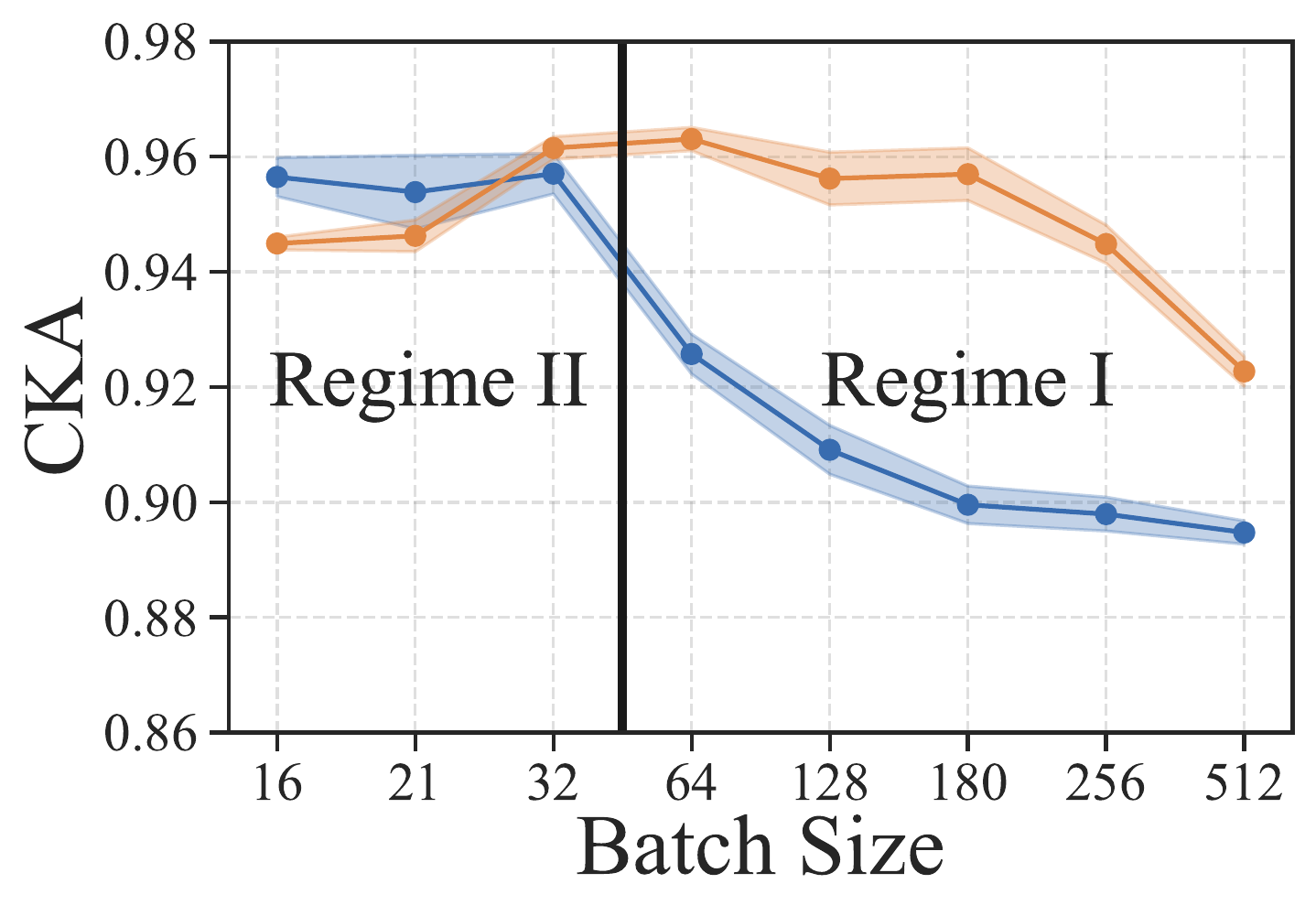}   

	\\
	\includegraphics[width=0.46\columnwidth,keepaspectratio]{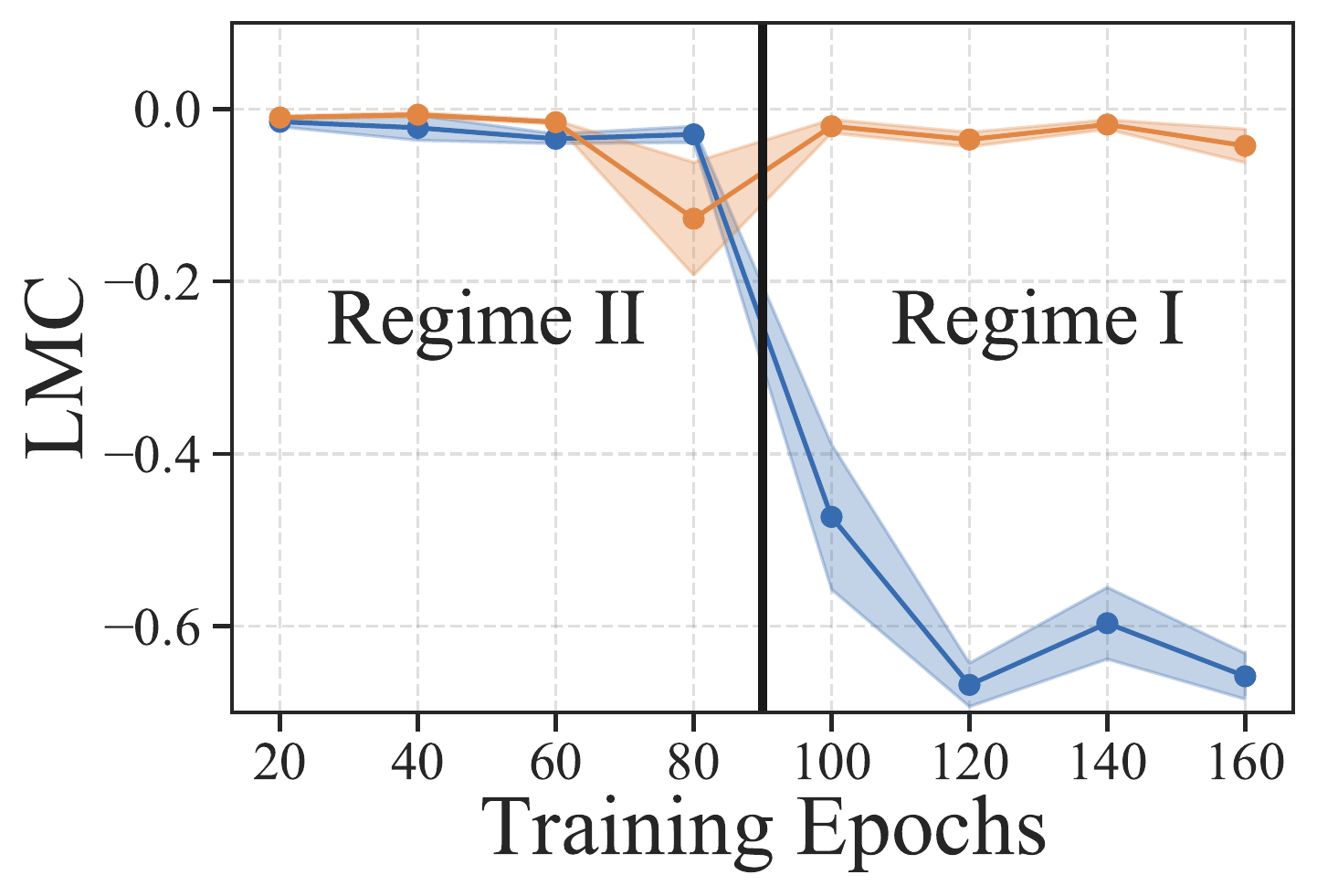} & 
	\includegraphics[width=0.46\columnwidth,keepaspectratio]{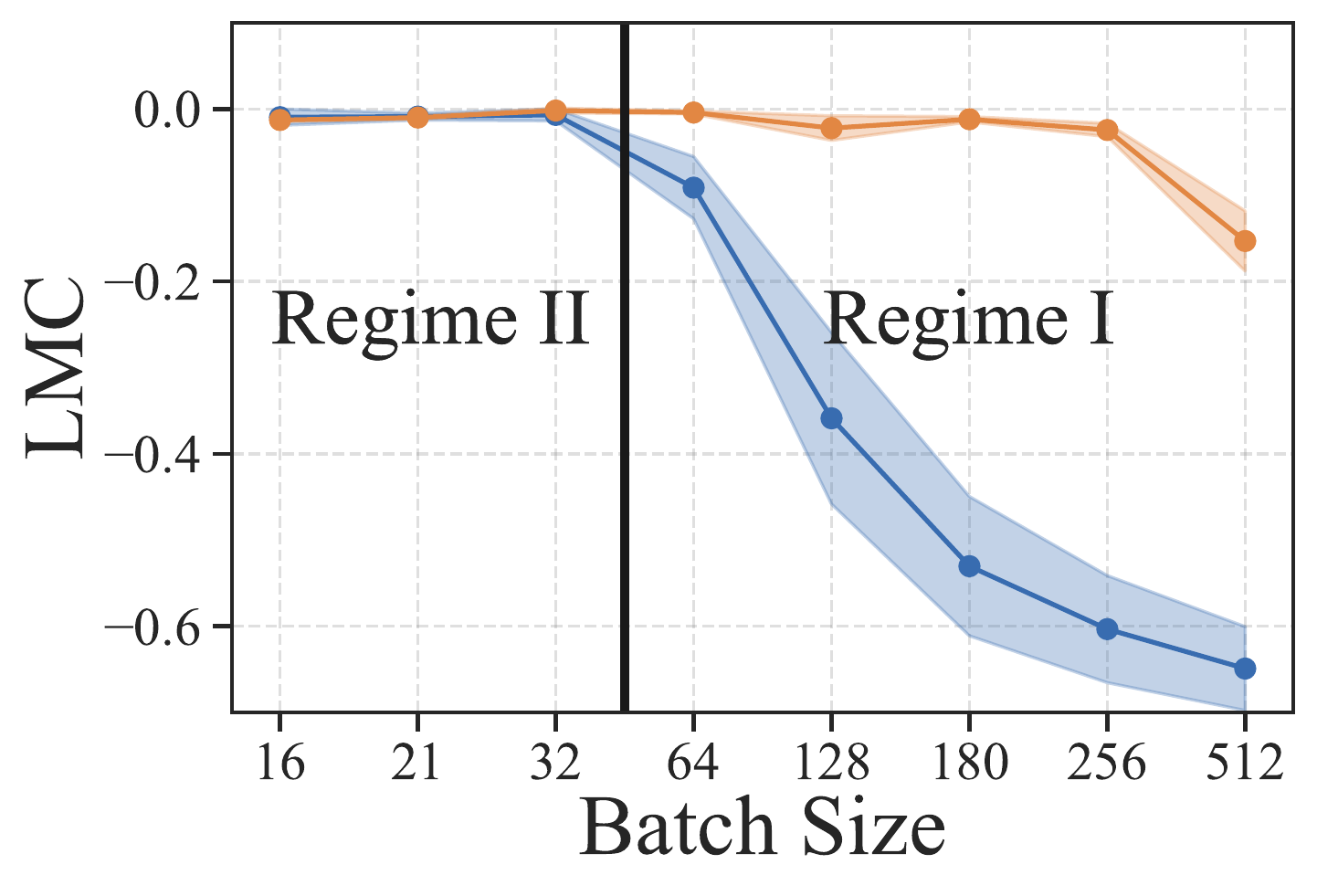} \\

    \multicolumn{2}{c}{\includegraphics[width=0.5\columnwidth,keepaspectratio]{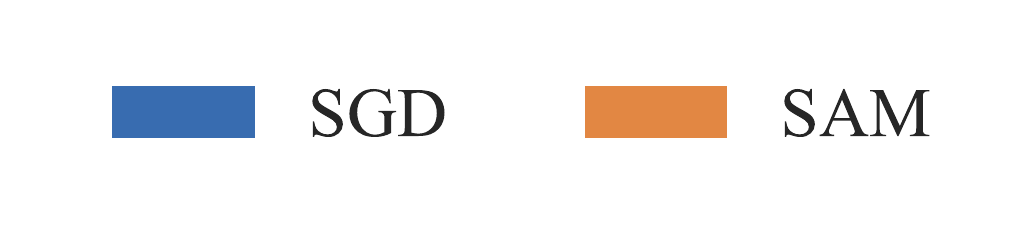}}

	\end{tabular} 
        \vspace{-6.5mm}
	\caption{Comparing final pruning performance with dense model training using SGD versus SAM in Regime I and Regime II. 
	}
	\label{fig:sam_sgd_base}
	\vspace{-5mm}
\end{centering}
\end{figure}

As explained in Section \ref{def:load-temp}, SAM can be viewed as a high-temperature training method, and the hyperparameter of SAM to control effective temperature is the neighborhood size ($\rho$). 
First, we find that the SAM has a dichotomous effect when applied to pruning, i.e., SAM can either improve or damage pruning depending on the regime to which the pruned model belongs.
This is explained by the proposed three-regime model. 
Then, we show that we can mitigate the negative effect of SAM by adaptively tuning the $\rho$ using our three-regime model.

\textbf{Experiment details.}
We study the effect of SAM on different regimes and vary the regimes of pruned models by varying temperature-like or load-like parameters.
In the first experiment, we vary the regimes between Regime I and II by changing the temperature, i.e., models are trained with a varying magnitude of temperature (batch size and training epochs) and pruned to a fixed density of 7\% using \textsc{UniformMP}. We set $\rho = 0.6$ for SAM.
In the second experiment, we vary the regimes between Regime I and II by changing the load, i.e., models are pruned to varying model densities and trained with a fixed magnitude of temperature. We also show a simple CKA-based method that can adaptively tune the $\rho$ of SAM for optimal pruning performance.

\textbf{Conventional wisdom.} Using SAM for dense model training is consistently beneficial for pruning performance~\citep{na-etal-2022-train}.

\textbf{Proposed method.} We find that the conventional wisdom is effective (SAM improves pruning) when the pruned model belongs to Regime I, while it damages pruning when the model belongs to Regime II and uses a large temperature ($\rho$), as this leads to an unfavorable transition from Regime II-B to a worse Regime II-A.
We propose a CKA-based hyperparameter tuning method to adaptively tune the $\rho$ of SAM according to the CKA similarity.
Specifically, for a given model density, we assess the CKA similarity of the pruned models, and we choose the $\rho$ that yields the highest CKA, thereby guiding the pruned models towards a more optimal Regime II-B.

\textbf{Results.}
The results of the first experiment are shown in Figure \ref{fig:sam_sgd_base}. We show that when the SGD-based pruning settings (blue lines) fall within Regime I (poor LMC and CKA), SAM (orange lines) significantly reduces the test error, as well as improving LMC and CKA.
This aligns with the conventional wisdom that SAM is effective in improving model compressibility, as well as our three-regime model that a large temperature is beneficial to pruning in Regime I.
However, in Regime II (near-zero LMC and benign CKA), SAM yields negligible improvements, which means conventional wisdom loses its effectiveness in this regime.

\begin{figure}[!thb]\centering

  \begin{subfigure}{0.44\columnwidth}
       \centering
       \includegraphics[width=\linewidth]{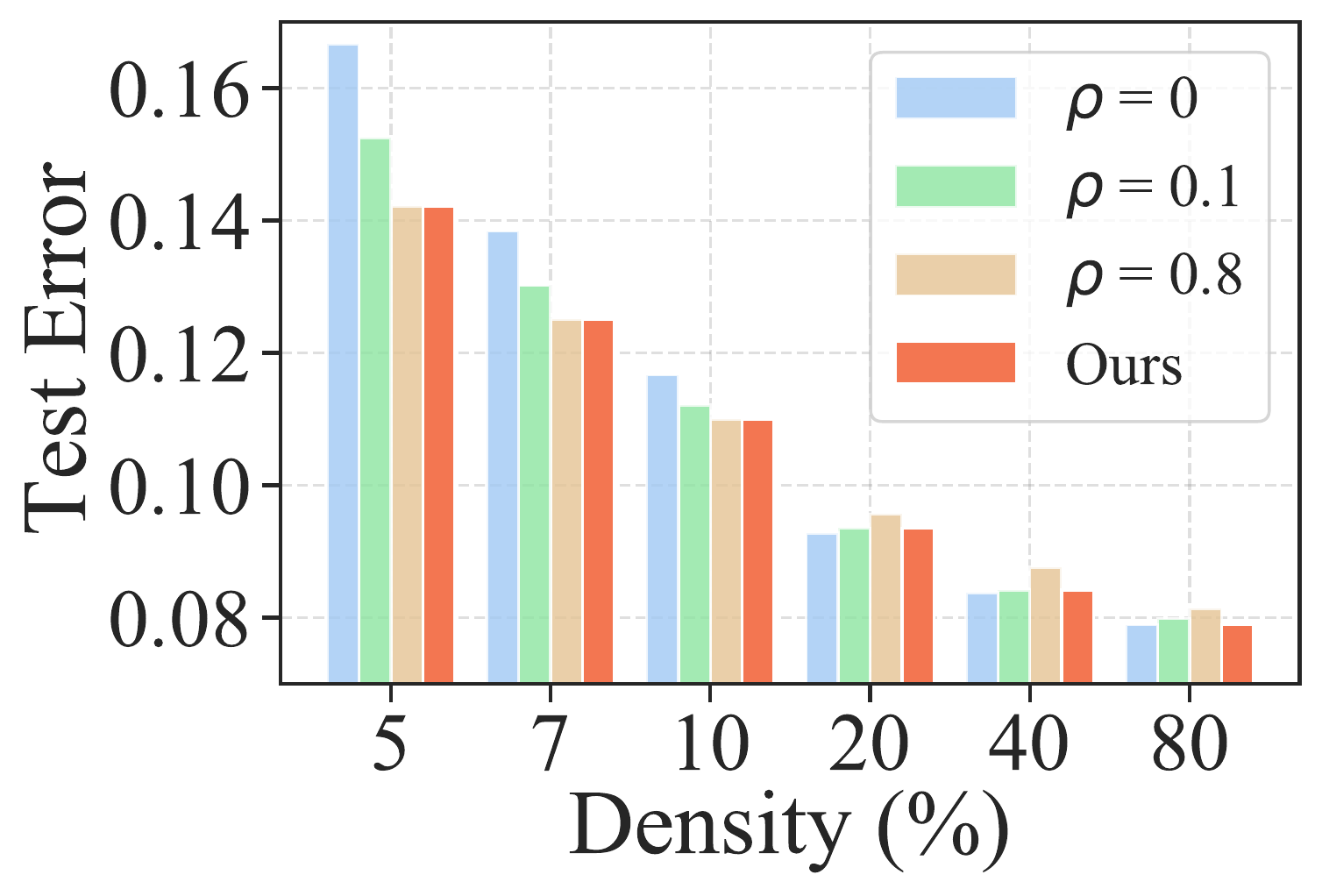}
       \caption{}
       \label{fig:rho-figure1}
       
   \end{subfigure}
  \begin{subfigure}{0.54\columnwidth}
       \centering
       \includegraphics[width=\linewidth]{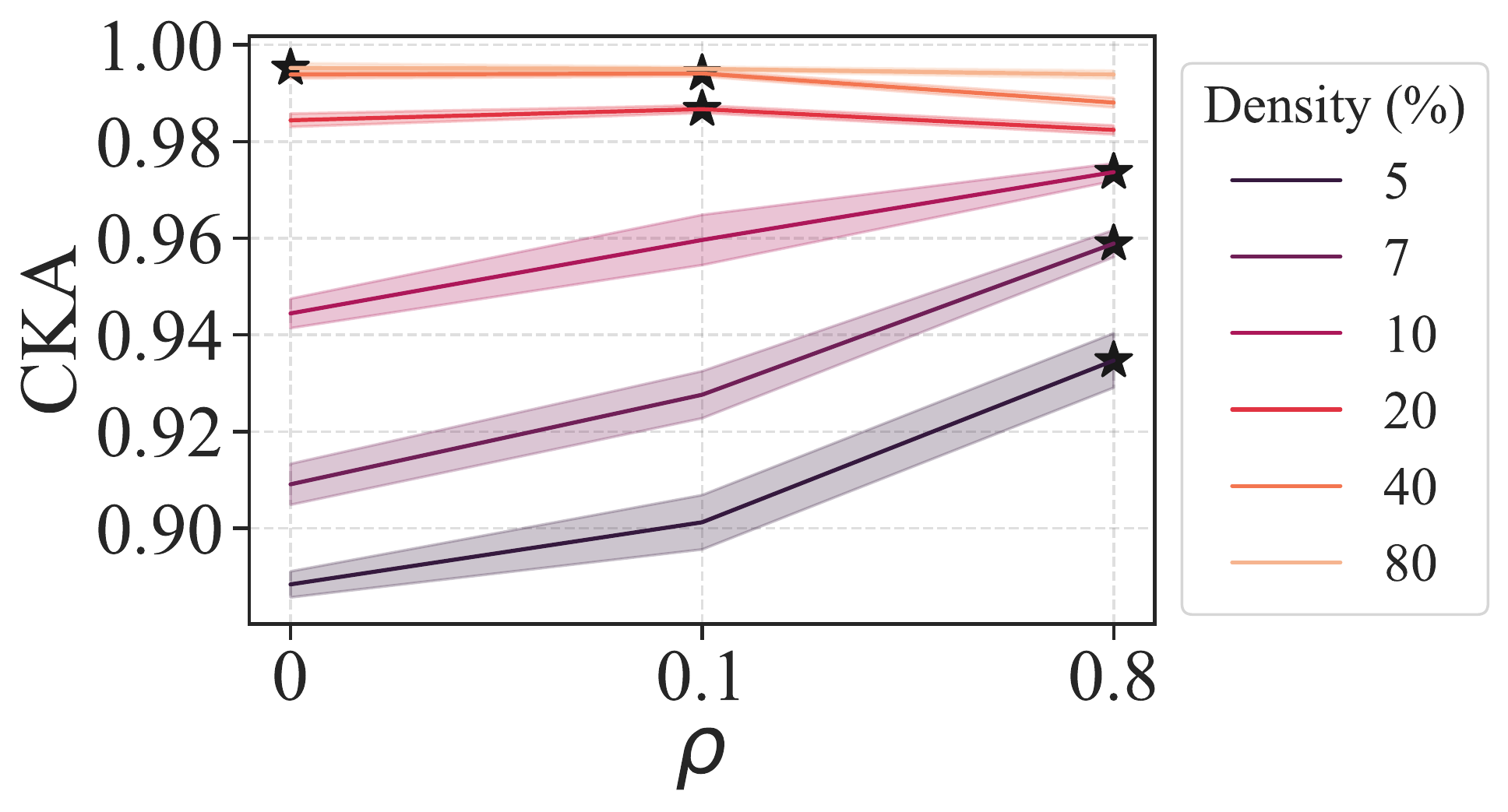}
       \caption{}
       \label{fig:rho-figure2}
   \end{subfigure}
   \vspace{-2mm}
    \caption{
    (a) Comparing the test error of small/large $\rho$ hyperparameter and proposed CKA-based hyperparameter tuning method. (b) The proposed hyperparameter tuning method selects the $\rho$ with the highest CKA value annotated by the black star markers.
    }
    \vspace{-1mm}
    \label{fig:sam-regime}
\end{figure}

The results of the second experiment are shown in Figure \ref{fig:sam-regime}. 
From Figure \ref{fig:rho-figure1}, we find that the low-density models benefit from using SAM with larger $\rho$, while high-density models suffer from using larger $\rho$: for densities 5\%, 7\%, and 10\%, compared with $\rho = 0 $ or $\rho = 0.1$, larger $\rho$ (0.8) significantly reduces the test error, while for densities 20\%, 40\%, and 80\%, larger $\rho$ makes the test error a bit worse. Figure \ref{fig:rho-figure2} demonstrates our CKA-based tuning method of selecting the $\rho$ under different model densities based on the CKA values, and the red bar in Figure \ref{fig:rho-figure1} shows that our approach always achieves optimal test error in tuning the $\rho$.

\vspace{-2mm}
\section{Conclusion}
\vspace{-1mm}
Based on recent work in the statistical mechanics of learning, we have proposed a three-regime model of network pruning, which depends on temperature-like and load-like parameters.
We discover a dichotomous phenomenon that arises when temperature-like and load-like parameters are varied: a higher temperature (used during the dense model training stage) results in good pruned model performance in heavily pruned networks, while a lower temperature (used during the dense model training stage) hurts final performance for lightly pruned networks. 
We find that popular metrics (such as the LMC and CKA similarity) closely track the transitions between regimes, providing operational ways of improving pruning. 
We anticipate the generalizability of our results and the reproducibility of the multi-regime taxonomy of load and temperature on other model compression techniques such as quantization and distillation.

\vspace{-2mm}
\paragraph{\textbf{Acknowledgments.}}
We would like to acknowledge the DOE, IARPA, NSF, and ONR 
%as well as a J. P. Morgan Chase Faculty Research Award 
for providing partial support of this work.


\newpage
\nocite{langley00}

\bibliography{example_paper}
\bibliographystyle{icml2023}

%%%%%%%%%%%%%%%%%%%%%%%%%%%%%%%%%%%%%%%%%%%%%%%%%%%%%%%%%%%%%%%%%%%%%%%%%%%%%%%
%%%%%%%%%%%%%%%%%%%%%%%%%%%%%%%%%%%%%%%%%%%%%%%%%%%%%%%%%%%%%%%%%%%%%%%%%%%%%%%
% APPENDIX
%%%%%%%%%%%%%%%%%%%%%%%%%%%%%%%%%%%%%%%%%%%%%%%%%%%%%%%%%%%%%%%%%%%%%%%%%%%%%%%
%%%%%%%%%%%%%%%%%%%%%%%%%%%%%%%%%%%%%%%%%%%%%%%%%%%%%%%%%%%%%%%%%%%%%%%%%%%%%%%
\newpage
\appendix
\onecolumn
%\section{You \emph{can} have an appendix here.}
\section{Related Work}
\label{sxn:related-work}

\subsection{Overview of related work}
\paragraph{Statistical mechanics of learning.}
Earlier work from the statistical mechanics of learning makes explicit connections between NN control parameters and load/temperature parameters, e.g., the temperature in the Boltzmann distribution~\cite{SST92,WRB93,DKST96,EB01_BOOK, BKPx20}; and more recent work has argued that these should empirically affect the output of training~\citep{martin2017rethinking}.
Based on this, \citet{martin2017rethinking} argues for adjusting load and temperature in mitigating overfitting, encouraging the model to exhibit Heavy-Tailed Self-Regularization (HT-SR), as its layer weight matrices develop stronger correlations over the course of training \cite{martin2018implicit_JMLRversion}.
The study of the correlation among HT-SR-related generalization metrics, model, task and test performance has been used to identify Simpsons paradoxes in deep learning contests \cite{MM21a_simpsons_TR}, which finds positive correlation between metrics when tasks are coupled and negative on certain task when performed alone. 
Importantly, our more phenomenological approach statistical mechanics view of NN learning is different from the commonly used (classical or more recent) works~\citep{SST92,WRB93,DKST96}.
However, it is consistent with the theoretical foundations of \citet{martin2017rethinking,martin2018implicit_JMLRversion}.
Our results provide additional evidence, complementing that of \citet{martin2020predicting_NatComm,yang2022evaluating}, that this more phenomenological approach to the statistical mechanics of learning can provide more principled basis for a practical theory of NN performance.
Also, recently we see a surge of interest in using statistical mechanics to analyze and improve learning, including~\citet{baity2018comparing} on the glassy behavior of neural networks,~\citet{barbier2019optimal} on optimal generalization error of generalized linear systems, and~\citet{sorscher2022beyond} on selecting easy versus hard samples used in training.

\paragraph{Load and temperature.}

The notation of load and temperature was introduced earlier in the statistical mechanics of learning, e.g., the load in the Hopfield model of associative memory~\citep{article, barra2008ergodic, barra2012equivalence} and the temperature in the Boltzmann distribution~\citep{SST92} and lsing model~\citep{Brush1967HistoryOT}.
As the load and temperature change, the learning system can change its performance and properties dramatically and qualitatively. 
Recent work studies the load and temperature in modern deep NN training. 
Several works characterize the temperature as the noise scale (or stochasticity) in the optimization process, specified as concrete hyperparameters that influence the training, e.g., the early stopping epochs~\citep{martin2017rethinking}, batch sizes, and learning rate~\citep{mccandlish2018empirical, yang2021taxonomizing}. 
\citet{martin2017rethinking} characterizes the load-like parameter as the effective capacity of the model or the ratio of the number of data points to a parameter characterizing the complexity of the model. 
This line of work also shows that varying load and temperature can make the NN training display phase behavior. 
There is no prior work that explicitly and systematically studies the load and temperature in the task of NN pruning and further observes and uses the phase transition behavior.
A few works have studied the effect of temperature-like parameters such as training epochs on network pruning.
\citet{li2020train} finds that larger models trained short of convergence outperform small models trained to convergence.
Although \citet{li2020train} gives a valuable heuristic on early stopping, the two hyperparameters training epochs and model size may be intertwined, and therefore, it is unclear whether early stopping is actually needed to reach the optimal pruning results. 
\citet{Shen_2022_CVPR} proposes a metric to indicate an early point to end the dense model training and begin pruning-retraining, showing that the model pruned at early epochs outperforms the one drawn at the later epoch.
\citet{liu2021sparse} shows that when pruning happens during the early training phase with large learning rates, models can easily recover performance via retraining. 
These studies provide evidence that varying temperature-like parameters in dense model training have a significant effect on pruning performance.
Via the VSDL model, our work provides a more explicit framework for these tasks.

\paragraph{Loss landscape.}
The study of loss landscapes has received attention among the machine learning community in recent years. 
Many of the important ideas have grounded in statistical mechanics and chemical physics~\cite{brooks2001, wales2003energy, stillinger2015energy, ballard2017energy}.
Recent work uses the loss landscape to analyze many modern techniques such as large-batch training~\cite{yao2020pyhessian} and pruning~\cite{frankle2020linear, evci2019the}.
\citet{yang2021taxonomizing} gives a systematic study on the local and global geometry properties of loss landscape in ``load-temperature'' framework. Their empirical finding shows that the global loss landscape metrics such as mode connectivity~\cite{garipov2018loss} and model similarity~\cite{kornblith2019similarity} perform well in indicating the phase transition of model training, and the transition is closely correlated with the generalization performance of trained models. 
\citet{frankle2020linear} uses linear mode connectivity to indicate the phase transition of drawing ``lottery tickets'' during dense model training.
Both works show that the global property of loss landscape is quite effective in explaining different phase transitions observed.

\subsection{Discussion on the difference between this work and \citet{yang2021taxonomizing}}

We discuss several aspects of our approach that differ from how \citet{yang2021taxonomizing} applied the VSDL model.

First, the astute reader will have noticed that, since the pruning pipeline is 
\vspace{-3mm}
\begin{center}
[Train-dense-model -- Prune -- Retrain-pruned-model], 
\end{center}
\vspace{-3mm}
there actually exists four variables of loads and temperatures that could potentially influence the pruning performance:   
1) load of dense-model;
2) temperature of dense-model;
3) load of pruned-model; and 
4) temperature of pruned-model.
Our main empirical finding is that, when one targets a given load of the pruned model, one should adjust the temperature of the dense model based on the phase behavior of the global loss landscape of pruned models. 
That is, in this special multiple-stage pipeline, where a large perturbation (pruning) is involved in the middle stage, changing the temperature in the first stage will significantly influence (and can be used to control) the global structure in the final stage.
Our modeling results demonstrate that the ``load-temperature'' framework of the VSDL model can simplify and help solve a complex multi-stage problem that is commonly seen in applications.
Other possible examples of this include transfer learning, distillation, quantization, etc. 

Second, the astute reader may also wonder what is the role of Hessian information, given its importance in how \citet{yang2021taxonomizing} taxonomized loss landscapes.
In \citet{yang2021taxonomizing}, Hessian information was used to identify the sharp transition at the interpolation threshold.
In our study of pruning, however, most of the pruned models are unable to interpolate the training data. 
Compared with Figure 3 in \citet{yang2021taxonomizing}, we observed empirically that the Hessian eigenvalues are too large to reach locally flat phases.
That is, we found that using metrics from \citet{yang2021taxonomizing} that were used to measure the \emph{local structure} of loss landscapes (such as Hessian trace or leading Hessian eigenvalues) is not as informative as the global LMC and CKA metrics in the pruning setting, and it does not significantly help identify the different regimes for the pruning problem. 

Finally, we would like to emphasize the novelties of our work in comparison to \citet{yang2021taxonomizing}: 1) While \citet{yang2021taxonomizing} builds a theory-driven framework, our paper proposes practical machine learning algorithms, including hyperparameter tuning and model selection, to address the specific challenges in pruning. 2) While \citet{yang2021taxonomizing} primarily focuses on taxonomizing various types of phase transitions, we leverage these phase transitions to design effective pruning algorithms with direct implications for practical applications.

\section{Additional Details for Three-regime Taxonomy}
\label{sxn:corrob}

\subsection{Implementation details} \label{sec:imple-detail-taxonomy}

\textbf{Datasets.}
For the image classification task, we consider CIFAR-10, CIFAR-100~\citep{krizhevsky2009cifar}, and SVHN~\citep{sermanet2011traffic}. 
CIFAR-10 comprises 50,000 training images and 10,000 testing images with 10 categories. 
CIFAR-100 comprises 50,000 training images and 10,000 testing images with 100 categories. 
SVHN comprises 73,257 training images and 26,032 testing images with 10 categories.
For the machine translation task, we consider WMT14~\citep{bojar-etal-2014-findings} German to English (DE-EN) dataset. We subsample 1.28M sentence pairs from the WMT14 training set for training and report the BLEU score on the validation set.

\textbf{Model architectures.}
For the image classification task, we use PreResNet-20~(Pre-Activation ResNet, \citep{he2016identity}), DenseNet-40~\citep{huang2017densely}, and VGG-19~\citep{simonyan2014very}. Our implementation is based on \citet{liu2018rethinking}.
For the machine transition task, we use Transformer-base~\cite{vaswani2017attention}.

\textbf{Training procedures.}
For full model training and pruned model retraining, we use SGD as the default optimizer. The default hyperparameters include a momentum of 0.9, weight decay of 1e-4, and a training duration of 160 epochs.
Learning rate decay is applied with an initial learning rate of 0.1, which decreases by a factor of 10 at epochs 80 and 120.
In the study of varying training epochs as temperature, we fix batch size as 64. In the study of varying batch sizes as temperature, we fix the training iterations as 62400. 
For each configuration, we train the dense model with three different random seeds and retrain the pruned sparse model with three random seeds as well. We report the average test error and loss landscape measures across these random seeds.

\textbf{Pruning procedures.}
We implement the unstructured magnitude pruning methods, \textsc{UniformMP} and \textsc{GlobalMP}, following \citet{liu2018rethinking, lee2021layeradaptive}. For image classification models, we only prune weights in convolution layers, as suggested by \citet{liu2018rethinking}, while keeping the last linear layer intact due to its relatively small number of parameters.

\textbf{Hyperparameters for different metrics.}
We implement the LMC and CKA similarity metrics based on \citet{yang2021taxonomizing}. For LMC measurement, we use the entire training set, while for CKA similarity measurement, we draw 6,400 samples from the training set.

\subsection{Additional Supporting Results for Three-regime Taxonomy} \label{sec:corrob-taxonomy}
\begin{figure*}[!th]\centering
  \begin{subfigure}{0.24\linewidth}
       \centering
       \includegraphics[width=\linewidth]{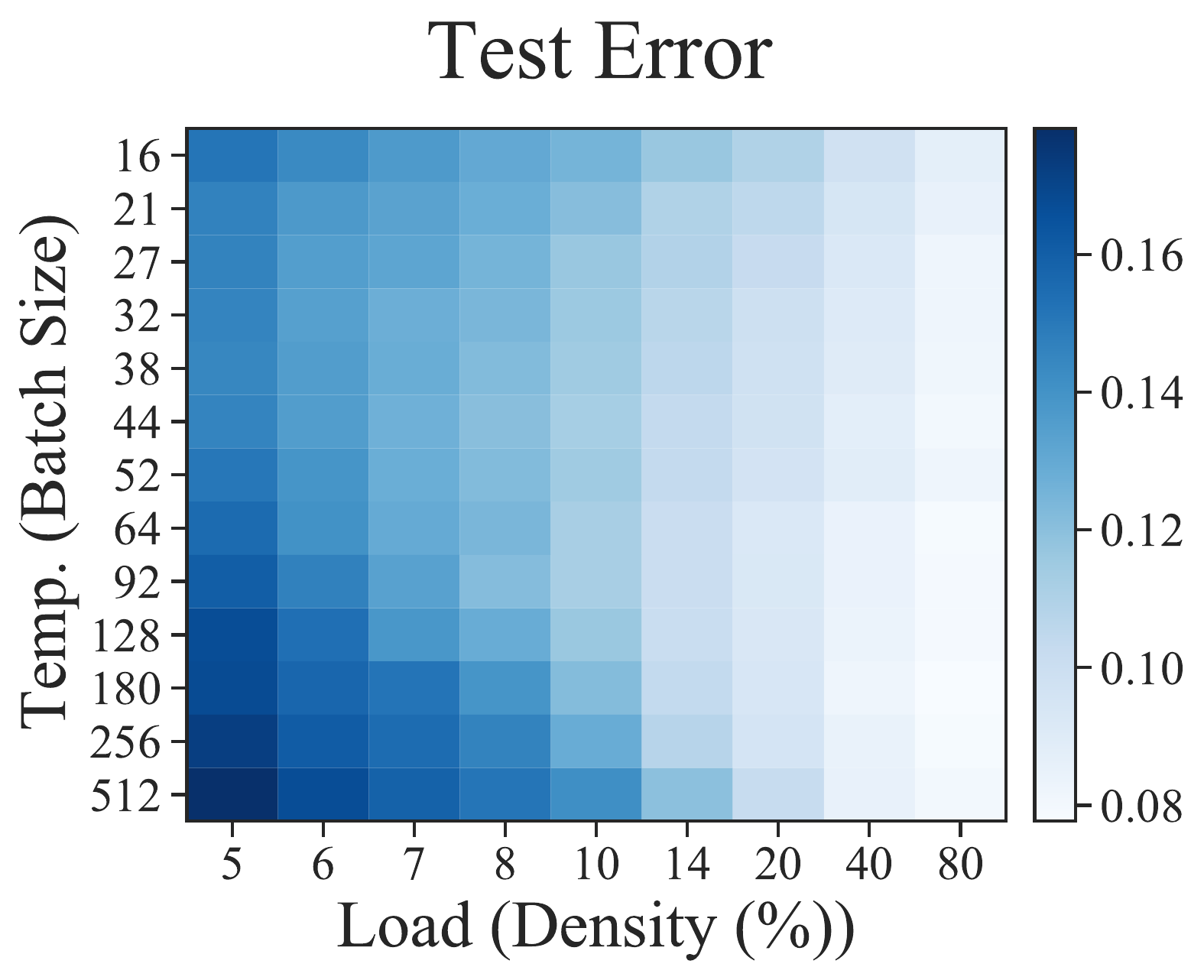}
       \caption{Test error}
       \label{fig:batchsize-error}
   \end{subfigure}
    \begin{subfigure}{0.24\linewidth}
       \centering
       \includegraphics[width=\linewidth]{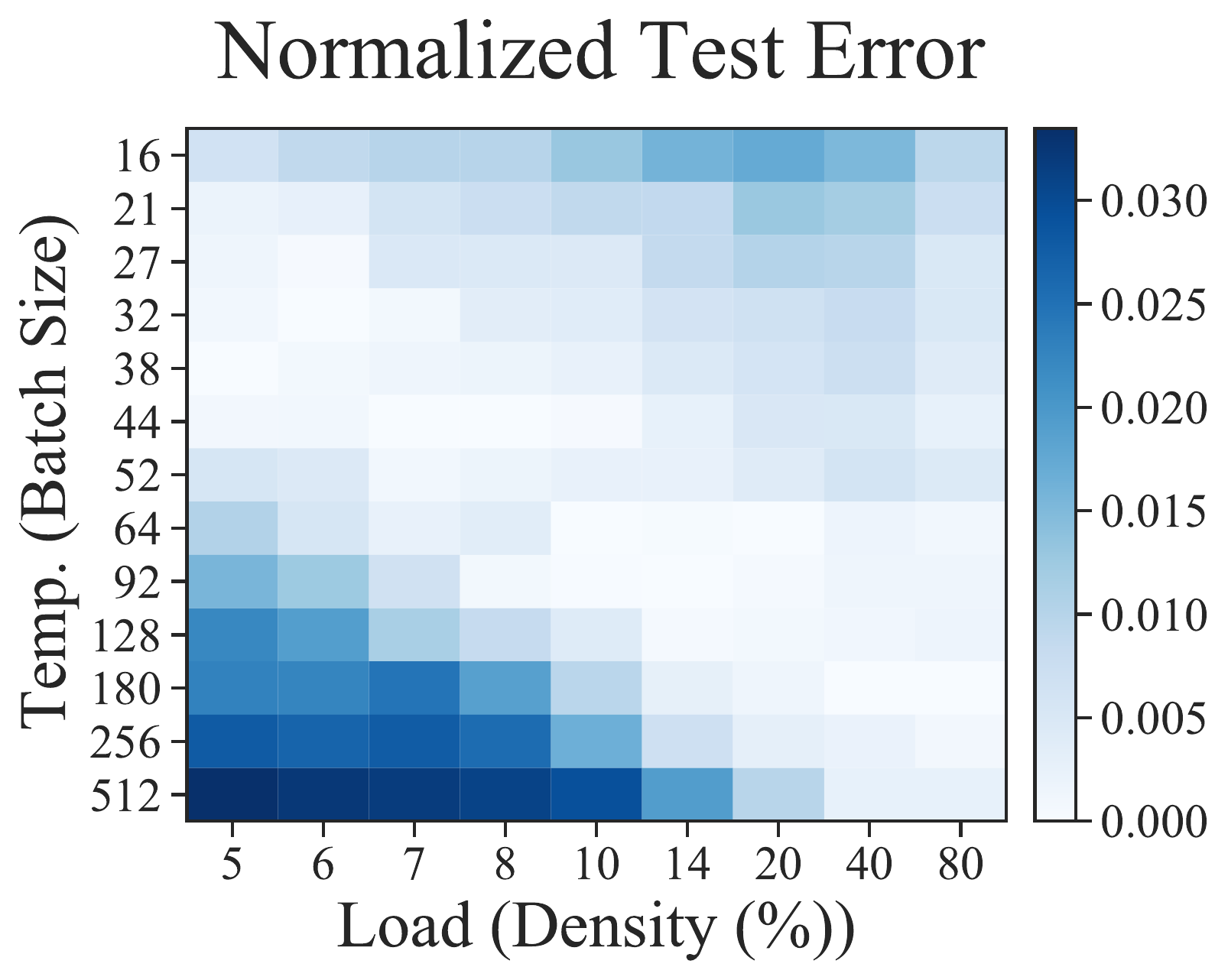}
       \caption{Normalized test error}
       \label{fig:batchsize-norm-error}
   \end{subfigure} 
   \begin{subfigure}{0.24\linewidth}
       \centering
       \includegraphics[width=\linewidth]{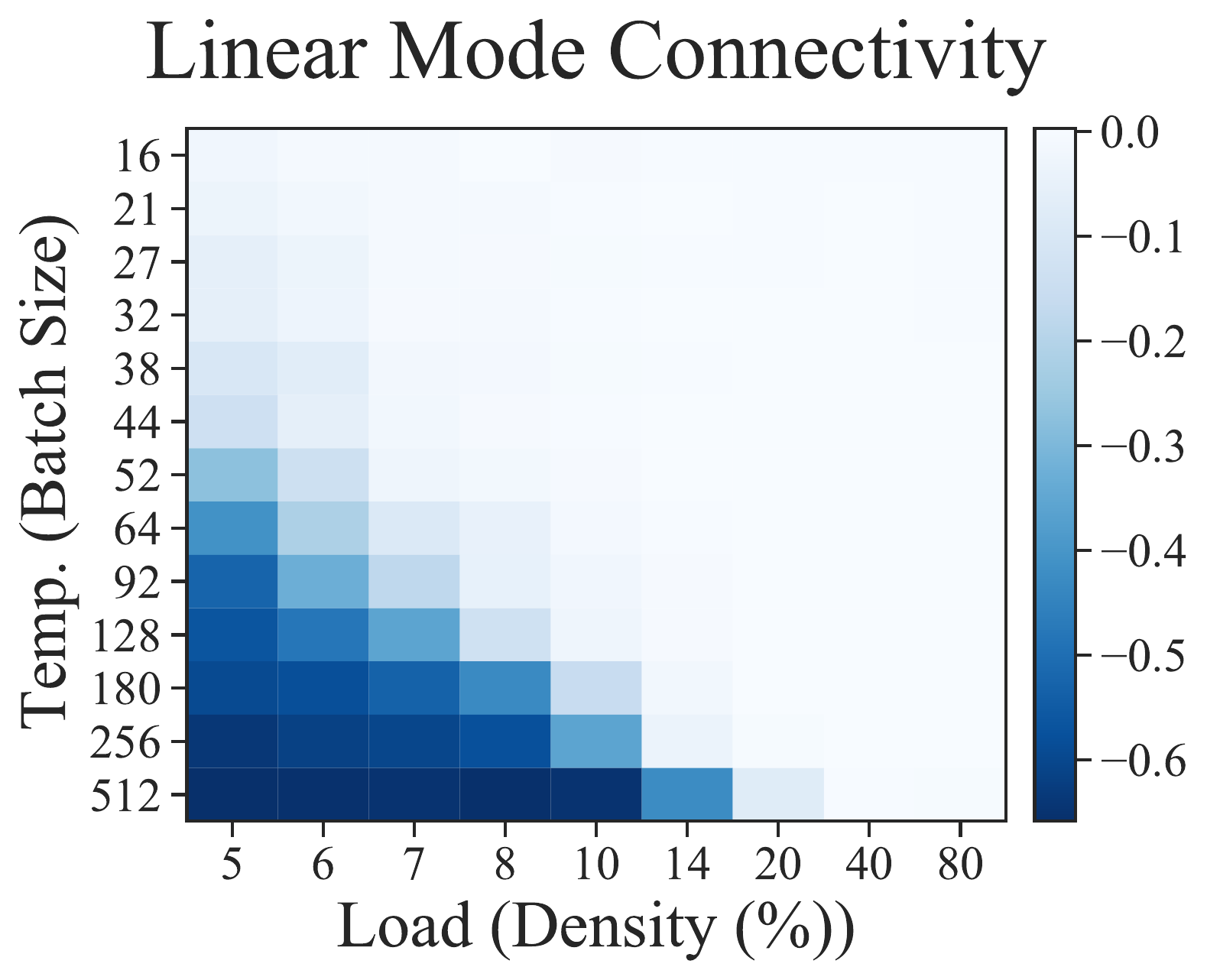}
       \caption{LMC}
       \label{fig:batchsize-lmc}
   \end{subfigure}
   \begin{subfigure}{0.24\linewidth}
       \centering
       \includegraphics[width=\linewidth]{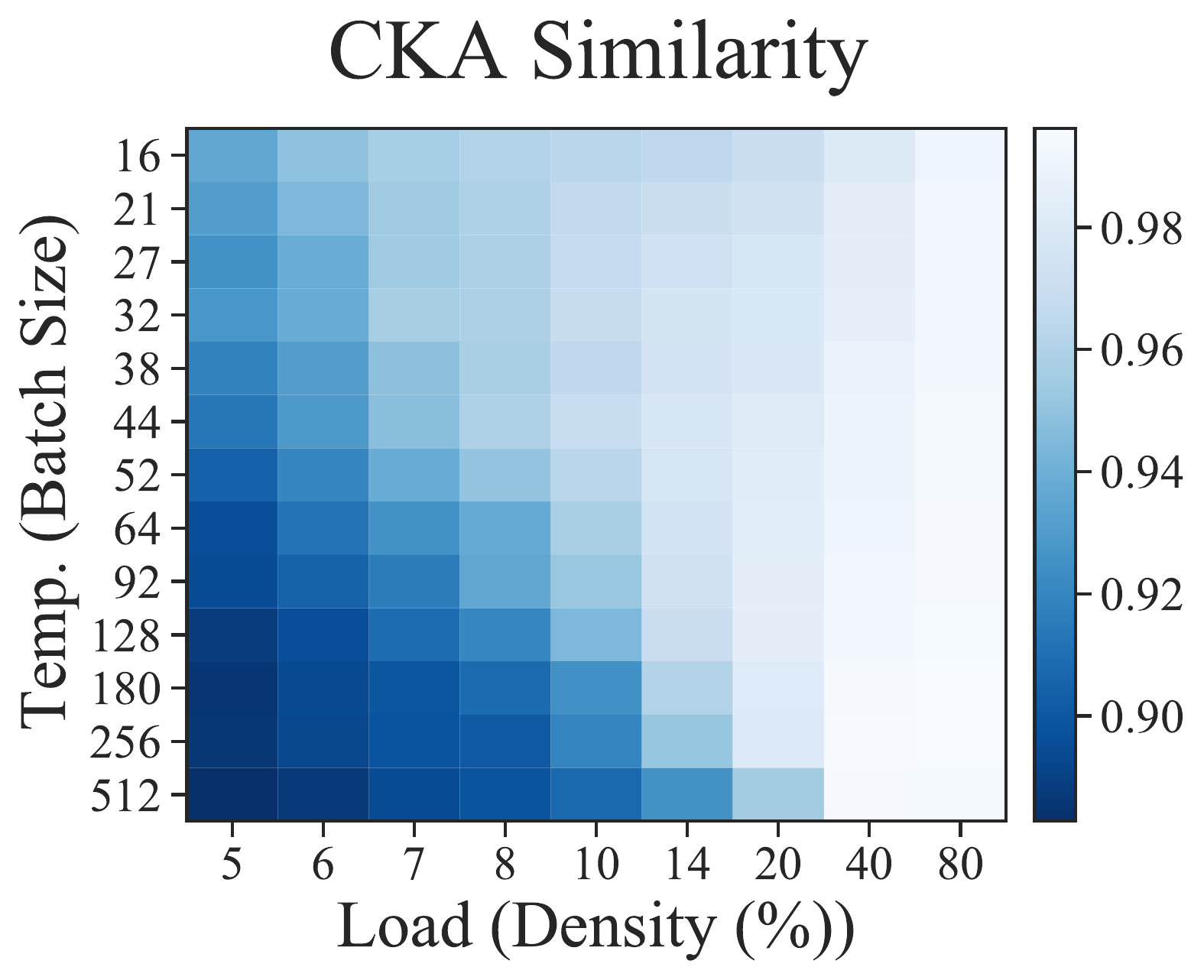}
       \caption{CKA similarity}
       \label{fig:batchsize-cka}
   \end{subfigure}
    \vspace{-2mm}
    \caption{\textbf{(Batch size as temperature).} Partitioning the 2D diagram of model density -- batch size into three regimes. Models are trained with PreResNet-20 on CIFAR-10.
    }
    \label{fig:batchsize-phase}
    
\end{figure*}

\textbf{Different temperature parameters.} 
We corroborate our main claim by studying batch size as an alternative temperature parameter, in addition to adjusting the number of training epochs. The results are shown in Figure \ref{fig:batchsize-phase}. Comparing Figure \ref{fig:batchsize-phase} with Figure \ref{fig:earlystop-regime}, we see that the three regimes are still present and the observation that high temperature improves pruning for poorly-connected Regime I while hurts pruning for well-connected Regime II-B also holds. Therefore, our central claim remains consistent even with the introduction of different temperature-like parameters.

\vspace{-1mm}

\begin{figure*}[!htb]\centering
  \begin{subfigure}{0.24\linewidth}
       \centering
       \includegraphics[width=\linewidth]{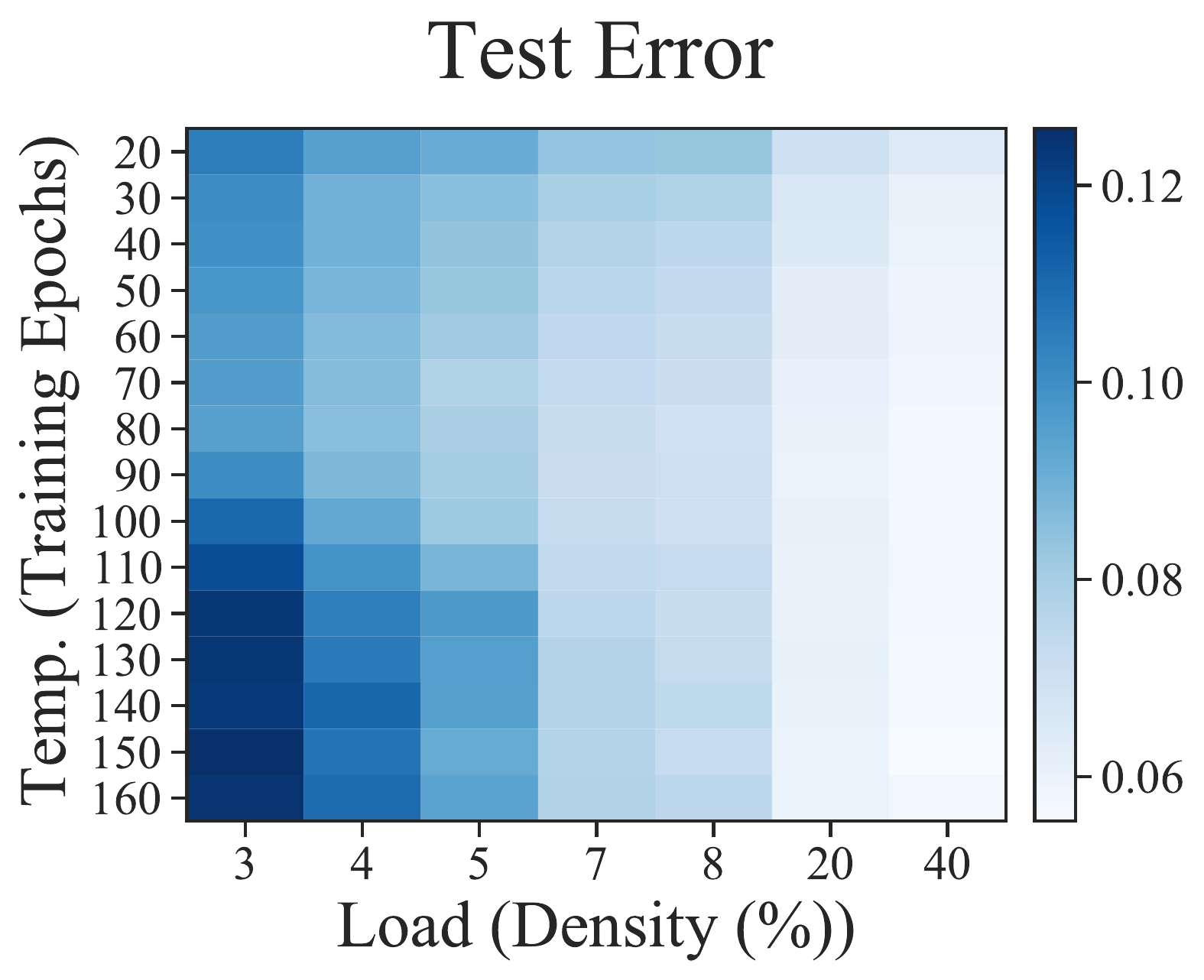}
       \caption{Test Error}
       \label{}
   \end{subfigure}
    \begin{subfigure}{0.24\linewidth}
       \centering
       \includegraphics[width=\linewidth]{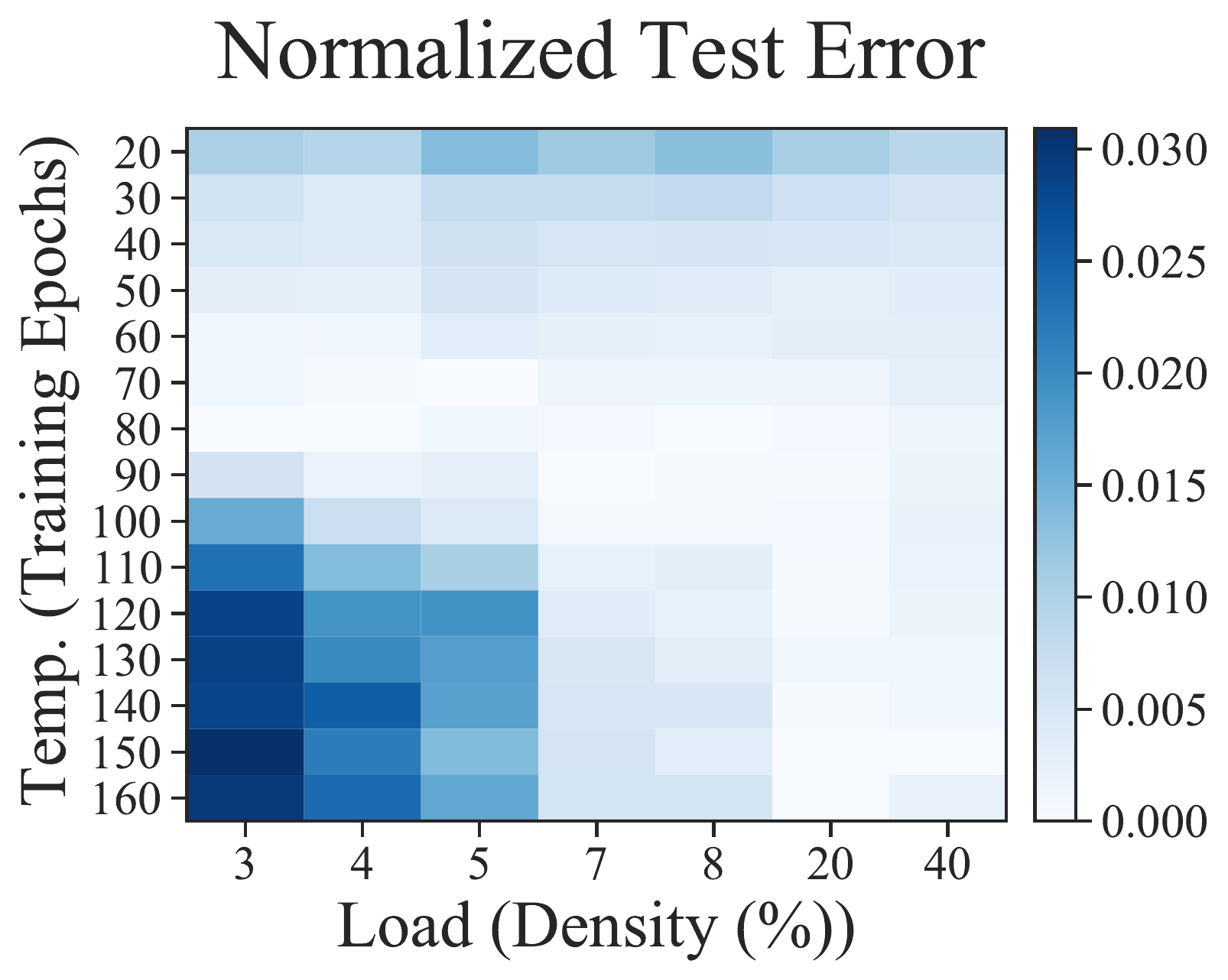}
       \caption{Normalized test error}
       \label{}
   \end{subfigure}
   \begin{subfigure}{0.24\linewidth}
       \centering
       \includegraphics[width=\linewidth]{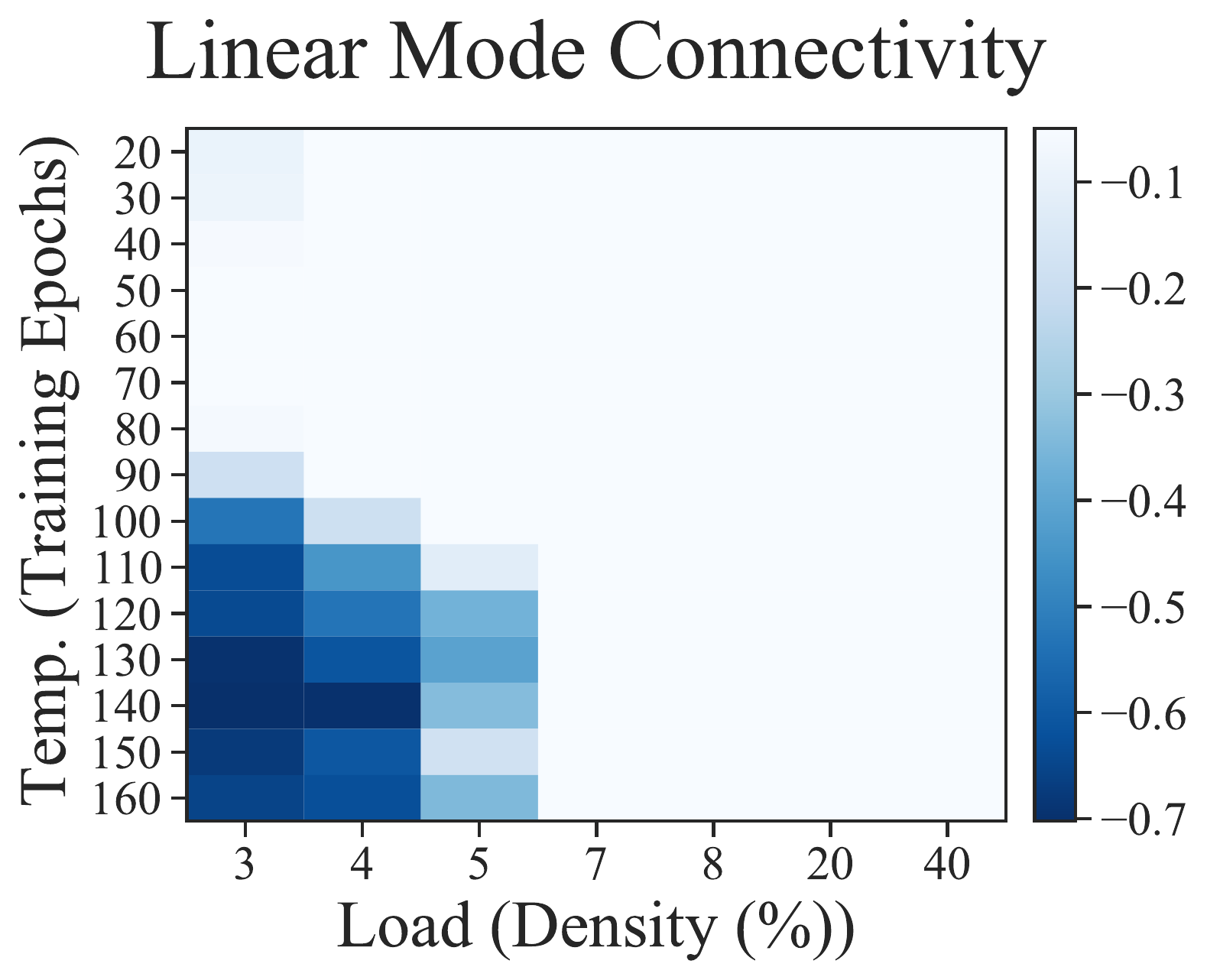}
       \caption{LMC}
       \label{}
   \end{subfigure}
   \begin{subfigure}{0.24\linewidth}
       \centering
       \includegraphics[width=\linewidth]{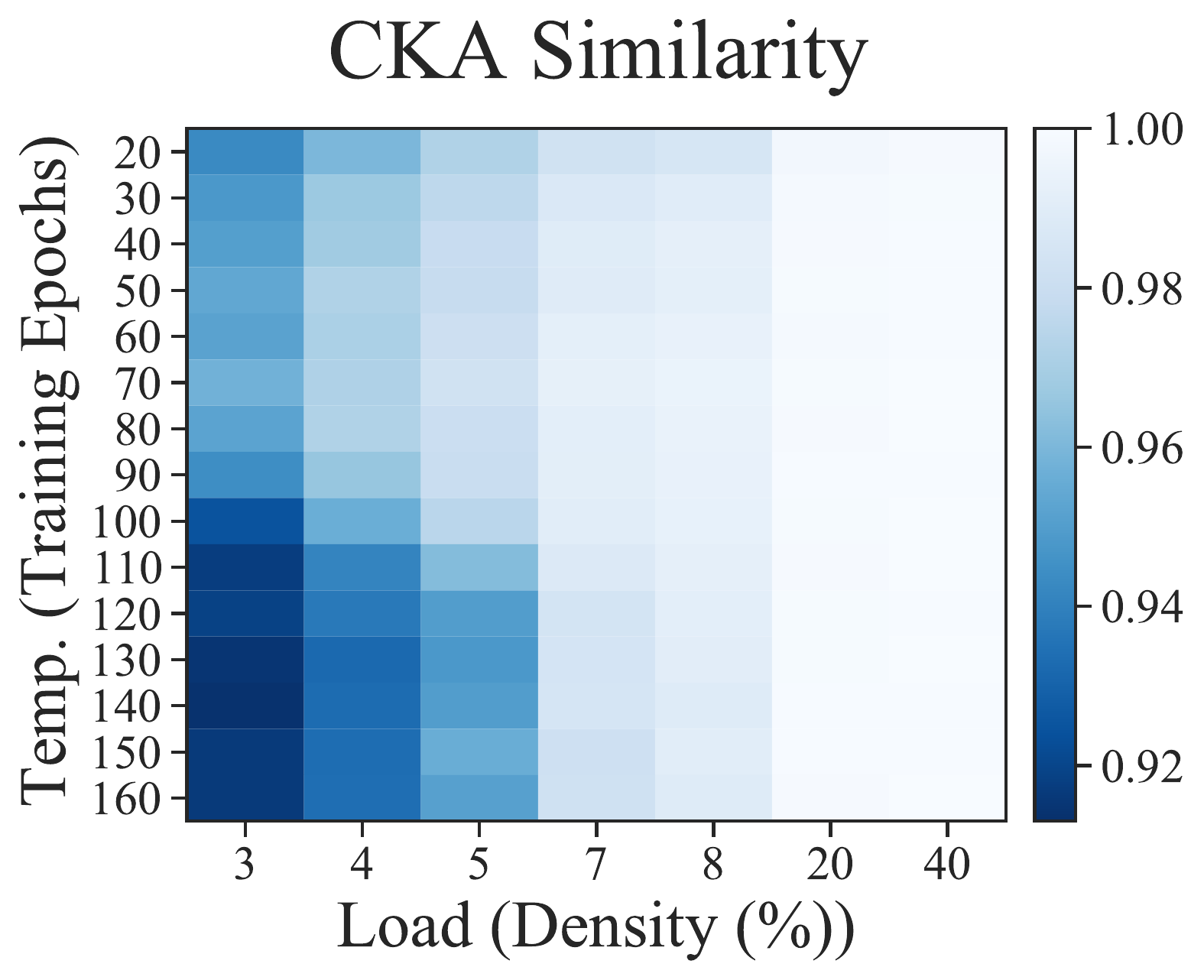}
       \caption{CKA similarity}
       \label{}
   \end{subfigure}
   \vspace{-2mm}
   \caption{\textbf{(DenseNet-40).} Partitioning the 2D model density -- training epochs diagram into three regimes. Models are trained with DenseNet-40 on CIFAR-10.}
   \label{fig:densenet40-phase}
\end{figure*}

\begin{figure*}[!htb]\centering
  \begin{subfigure}{0.24\linewidth}
       \centering
       \includegraphics[width=\linewidth]{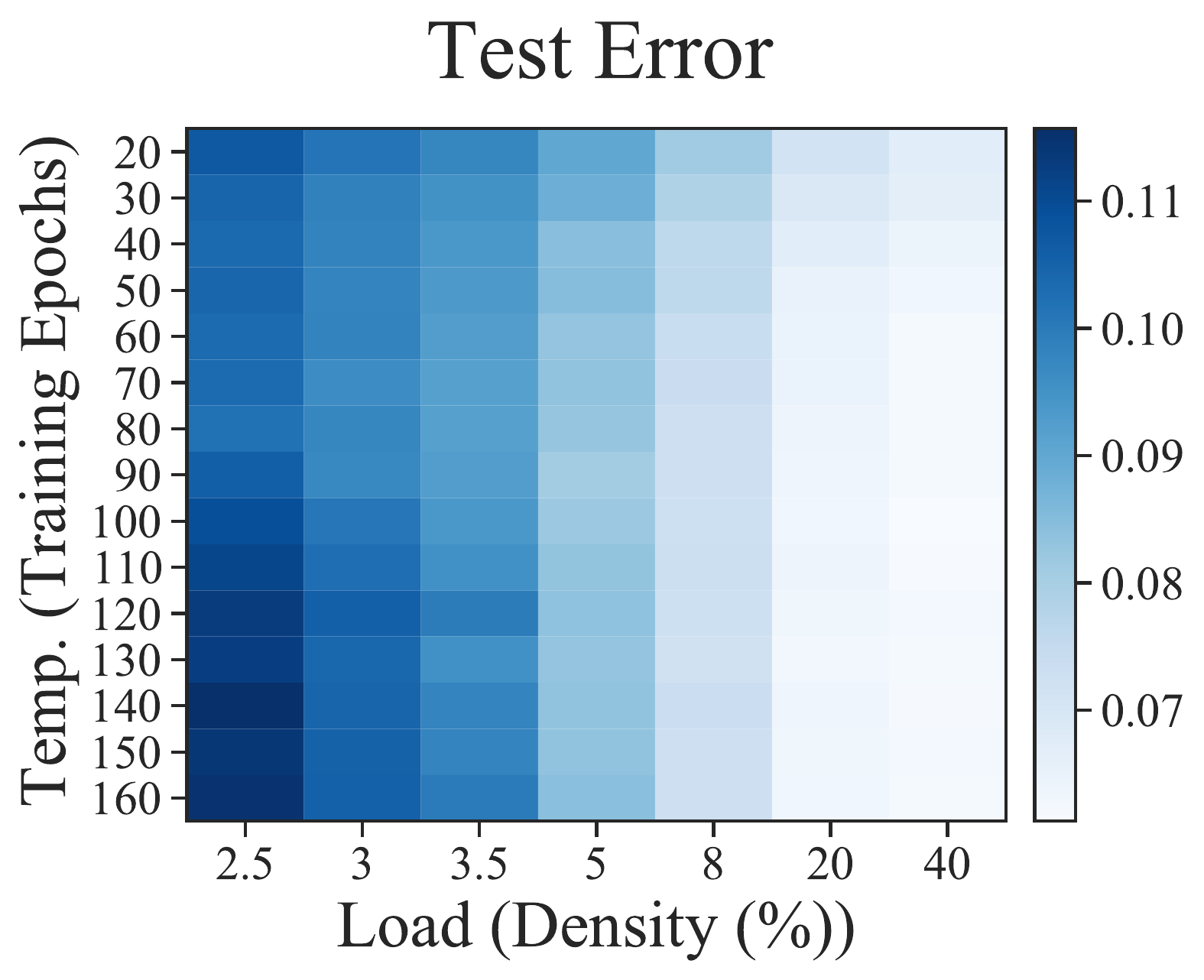}
       \caption{Test error}
       \label{}
   \end{subfigure}
    \begin{subfigure}{0.24\linewidth}
       \centering
       \includegraphics[width=\linewidth]{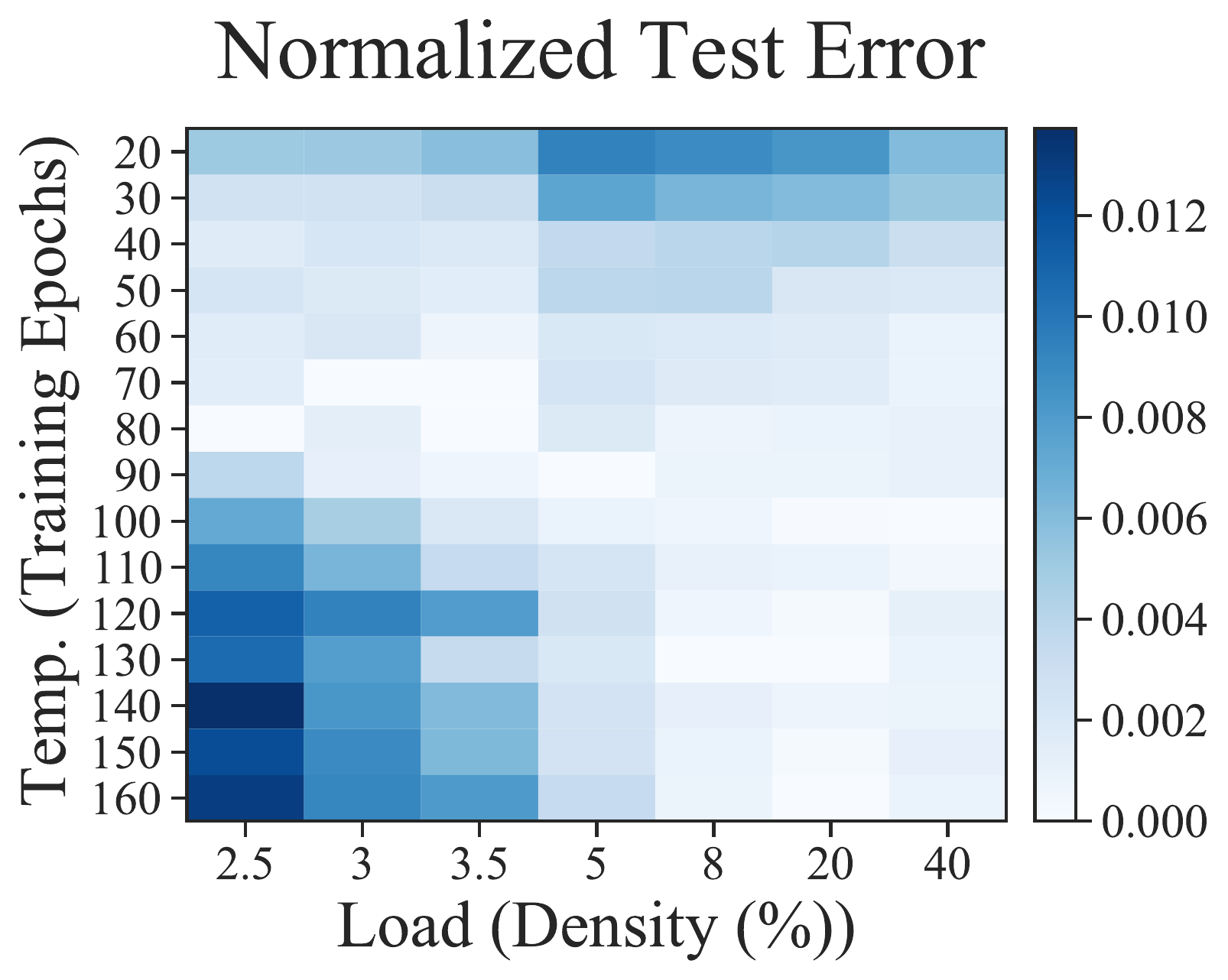}
       \caption{Normalized test error}
       \label{}
   \end{subfigure}
   \begin{subfigure}{0.24\linewidth}
       \centering
       \includegraphics[width=\linewidth]{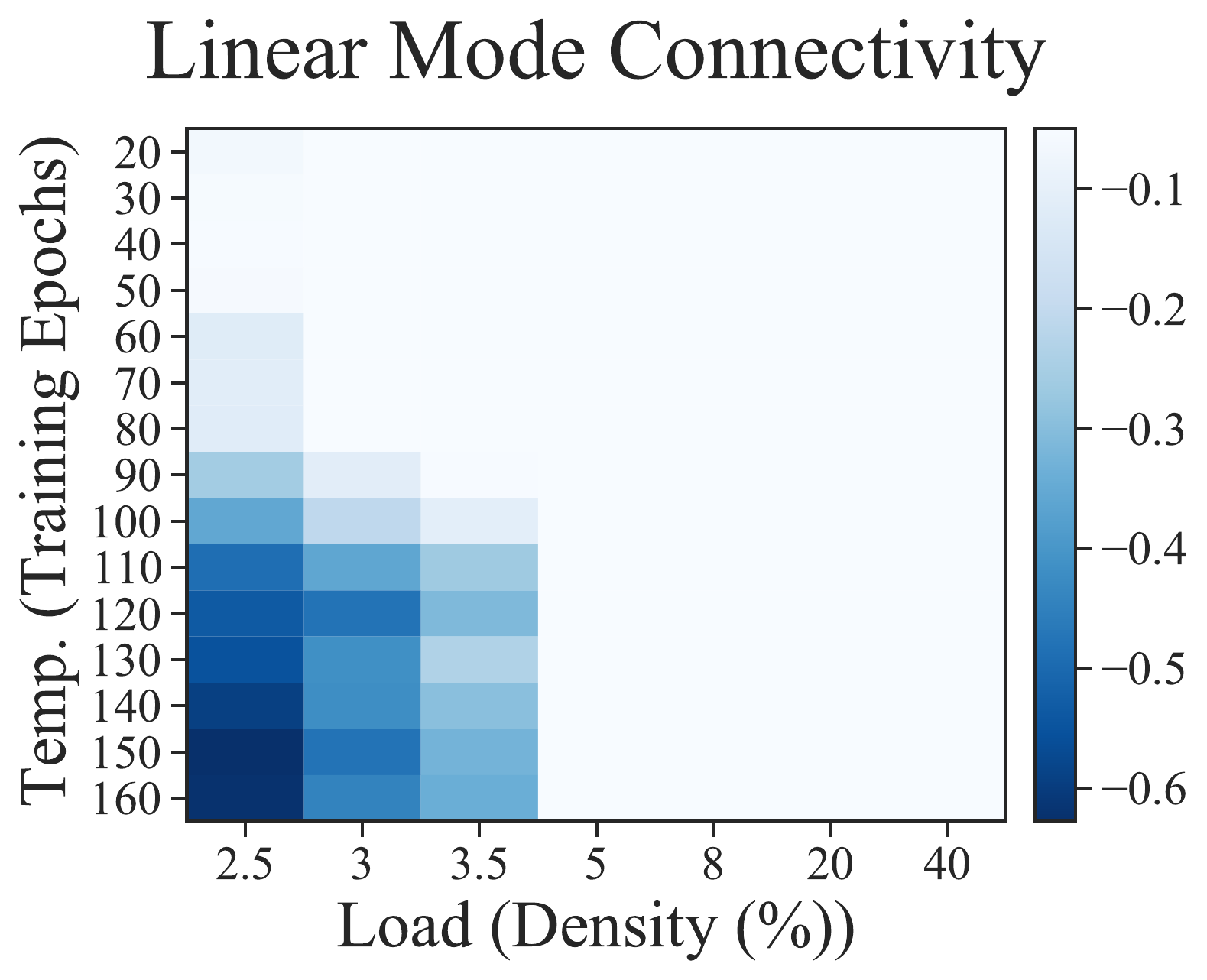}
       \caption{LMC}
       \label{}
   \end{subfigure}
   \begin{subfigure}{0.24\linewidth}
       \centering
       \includegraphics[width=\linewidth]{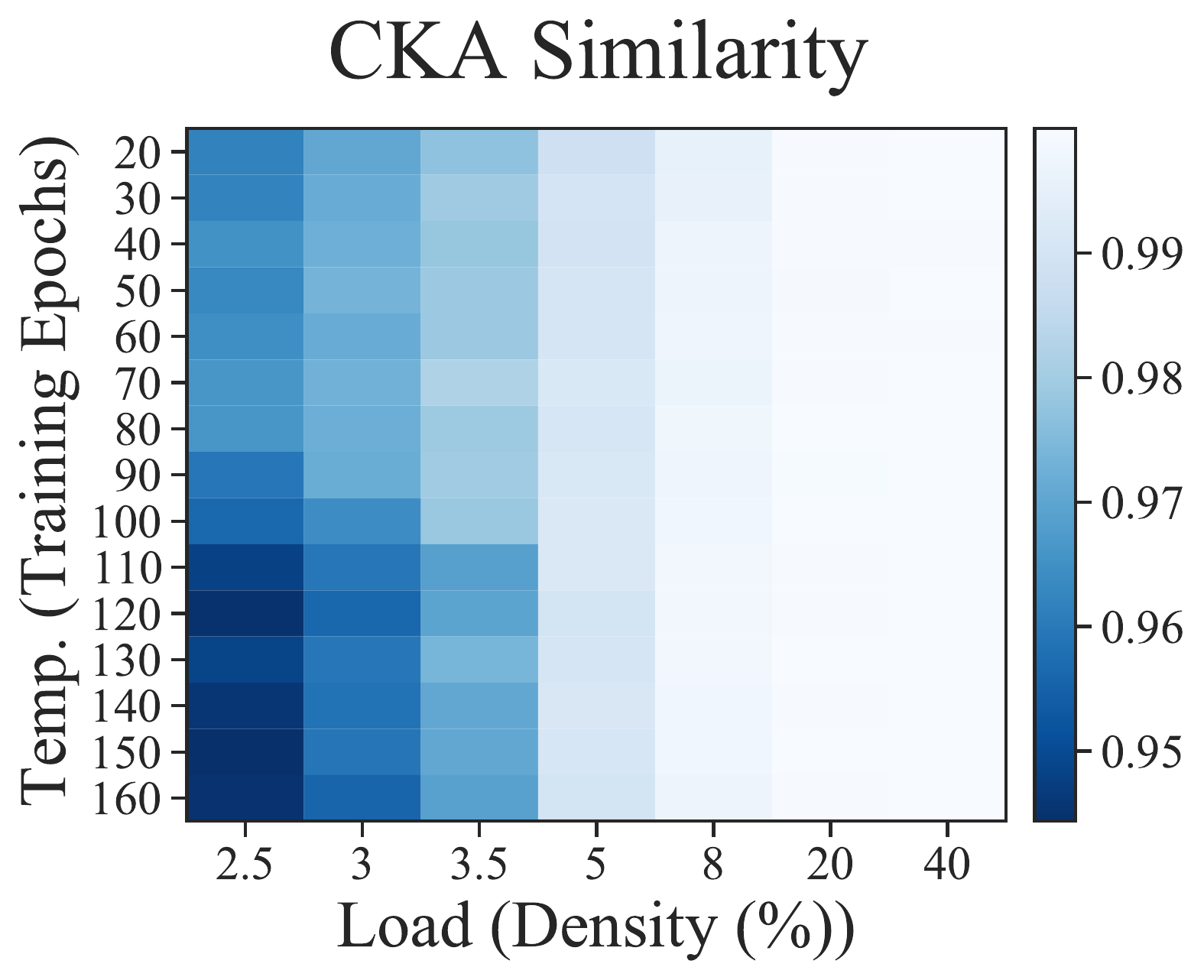}
       \caption{CKA similarity}
       \label{}
   \end{subfigure}
   \vspace{-2mm}
   \caption{\textbf{(VGG-19).} Partitioning the 2D model density -- training epochs diagram into three regimes. Models are trained with VGG-19 on CIFAR-10.}
   \label{fig:vgg19-phase}
\end{figure*}

\textbf{Different architectures.}
We extend the experiments with DenseNet-40 and VGG-19 on CIFAR-10 in Figure \ref{fig:densenet40-phase} and Figure \ref{fig:vgg19-phase}. The experiment produces analogous results to Figure \ref{fig:earlystop-regime} and supports our main claim. We note that DenseNet-40 and VGG-19 are larger architectures with more parameters (1M and 20M parameters) than the basic architecture PreResNet-20 (0.27M parameters), so we are able to prune it to a smaller density.
\vspace{-2mm}

\begin{figure*}[!th]\centering
  \begin{subfigure}{0.24\linewidth}
       \centering
       \includegraphics[width=\linewidth]{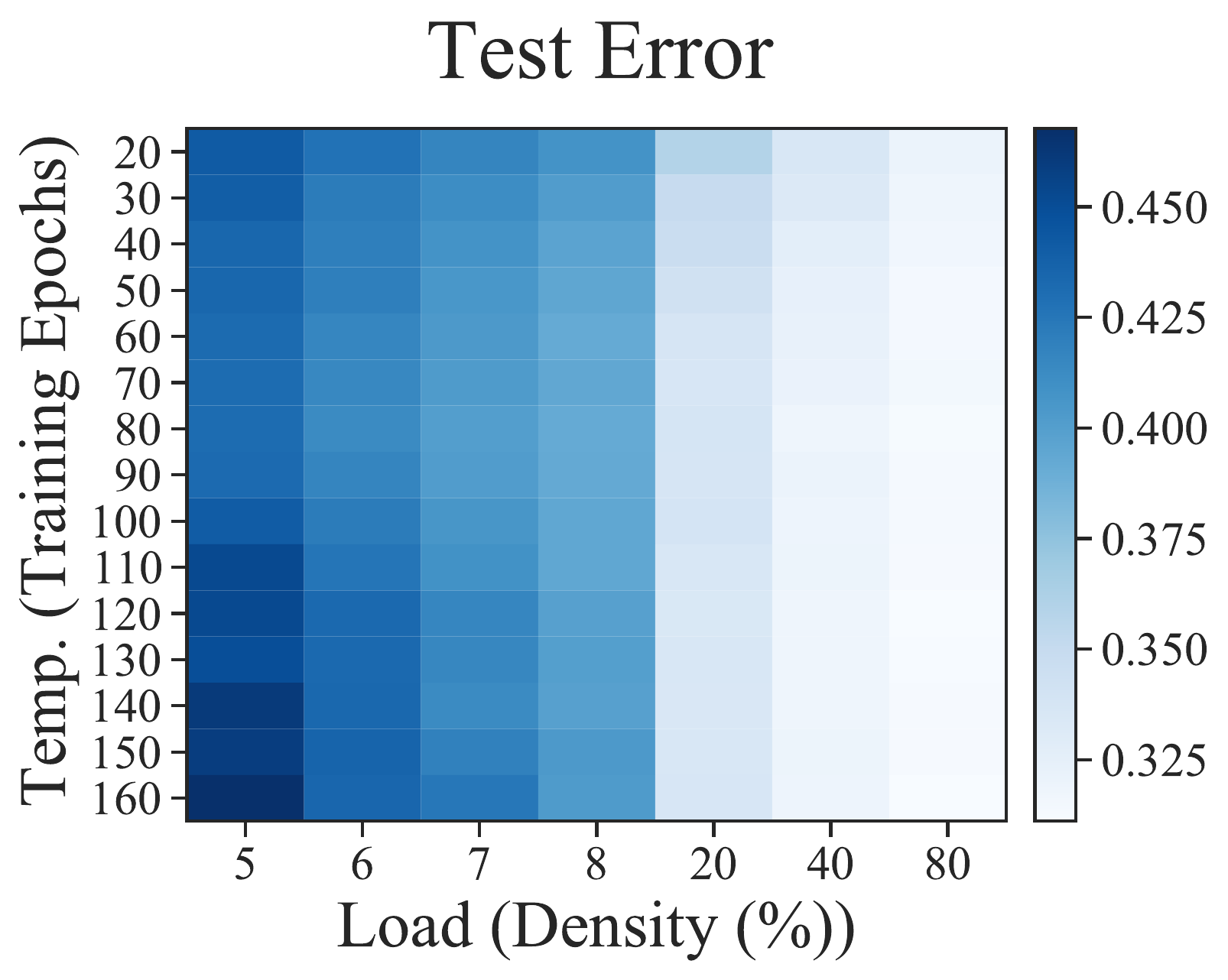}
       \caption{Test error}
       
       \label{fig:cifar100-error}
   \end{subfigure}
    \begin{subfigure}{0.24\linewidth}
       \centering
       \includegraphics[width=\linewidth]{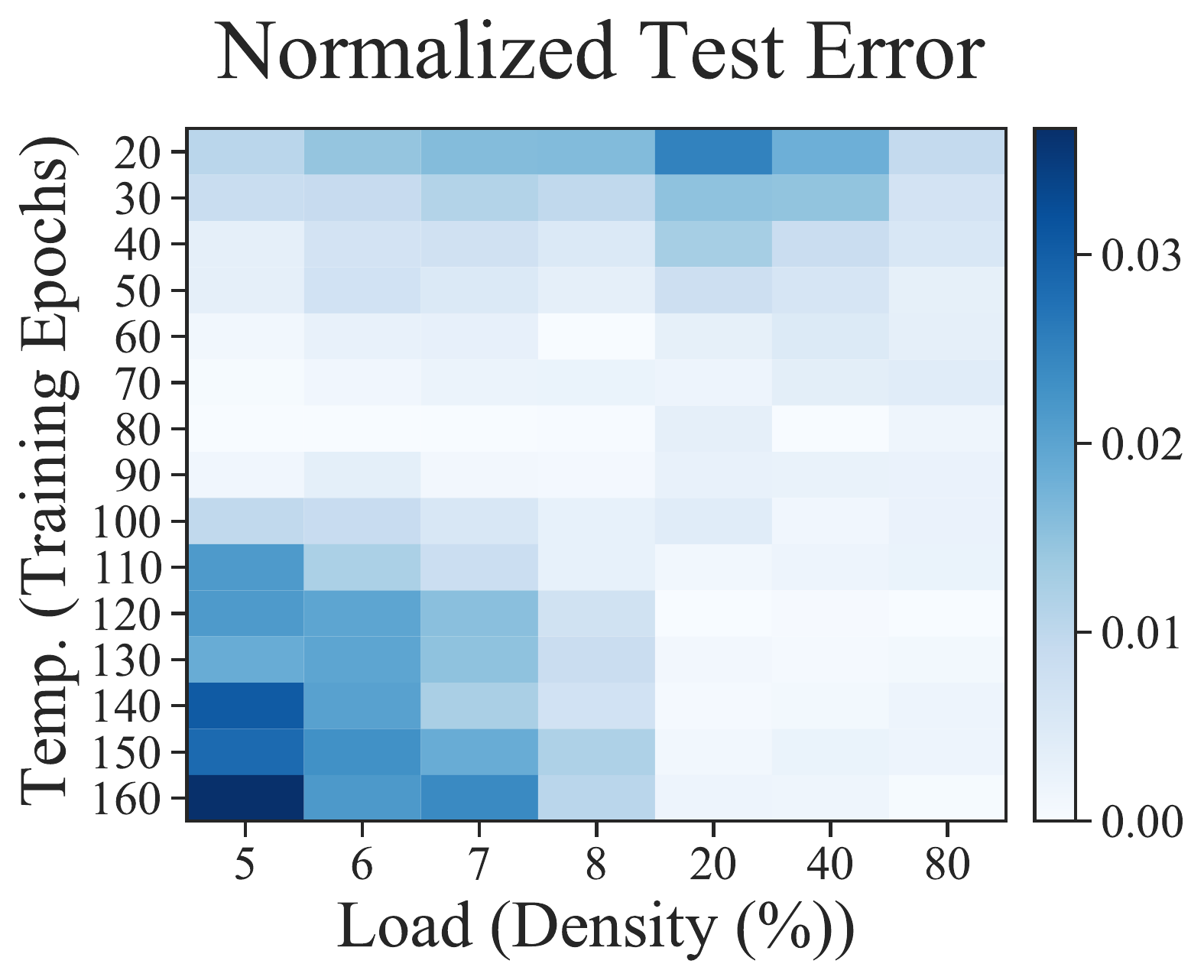}
       \caption{Normalized test error}
       \label{fig:cifar100-norm-error}
   \end{subfigure} 
   \begin{subfigure}{0.24\linewidth}
       \centering
       \includegraphics[width=\linewidth]{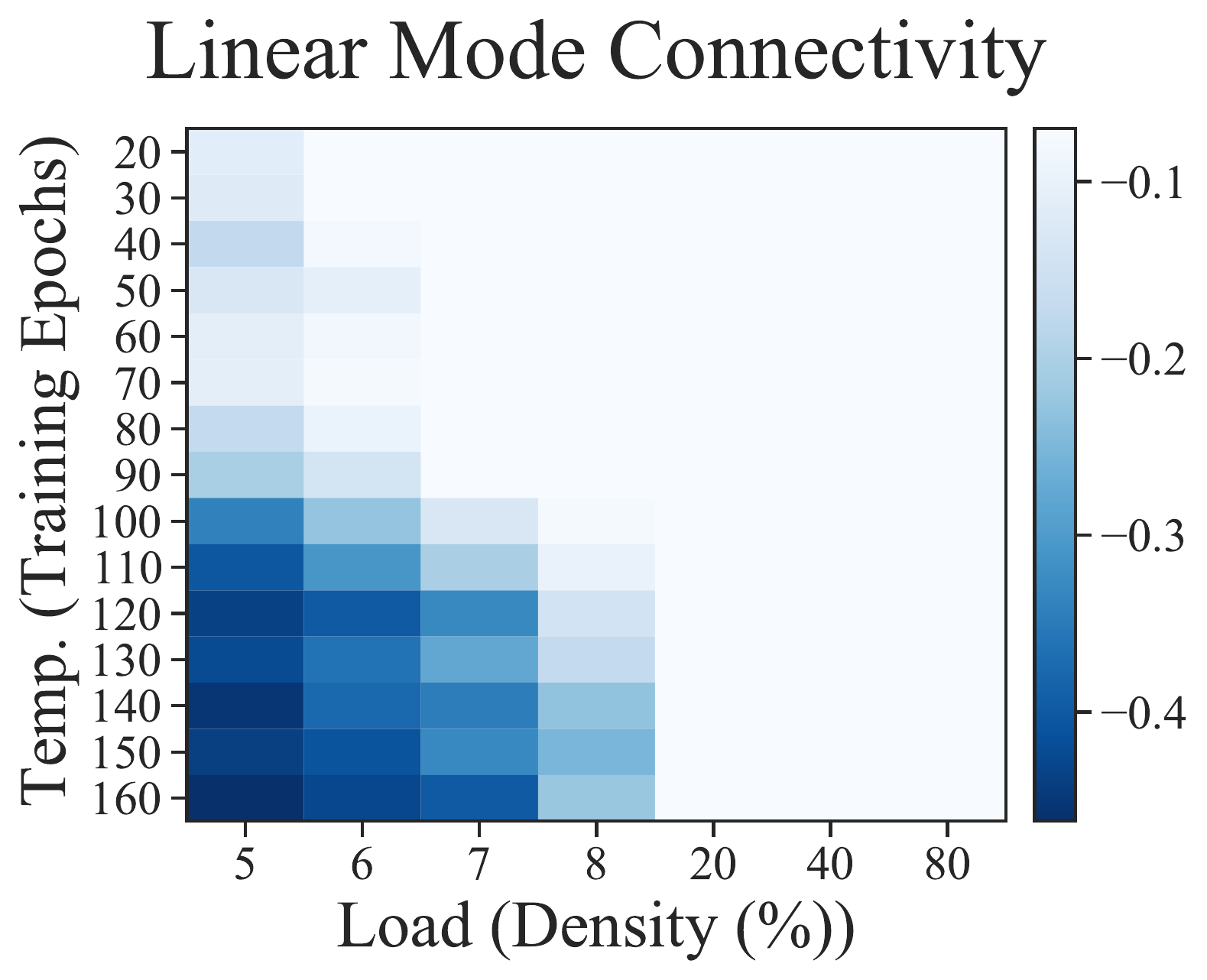}
       \caption{LMC}
       \label{fig:cifar100-lmc}
   \end{subfigure}
   \begin{subfigure}{0.24\linewidth}
       \centering
       \includegraphics[width=\linewidth]{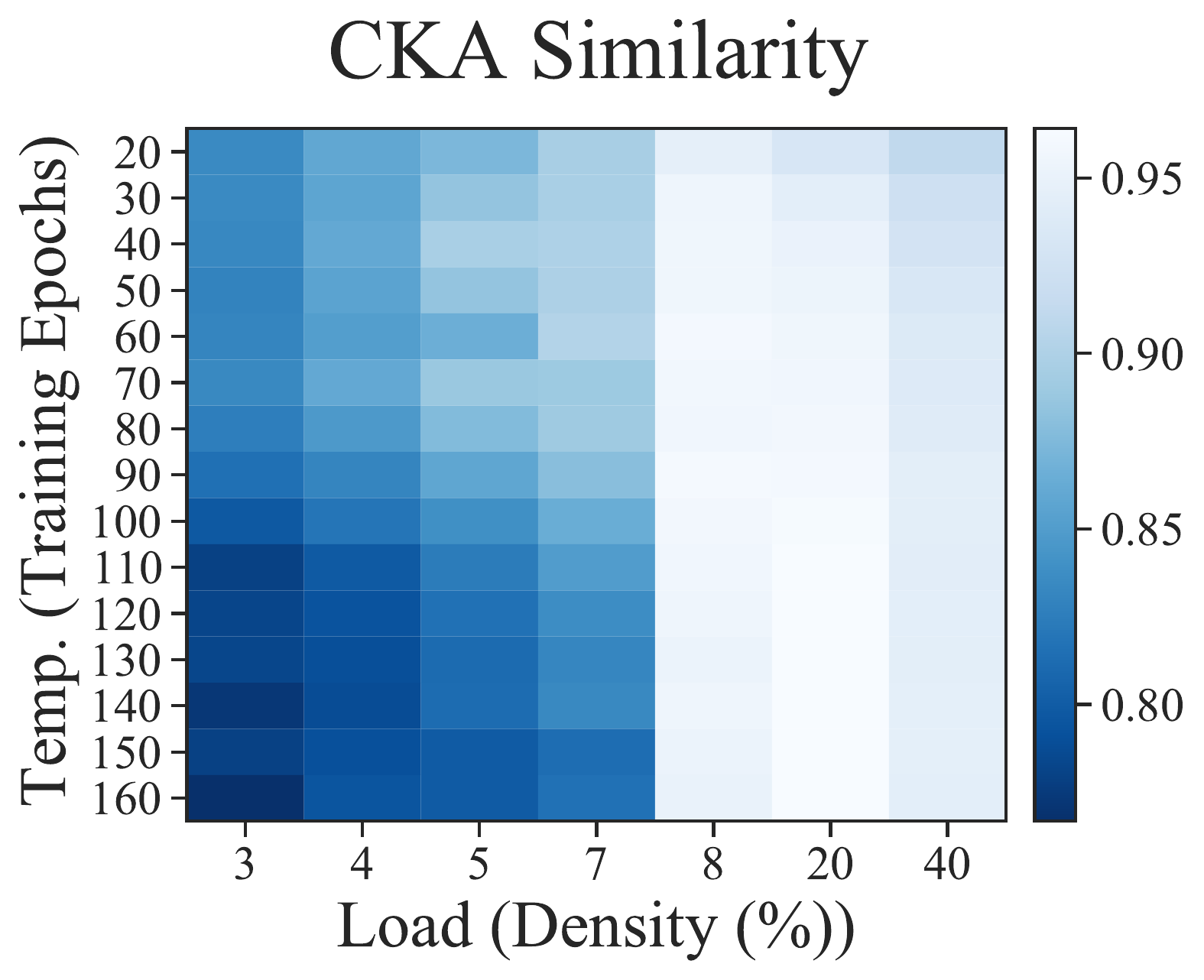}
       \caption{CKA similarity}
       \label{fig:cifar100-cka}
   \end{subfigure}
    \caption{\textbf{(CIFAR-100).} Partitioning the 2D diagram of model density—training epochs into three regimes. Models are trained with PreResNet-20 on CIFAR-100.
    }
    \label{fig:cifar100-phase}
\end{figure*}

\begin{figure*}[!th]\centering
  \begin{subfigure}{0.24\linewidth}
       \centering
       \includegraphics[width=\linewidth]{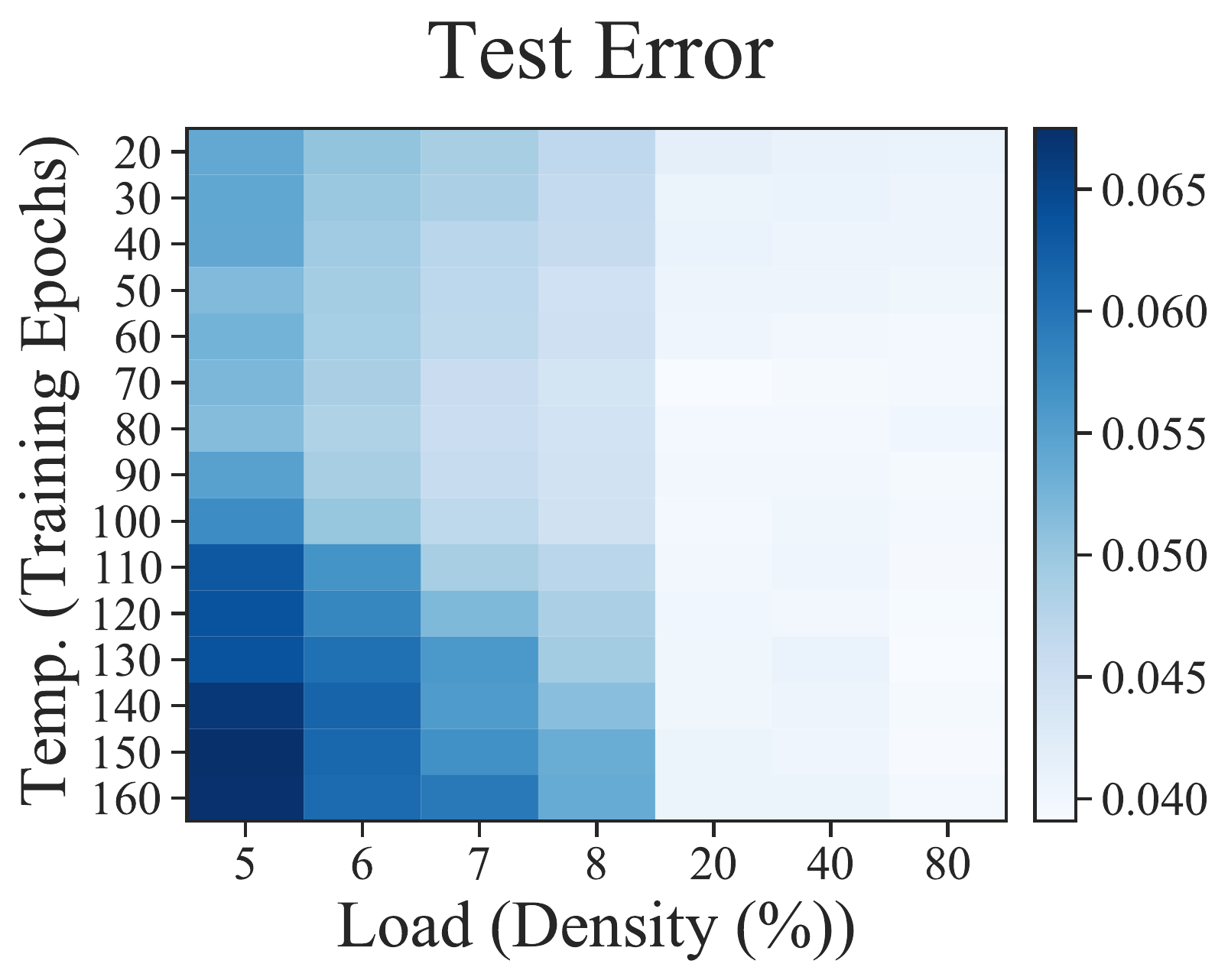}
       \caption{Test error}
       \label{fig:svhn-error}
   \end{subfigure}
    \begin{subfigure}{0.24\linewidth}
       \centering
       \includegraphics[width=\linewidth]{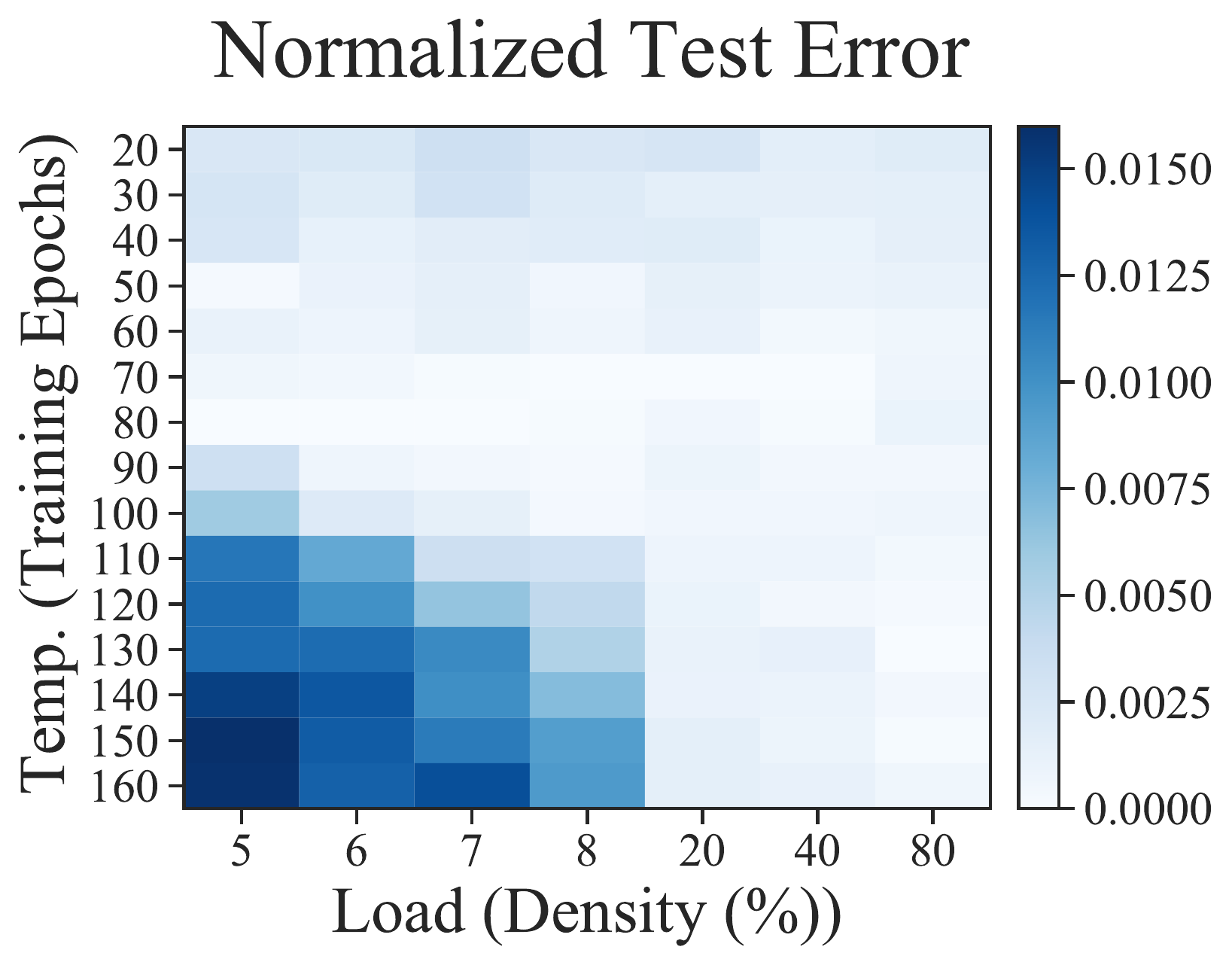}
       \caption{Normalized test error}
       \label{fig:svhn-norm-error}
   \end{subfigure} 
   \begin{subfigure}{0.24\linewidth}
       \centering
       \includegraphics[width=\linewidth]{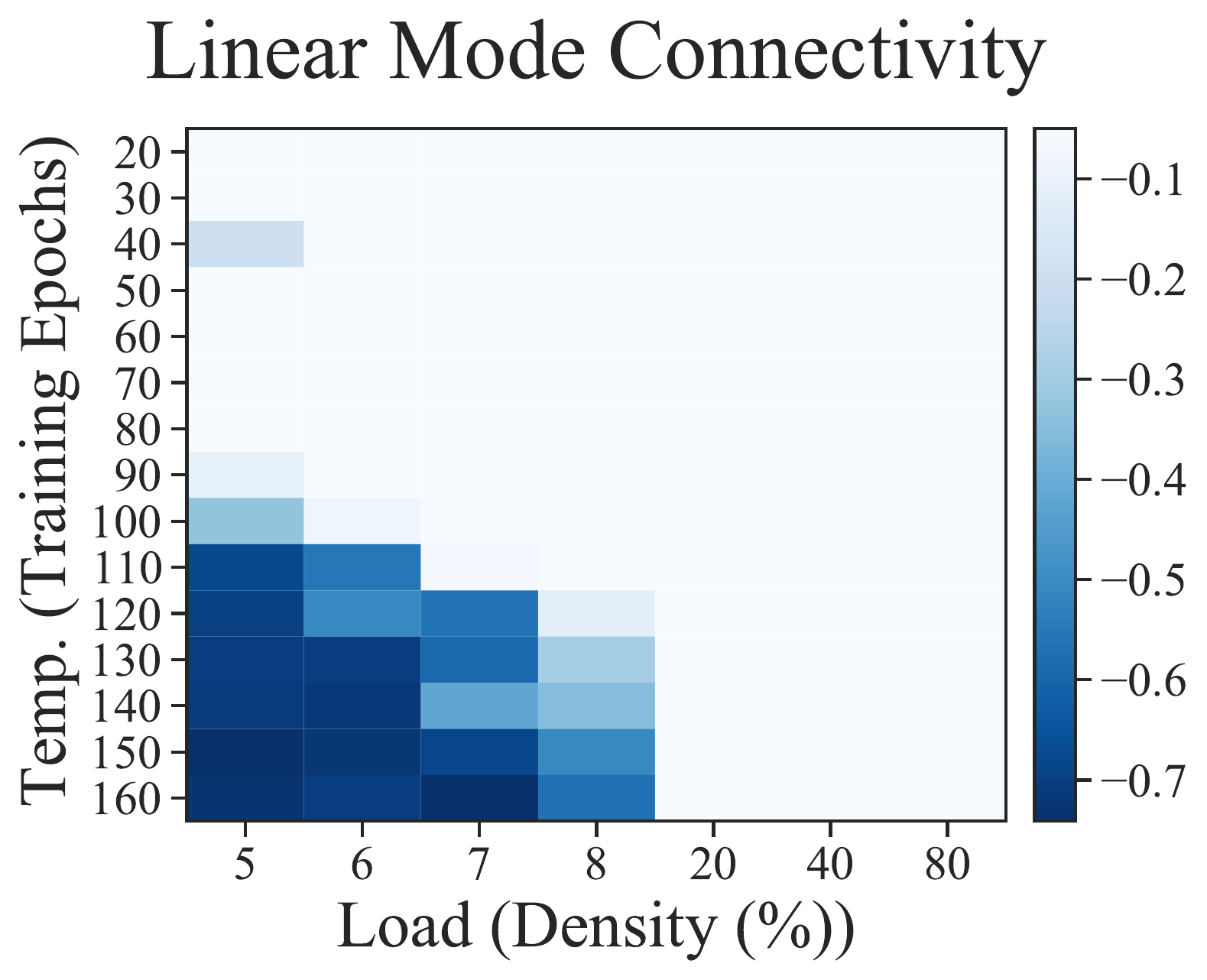}
       \caption{LMC}
       \label{fig:svhn-lmc}
   \end{subfigure}
   \begin{subfigure}{0.24\linewidth}
       \centering
       \includegraphics[width=\linewidth]{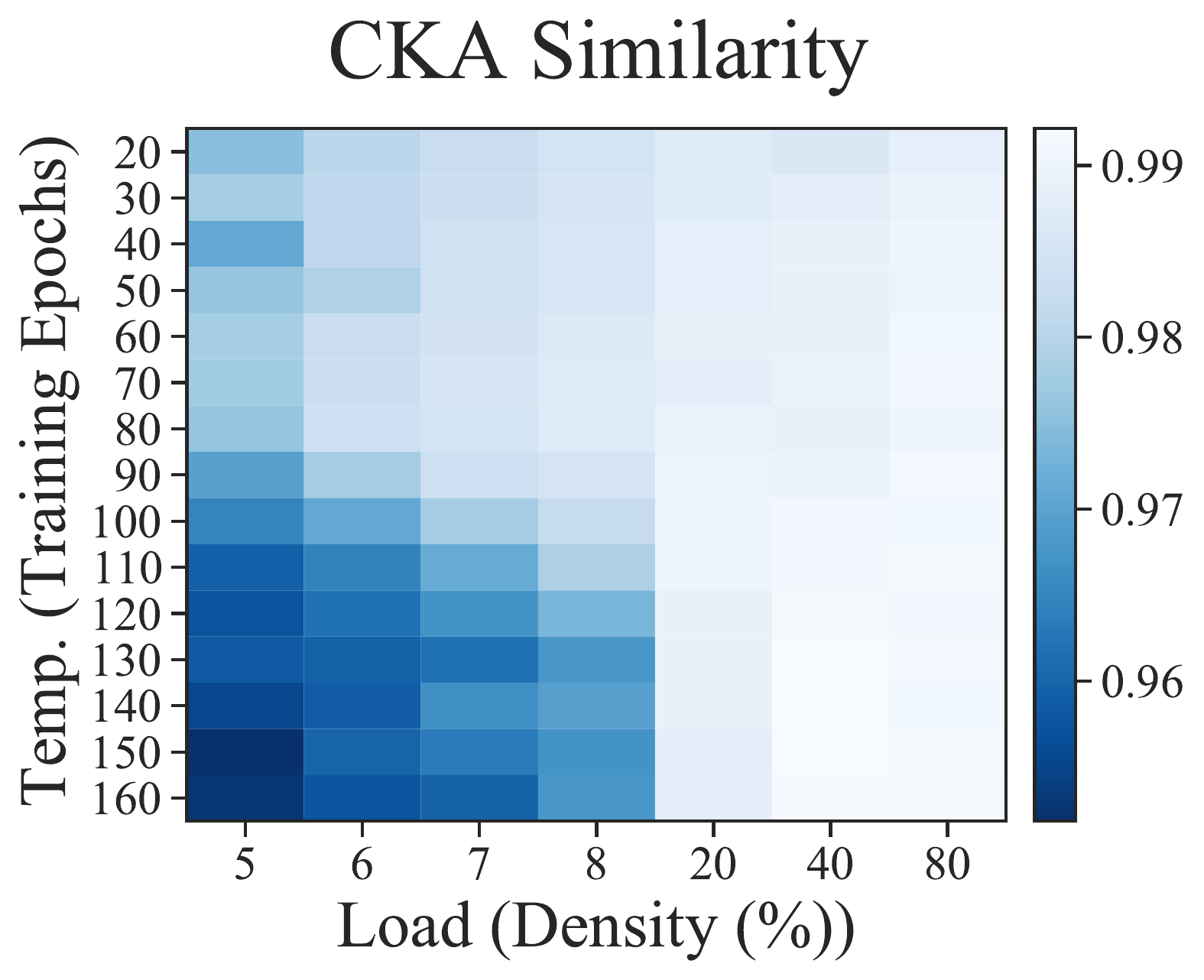}
       \caption{CKA similarity}
       \label{fig:svhn-cka}
   \end{subfigure}
    \caption{\textbf{(SVHN).} Partitioning the 2D diagram of model density—training epochs into three regimes. Models are trained with PreResNet-20 on SVHN.
    }
    \label{fig:svhn-phase}
\end{figure*}

\FloatBarrier
\textbf{Different datasets.}
We extend the experiments with different datasets CIFAR-100 and SVHN in Figure \ref{fig:cifar100-phase} and Figure \ref{fig:svhn-phase}. The experiment produces consistent results with Figure \ref{fig:earlystop-regime} and supports our main claim.

\begin{figure*}[!htb]\centering
  \begin{subfigure}{0.24\linewidth}
       \centering
       \includegraphics[width=\linewidth]{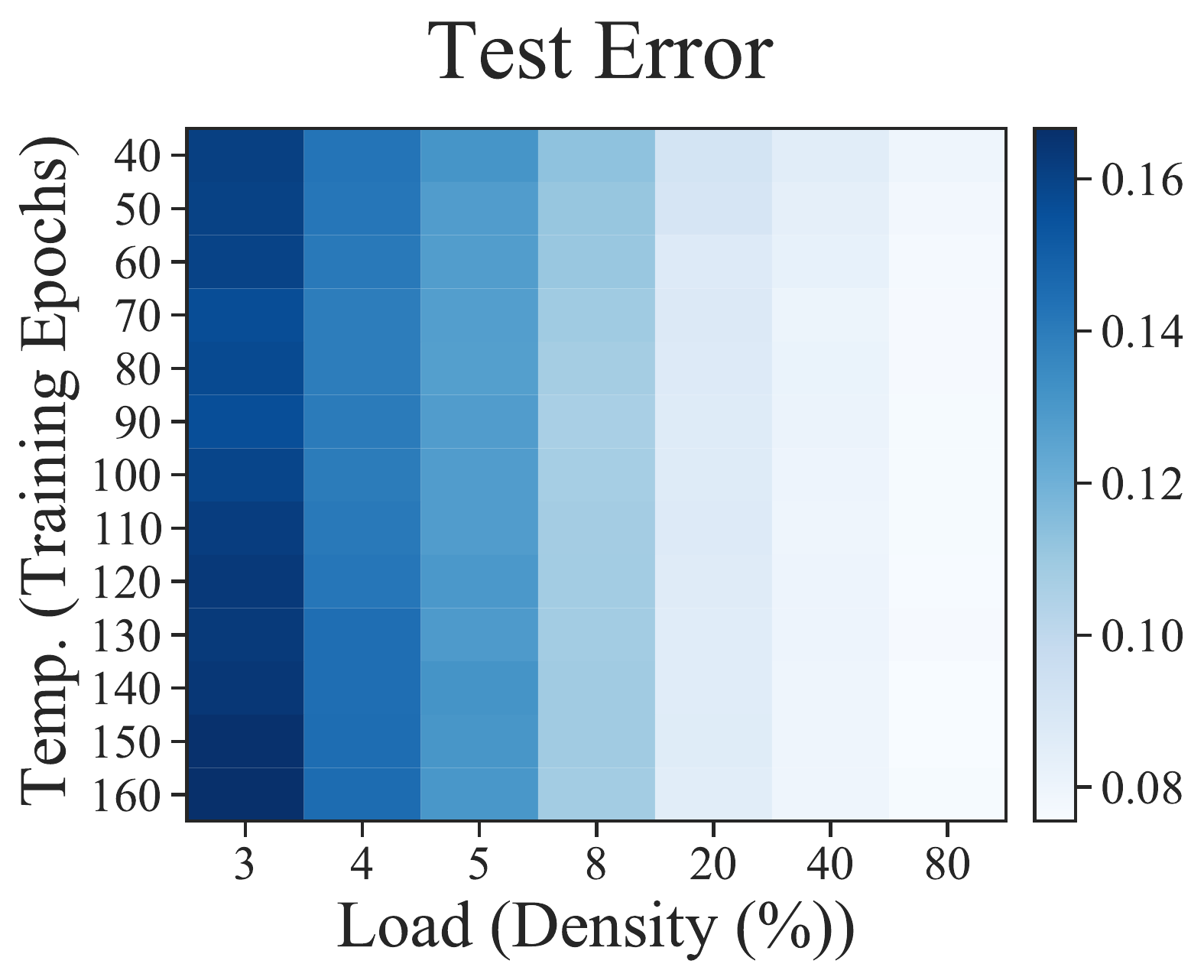}
       \caption{Test error}
   \end{subfigure}
    \begin{subfigure}{0.24\linewidth}
       \centering
       \includegraphics[width=\linewidth]{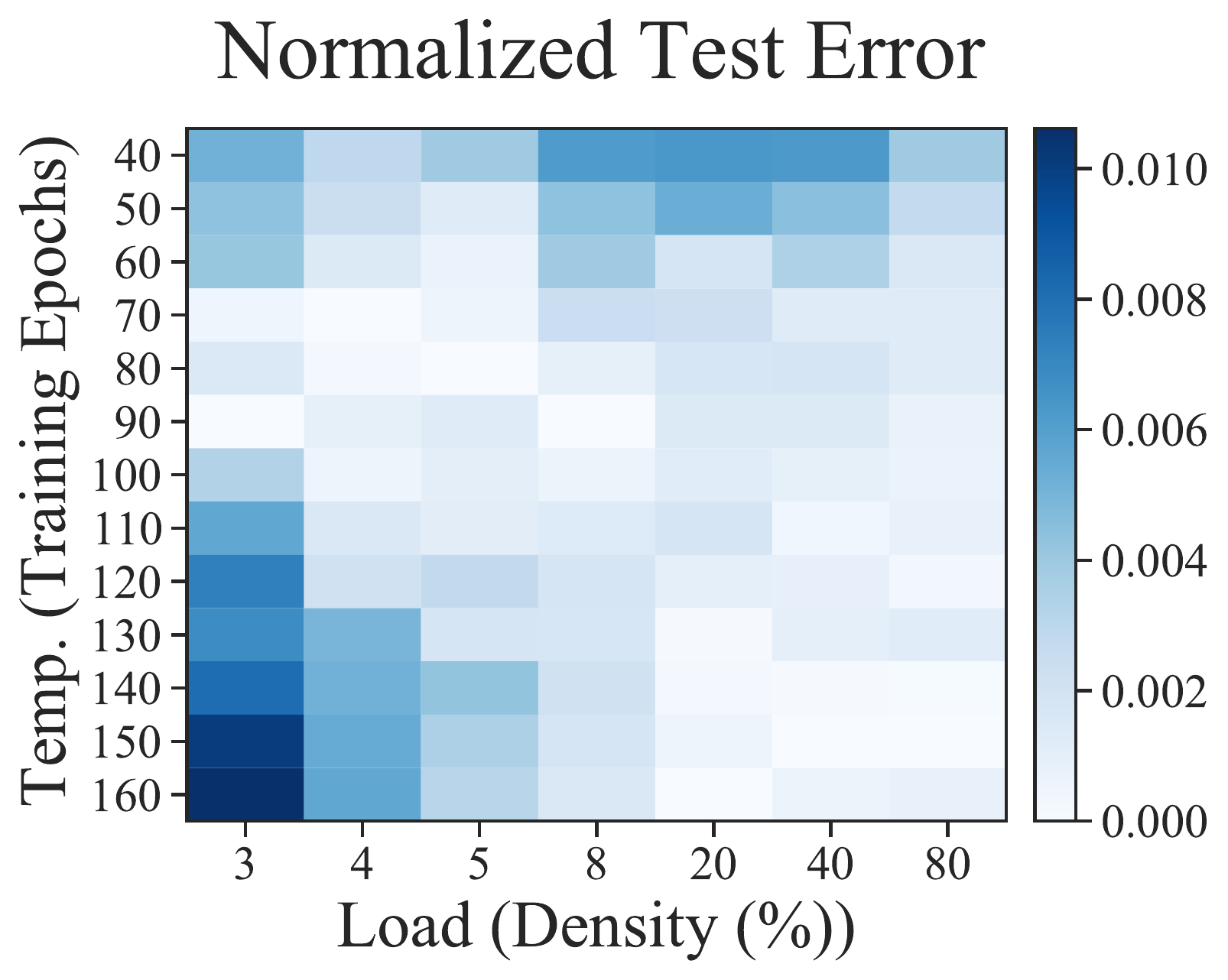}
       \caption{Normalized test error}
   \end{subfigure}
   \begin{subfigure}{0.24\linewidth}
       \centering
       \includegraphics[width=\linewidth]{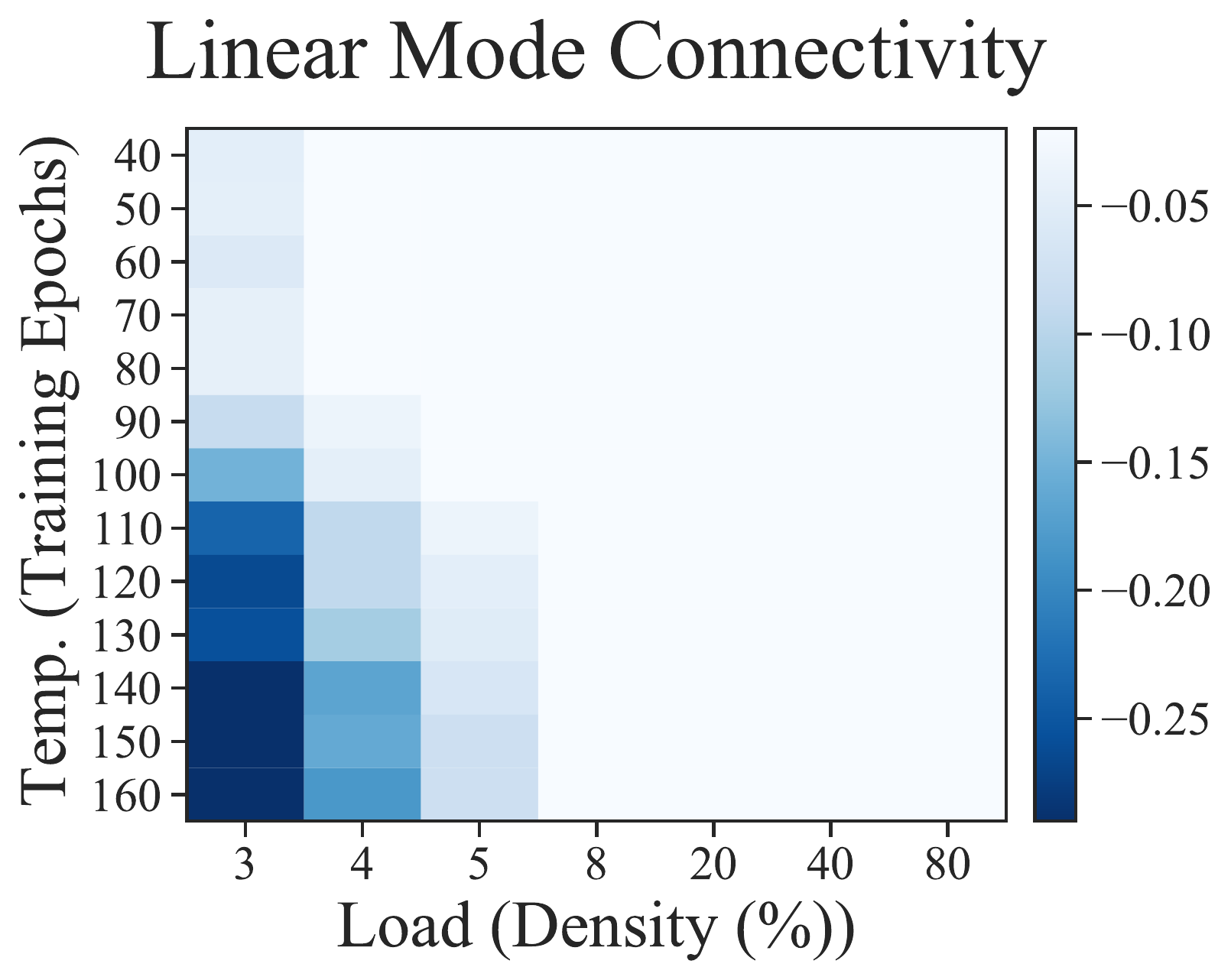}
       \caption{LMC}
   \end{subfigure}
   \begin{subfigure}{0.24\linewidth}
       \centering
       \includegraphics[width=\linewidth]{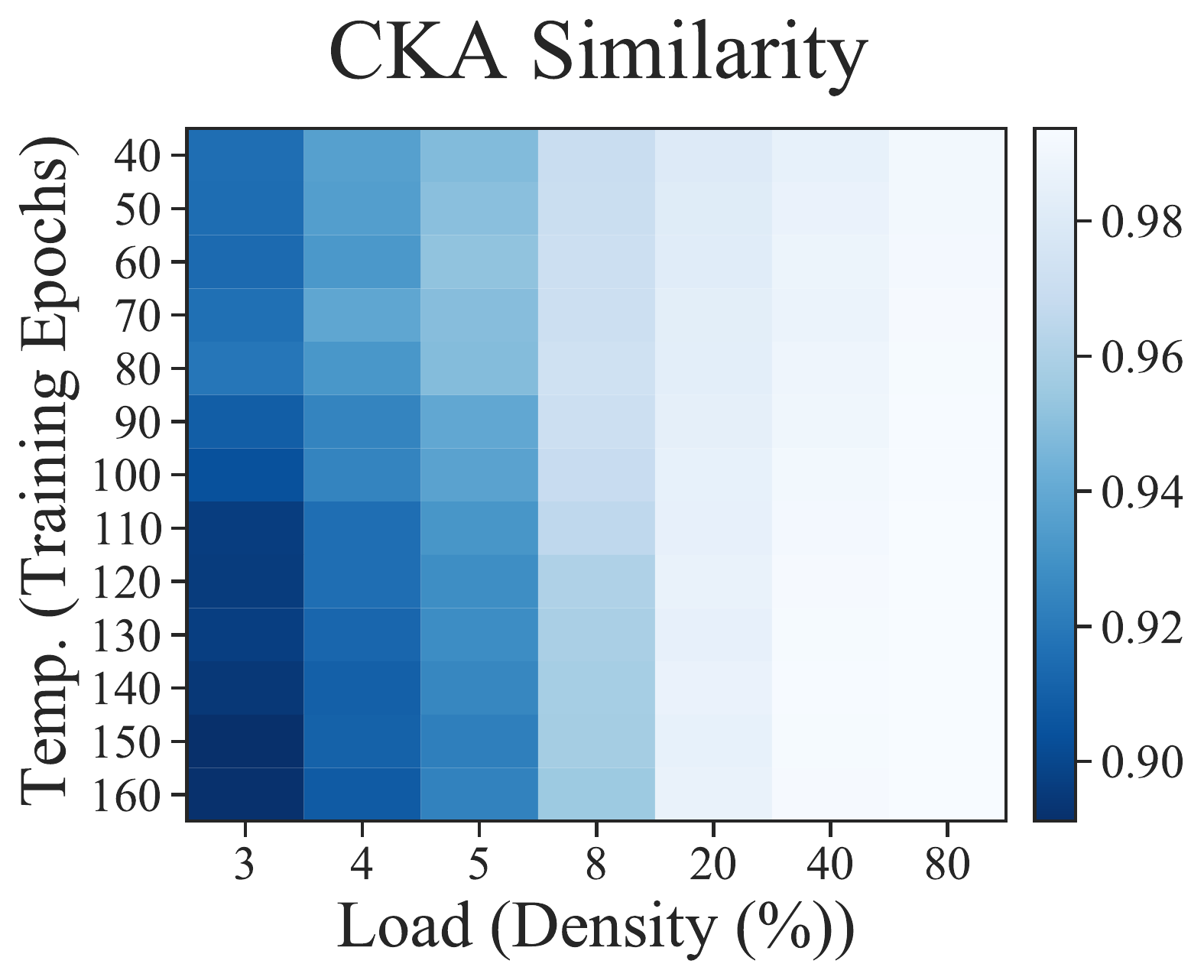}
       \caption{CKA similarity}
   \end{subfigure}
   \caption{\textbf{(Global magnitude pruning).} Partitioning the 2D model density—training epochs diagram into three regimes where we use global magnitude pruning (\textsc{GlobalMP}) instead of uniform magnitude pruning (\textsc{UniformMP}). Models are trained with PreResNet-20 on CIFAR-10. }
   \label{fig:gmp}
\end{figure*}

\textbf{Different pruning strategy.} We confirm the three-regime taxonomy still holds under a different pruning strategy, namely performing \textsc{GlobalMP} in place of \textsc{UniformMP}, as shown in Figure \ref{fig:gmp}. We maintain a similar experimental setup, varying training epochs to adjust the temperature, and we produce the same set of 2D diagrams as Figure \ref{fig:earlystop-regime}.
Comparing Figure \ref{fig:gmp} with Figure \ref{fig:earlystop-regime}, we see that the three regimes are visibly present, and the higher temperature has a dichotomous effect on test error depending on whether LMC is negative or not. Therefore, our central claim is robust to the choice of the pruning method. 

\textbf{Different optimizer.}
We conducted ablation studies on SGD versus Adam with four different settings for network pruning: 1) dense model training with Adam, retrain with SGD, 2) dense model training with Adam, retrain with Adam, 3) dense model training with SGD, retrain with SGD, 4) dense model training with SGD, retrain with Adam. All four settings focus on heavily-pruned models (pruned to the density of 5\%, 6\%), and we expect to see that a large temperature potentially helps improve the test error.

In more detail, here is the experimental setup. We trained PreResNet-20 on CIFAR-10, using the Adam optimizer with 0.001 as the initial learning rate and the SGD optimizer with 0.1 as the initial learning rate. The weight decay is 1e-4 for both. The setup of the learning rate decay schedule, model architecture, dataset, and pruning is the same as the setup in the main paper. We grid-searched the initial learning rate of Adam using values from {0.1, 0.01, 0.001, 0.0001}. We trained the dense model with one random seed and fine-tuned each pruned model with three random seeds.

\begin{figure*}[!htb]\centering
  \begin{subfigure}{0.32\linewidth}
       \centering
       \includegraphics[width=\linewidth]{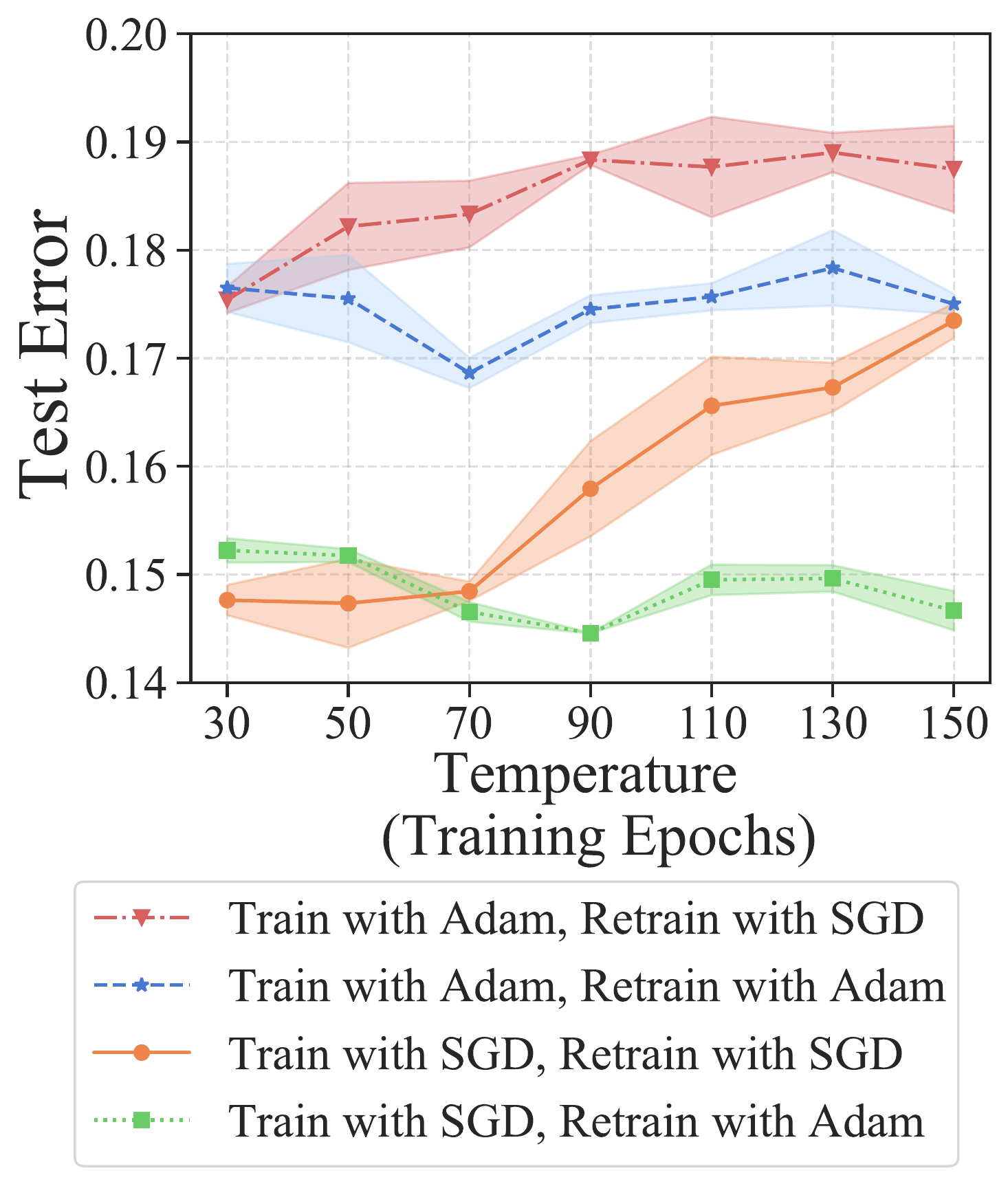}
       \caption{Density = 5\%}
   \end{subfigure}
    \begin{subfigure}{0.32\linewidth}
       \centering
       \includegraphics[width=\linewidth]{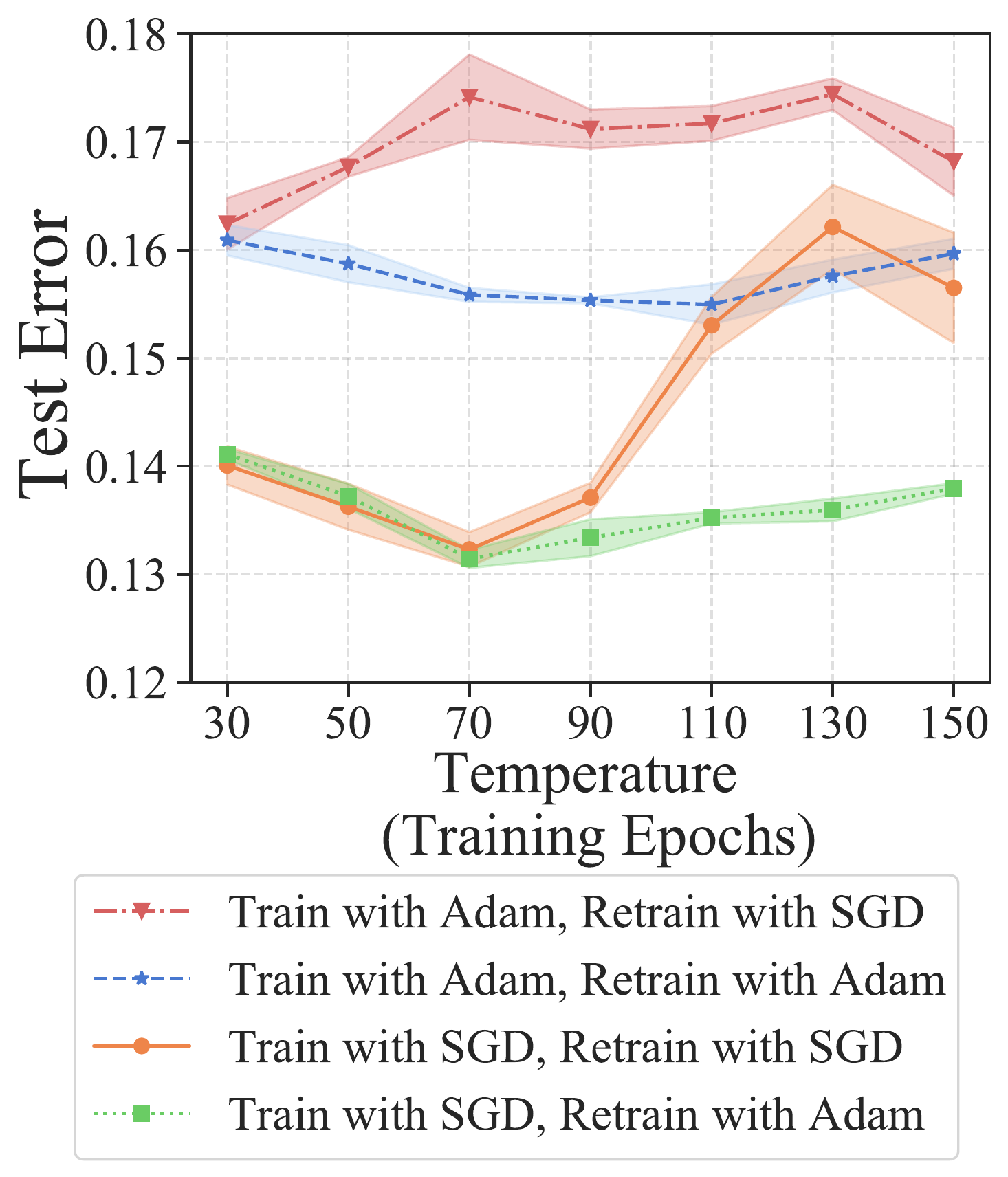}
       \caption{Density = 6\%}
   \end{subfigure}
   \caption{\textbf{(Adam optimizer for pruning.)} Using a large temperature (reducing the training epochs) improves the test error of heavily pruned models consistently when using Adam and/or SGD optimizers during pruning. Models are trained with PreResNet-20 on CIFAR-10.}
   \label{fig:adam}
\end{figure*}

The results are presented in \cref{fig:adam}. First, we noticed that reducing the training epochs (using a large temperature) improves the test error for all four settings, which is consistent with our main claim. Additionally, we noticed that Adam-based dense model training does not work as well as SGD, which is observed in the literature. Interestingly, retraining with Adam significantly improves our result (see the green curves). Our interpretation is that Adam during the retraining works by stabilizing the training with heavily-pruned bottleneck layers.

\textbf{Different task.}
We show that our model extends to machine translation by performing an experiment on the WMT14~\citep{bojar-etal-2014-findings} German to English (DE-EN) dataset using the standard Transformer~\citep{vaswani2017attention} model. We use the Adam optimizer with an inverse square-root learning rate schedule and 4000 warm-up iterations for training and fine-tuning the model. We subsample 1.28M sentence pairs from the WMT14 training set and report the validation BLEU score. The Transformer-base model has 6 layers, 8 attention heads, and an embedding dimension of 512. We varied the temperature over 10 training epochs and the load over 6 densities, ranging from 5\% to 80\%, using \textsc{UniformMP} defined in the paper. We trained the dense model with one random seed and retrain each pruned model with three random seeds.

\begin{figure*}[!htb]\centering
  \begin{subfigure}{0.25\linewidth}
       \centering
       \includegraphics[width=\linewidth]{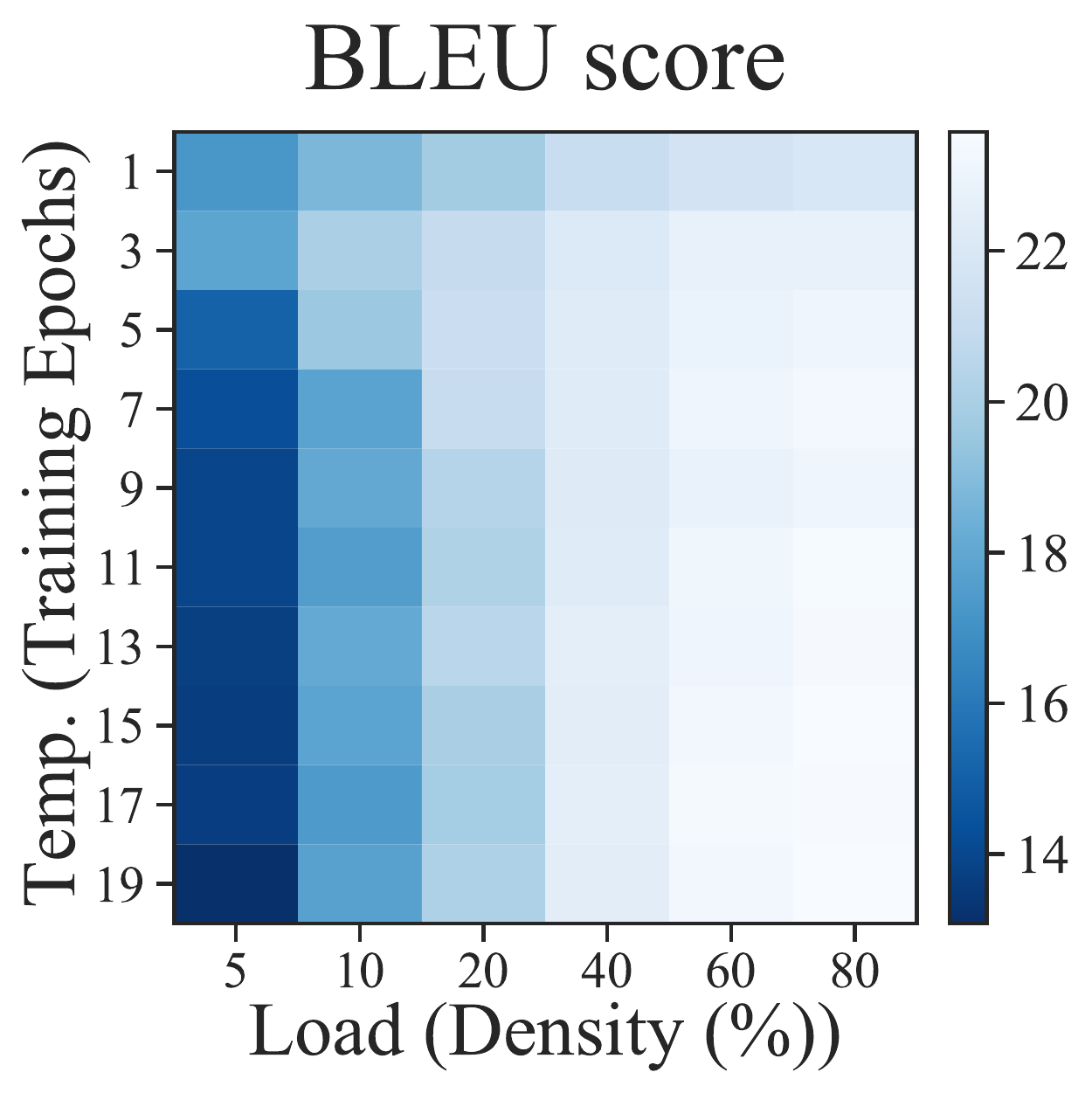}
       \caption{BLEU score}
   \end{subfigure}
    \begin{subfigure}{0.25\linewidth}
       \centering
       \includegraphics[width=\linewidth]{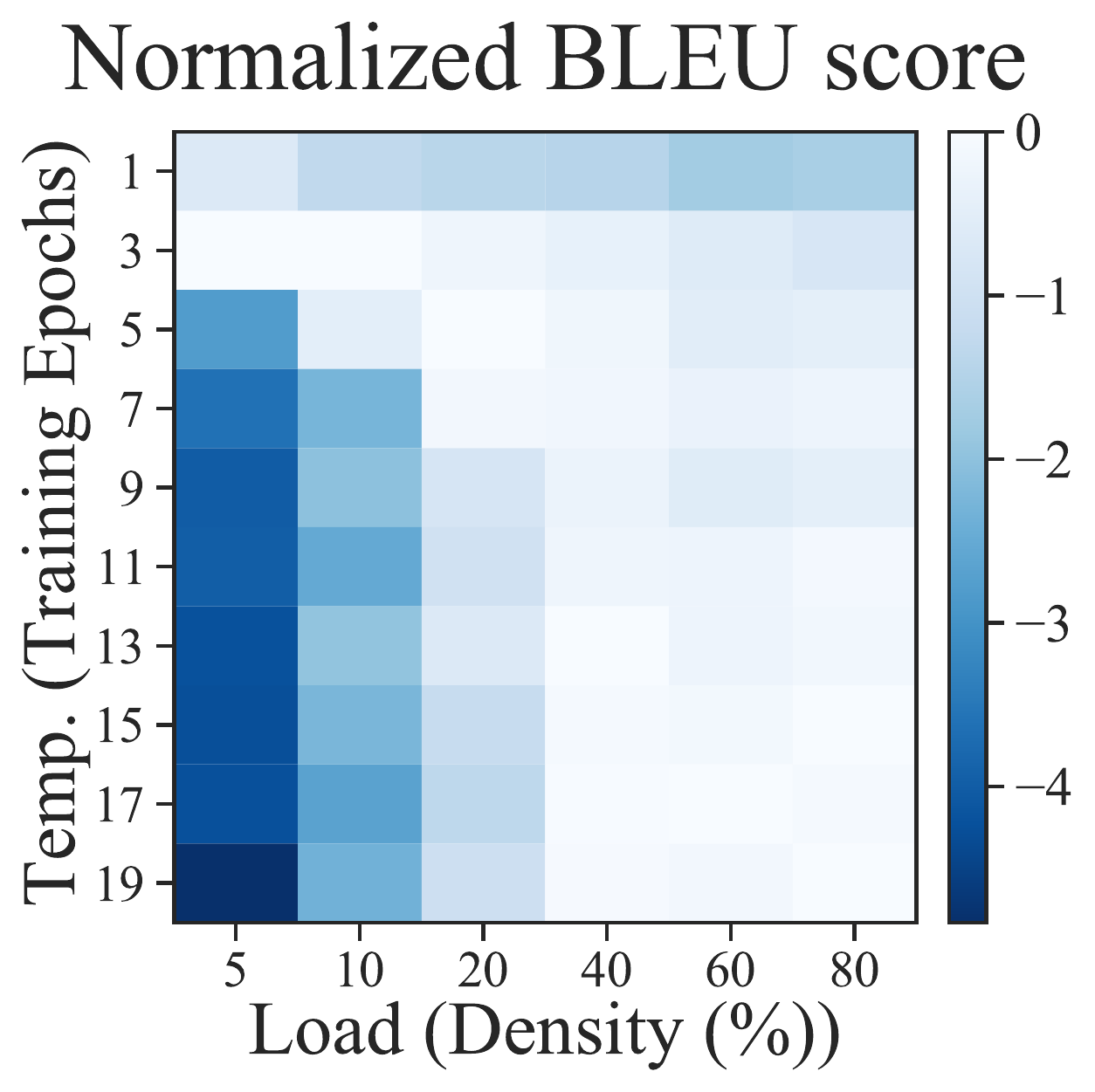}
       \caption{Normalized BLEU score}
   \end{subfigure}
   \caption{\textbf{(Machine transition.)} Larger temperature (fewer training epochs) enhances the BLEU score in low densities, while smaller temperatures help high-density settings. Models are trained with Transformer-base on WMT14.}
   \label{fig:wmt}
\end{figure*}

The results are presented in \cref{fig:wmt}. They demonstrate the same dichotomous phenomenon that the large temperature enhances the BLEU score in low densities, while smaller temperatures help high-density settings, which is consistent with our main findings in image classification tasks.

\section{Additional Details on Three-regime based Applications}

\subsection{Determining the LMC threshold value} \label{sec:thre-explain}
In \cref{sec:app1} and \cref{sec:app2}, we consistently use -0.05 as the LMC threshold value $\epsilon$, which is used to distinguish whether a pruned model belongs to Regime I or Regime II-B (see \cref{fig:three_regime}) in our proposed \cref{alg:tmp-tune} and \cref{alg:model-sele-lmc}. 
This threshold is empirically determined by our three-regime taxonomy results of different datasets and architectures in \cref{sec:phase-plots}. 
Among the three-regime results of multiple architectures and datasets, we consistently find that the transition between Regime I and II-B is very sharp: Regime I, which is the darker region on the lower left, always has a much smaller LMC than Regime II-B, the LMC changing quickly from close to zero to lower than -0.4 when the transition happens. Therefore, -0.05 is a reasonable critical value and is sufficient to distinguish the two regimes. Furthermore, our proposed methods are robust to the choices of the threshold.
Researchers who apply our algorithms in \cref{sec:app1} and \cref{sec:app2} also do not need to generate the three-regime taxonomy results again to determine the threshold value. In other words, they can apply our prior observation in \cref{sec:phase-plots} and employ the threshold of -0.05 as a hyperparameter initialization for their specific tasks.

\begin{table*}[ht]
\centering
\caption{The range of hyperparameters considered in the experiment setup: a dense model is trained with initial temperature $T$ and subsequently pruned to target level of load $L$.}
\label{tab:vary-depth}
\begin{tabular}{@{}c|c|c@{}}
\toprule
\multicolumn{2}{c}{Initial temperature to adjust} & Target level of load \\ \cmidrule(r){1-2} \cmidrule(r){3-3}
Training Epochs (\small{$T_{\text{epoch}}$})  & Batch Size (\small{$T_{\text{batch}}$})   & Model Depth (\small{$L_{\text{depth}}$})  \\ \midrule
\small{160}   &  \small{128}  & \small{$L_{\text{depth}}$} $\subseteq$ \{20, 32, 44, 56, 80, 152\} 
\\ \bottomrule
\end{tabular}
\end{table*}

\subsection{Additional results on hyperparameter-tuning scheme} \label{sec:hyper-tune}

This subsection provides supplementary results to Section~\ref{sec:app1}, demonstrating how to use the proposed temperature-tuning scheme to predict if we should increase or decrease temperature given a specific pruning configuration with load and temperature.
We study model depth $T_\text{depth}$ as a load-like parameter.

\begin{figure*}[thb!]
\begin{centering}
	\begin{tabular}{cc}
	\toprule
    \multicolumn{2}{c}{Varying $L_\text{depth}$}  \\
    \toprule 
    $T_\text{epoch} \times L_\text{depth} $  &  $T_\text{batch} \times L_\text{depth}$  \\

     \includegraphics[width=0.24\linewidth,keepaspectratio]{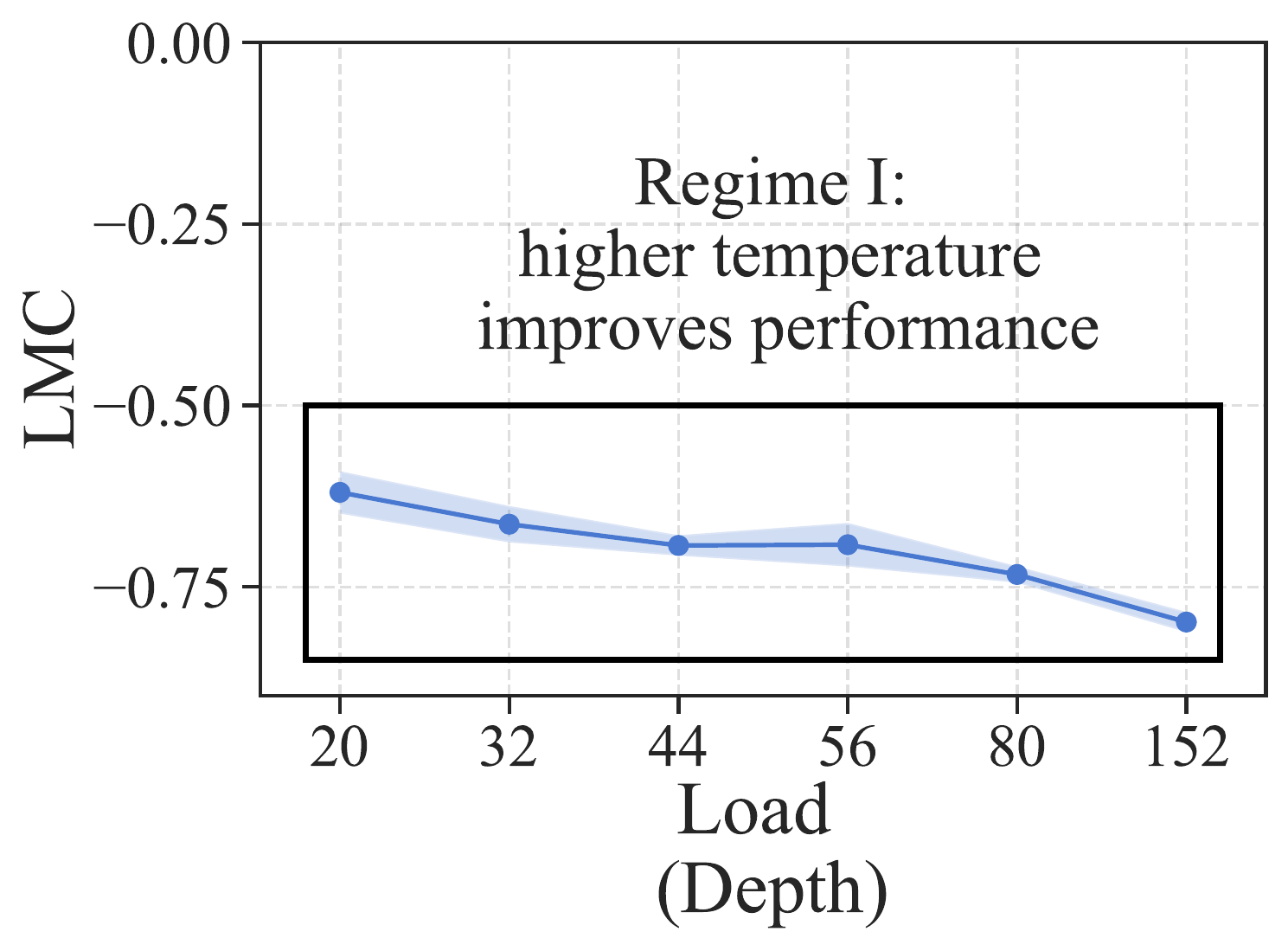} &\includegraphics[width=0.24\linewidth,keepaspectratio]{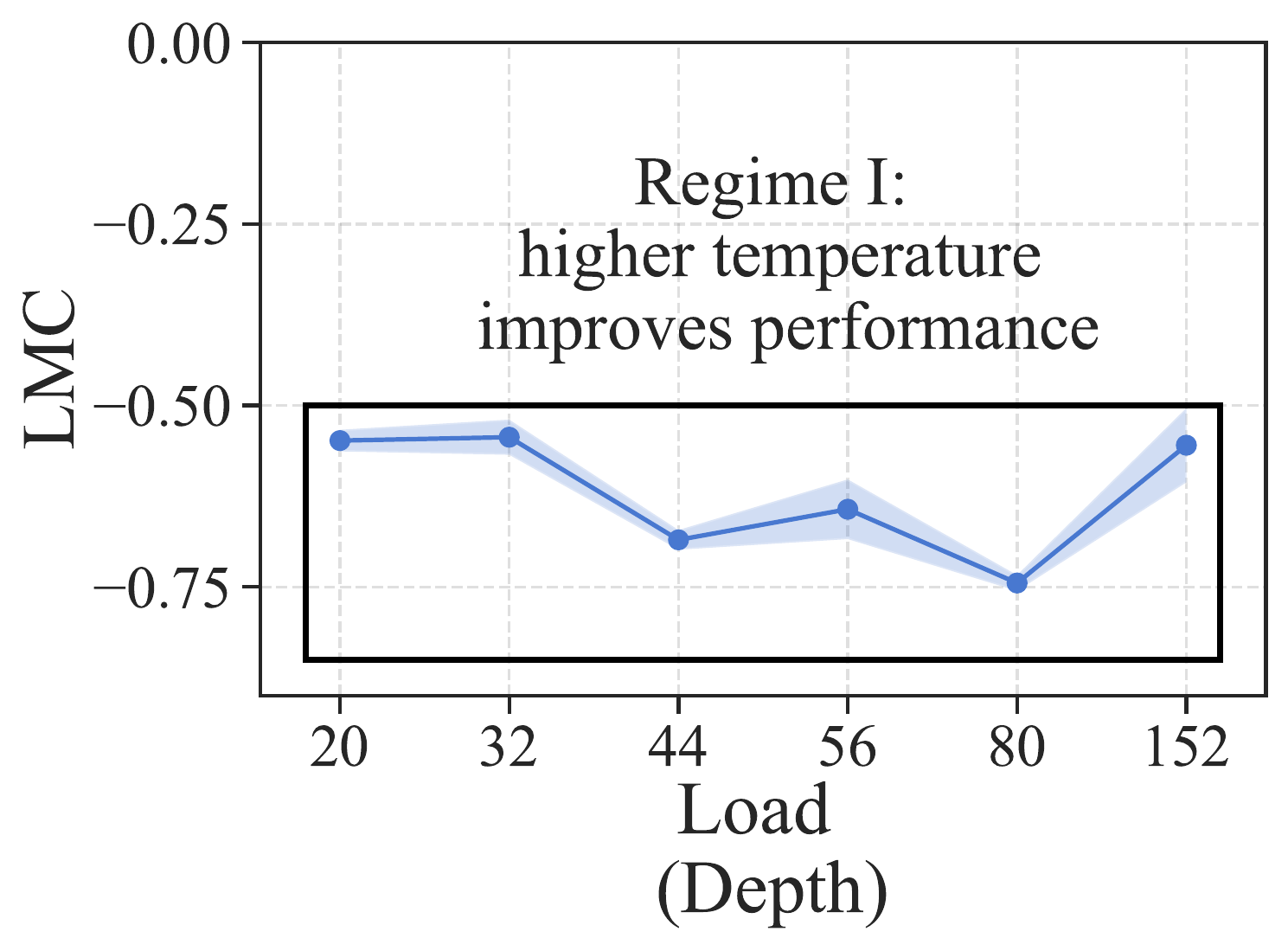} \\
 
  \includegraphics[width=0.24\linewidth,keepaspectratio]{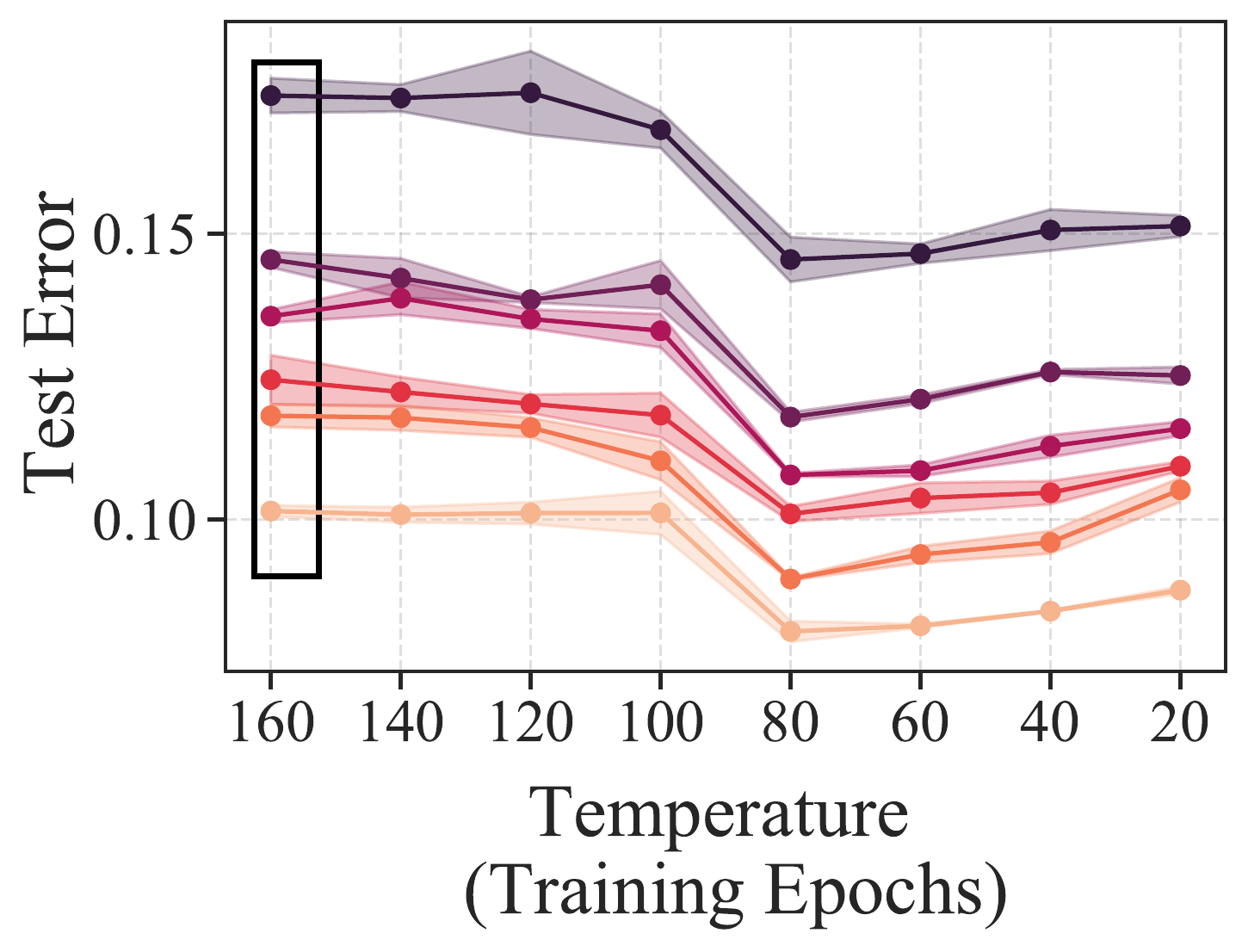}
  &\includegraphics[width=0.24\linewidth,keepaspectratio]{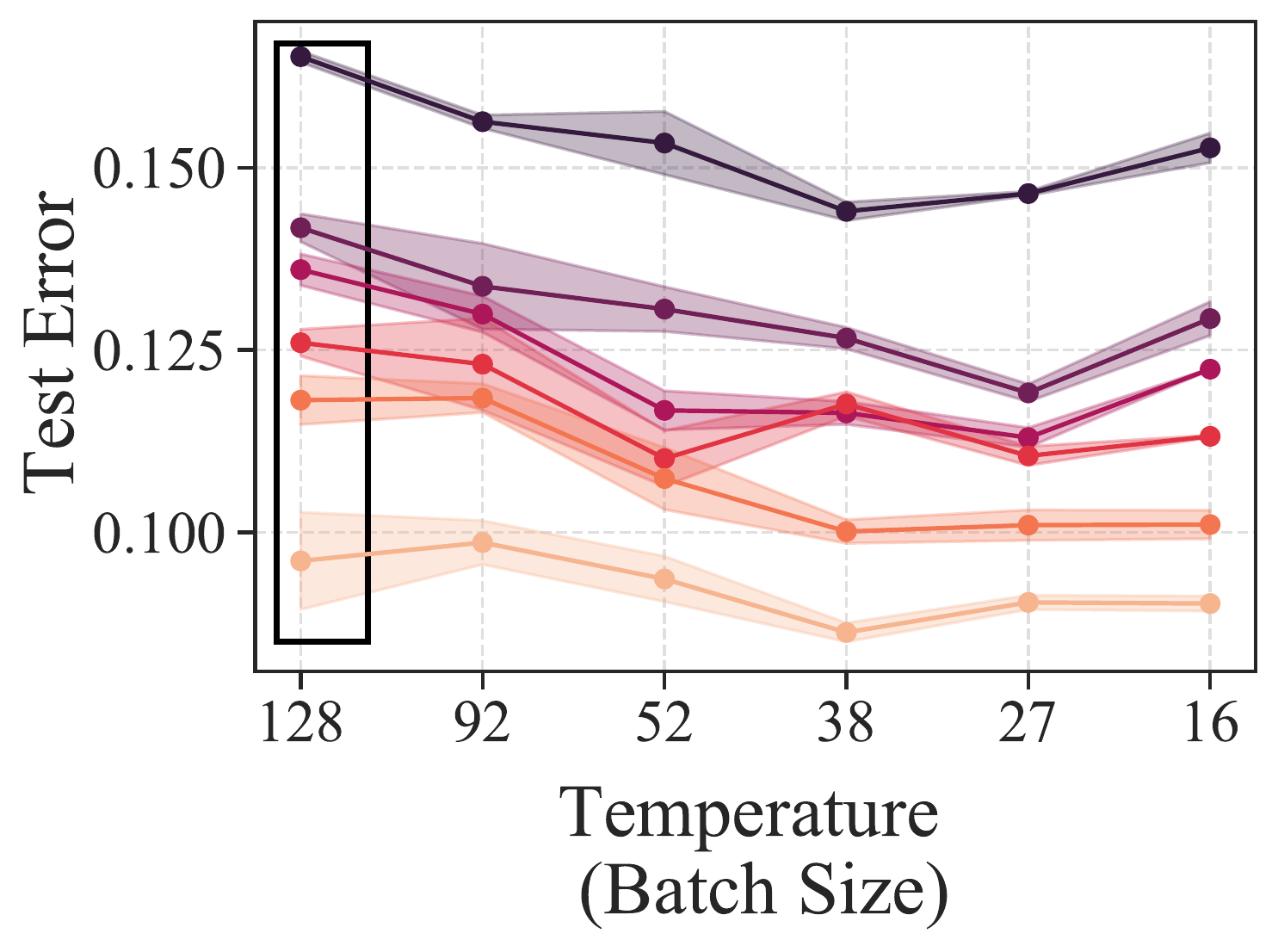} \\

    \multicolumn{2}{c}{\includegraphics[width=0.50\linewidth,keepaspectratio]{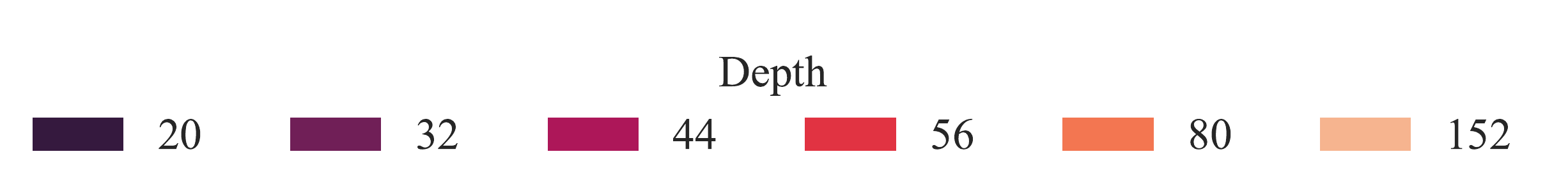}}
	\end{tabular} 
	\caption{Using LMC to determine whether to increase or decrease temperature: models with negative LMC are located in Regime I (annotated by the black box), and their test error can be reduced by increasing temperature.} 
	\label{fig:depth-vary}
\end{centering}
\end{figure*}

{\bf Experiment details}.
We study PreResNet-20 with width scaling 1 on CIFAR-10 and prune the model to 5$\%$ of model density with \textsc{UniformMP}. We consider studying the varying temperature $T$ and load $L$, as detailed in Table \ref{tab:vary-depth}.

{\bf Results}. Results are presented in Figure \ref{fig:depth-vary}. From the LMC results on the first row, our scheme determines that all the depths should be tuned with higher temperatures since they have LMC $< \epsilon$ ($\epsilon$ = -0.05, annotated by the black box). This prediction is verified by the test error results on the second row, which shows that the test error can be reduced by increasing the temperature. In contrast with the other two load-like parameters density and width scaling, one special finding here is that increasing model depth does not help the configuration transit to Regime II (and thus all the cases studied can be benefited from higher temperatures).

\FloatBarrier

\renewcommand{\thealgorithmfloat}{2.1}
\begin{algorithmfloat}[!thb]
   \caption{Model Selection via LMC and CKA}
\begin{algorithmic}
   \STATE {\bfseries Input:} a set of trained dense models $\{\Theta_i\}_{i=1}^{n}$, and corresponding test errors $\{\operatorname{error}_\text{test}(\Theta_i)\}_{i=1}^{n}$,
   LMC threshold $\epsilon$, training procedure $\operatorname{Train}$ (Section \ref{def:preli}), pruning procedure $\operatorname{Prune}$ (Section \ref{def:network-prune}),
   target load $L$, retraining epochs $\alpha$
   \STATE {\bfseries Output:}  dense model $\Theta_{i^{*}}$ to prune
   \vskip 0.5em
   \STATE $i^{*} = \mathop{\arg\min}_i \{ \operatorname{error}_\text{test}(\Theta_i)\}_{i=1}^{n}$
   
   \STATE $\Theta_{i^{*}} \odot \mathcal{M} = \operatorname{Prune}(\Theta_{i^{*}}, L)$
   \STATE Compute LMC on $\Theta_{i^{*}} \odot \mathcal{M}$ via \cref{eq:mode_conn}
   \IF{$\text{LMC} < \epsilon$}
    \STATE $\Theta_{i} \odot \mathcal{M}_{i}  = \operatorname{Train}(\operatorname{Prune}(\Theta_{i}, L) , \alpha)~\text{for}~i \in [1, n]$
    \STATE $ i^{*} = \mathop{\arg\max}_i \{ \text{CKA}(\Theta_{i} \odot \mathcal{M}_{i}) \}_{i=1}^n$ via \cref{eq:cka}
   \ENDIF
\end{algorithmic}
\label{alg:model-sele-lmc-cka}
\end{algorithmfloat}

\subsection{Additional details for model selections} 
\label{sec:sup-model-select-method}
In this subsection, we provide diagrams to describe the algorithms used in model selection tasks. 
Algorithm \ref{alg:model-sele-lmc} shows the proposed method of using LMC and test error for selection with access to test data.
Algorithm \ref{alg:model-sele-lmc-cka} shows the proposed method using LMC and CKA for selection without access to test data.

\begin{figure}[!htb]\centering
  \begin{subfigure}{0.33\columnwidth}
       \centering
       \includegraphics[width=\linewidth]{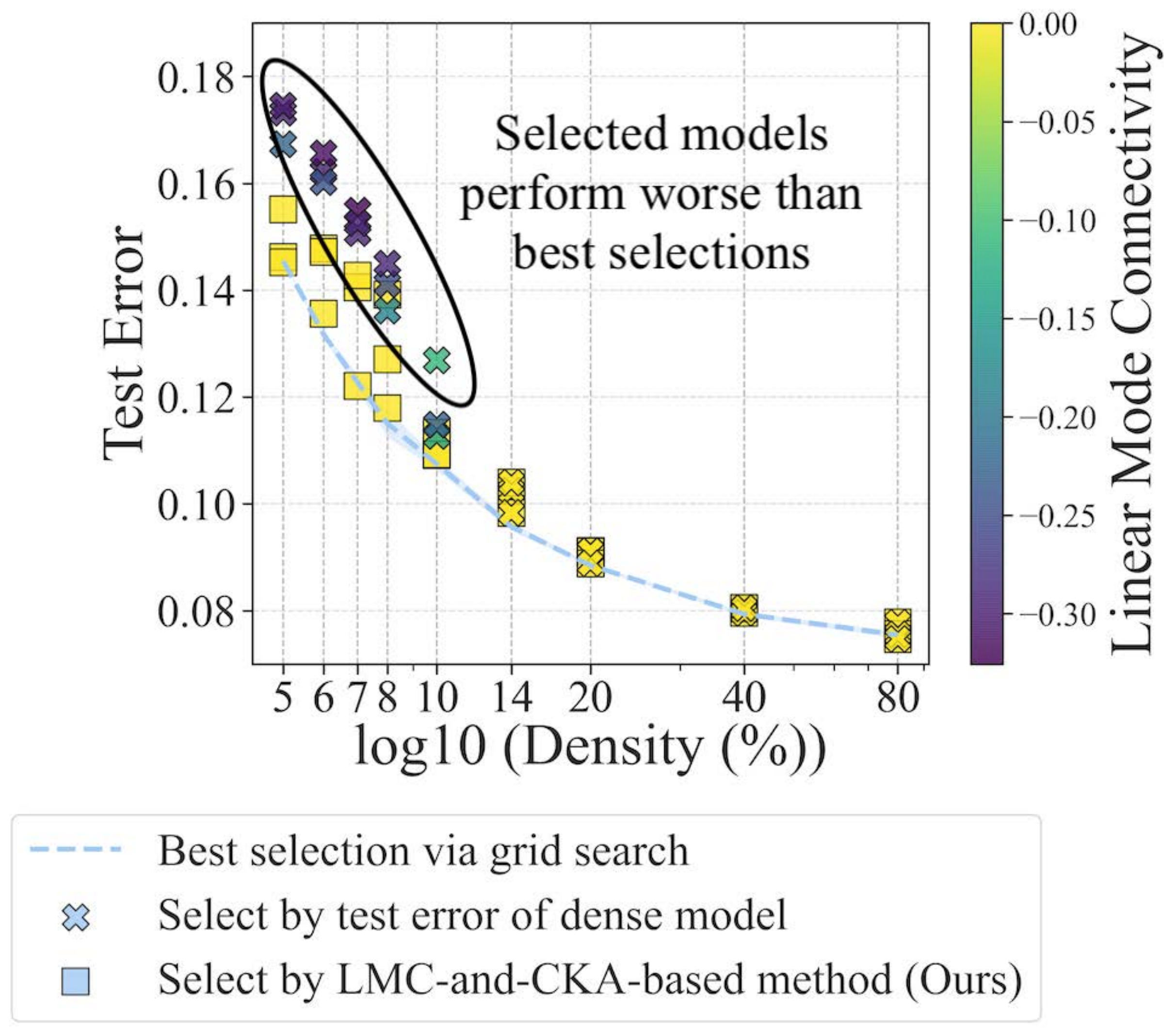}
       \caption{Selecting the best model from those trained with various training epochs.}
   \end{subfigure}
    \begin{subfigure}{0.32\columnwidth}
       \centering
       \includegraphics[width=\linewidth]{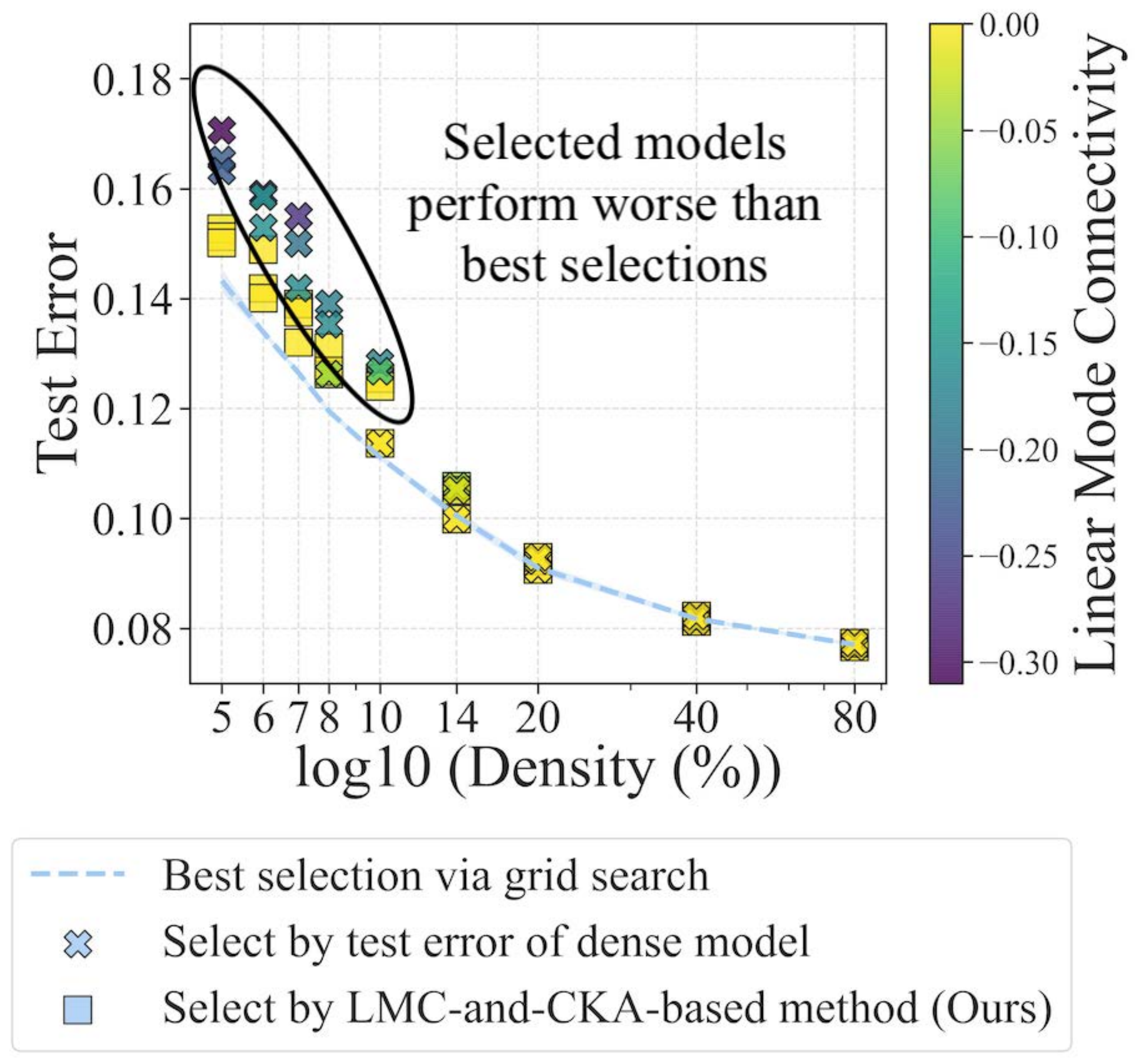}
       \caption{Selecting the best model from those trained with various batch sizes.}
   \end{subfigure} 
    \caption{Selecting temperature using LMC and CKA similarity (squares) leads to a smaller test error than selecting temperature using the test error of the unpruned dense model (crosses). The retraining epochs of the proposed method $\alpha = 2$. 
    The performance of CKA-based selection is close to the best test error found by grid search (dashed lines).
    (Left) Selecting the best training epoch. (Right) Selecting the best batch size. 
    Models that perform significantly worse than grid search tend to have worse LMC, shown by the dark color of markers.
    }
    \label{fig:prediction-via-lmc-cka}
\end{figure}

We provide the results of evaluating Algorithm \ref{alg:model-sele-lmc-cka}. See Figure \ref{fig:prediction-via-lmc-cka}. 
In the high-density regime, this method shows comparable performance as the grid search but uses fewer retraining epochs: this method takes 2 retraining epochs to determine that the baseline selection performs well for this density, while the grid search takes 160 retraining epochs for each candidate in the set.
In the low-density regime, this method performs better than the baseline selections by using the LMC to avoid selecting the bad model to prune.

\FloatBarrier

\subsection{SAM hyperparameter $\rho$ range}
SAM was originally a method proposed to improve generalization in \citet{foret2021sharpnessaware}; here we use it to train dense models to improve network pruning which is a different task. Thus, we would not expect our optimal $\rho$ to be the same as the range in \citet{foret2021sharpnessaware}. Furthermore, in our experiments, the range of $\rho$ that we sweep over indeed includes {0.01, 0.02, 0.05, 0.1, 0.2, 0.5}, and we find that using the $\rho$ slightly larger than the original range provides more improvement on pruning.

%You can have as much text here as you want. The main body must be at most $8$ pages long.
%For the final version, one more page can be added.
%If you want, you can use an appendix like this one, even using the one-column format.
%%%%%%%%%%%%%%%%%%%%%%%%%%%%%%%%%%%%%%%%%%%%%%%%%%%%%%%%%%%%%%%%%%%%%%%%%%%%%%%
%%%%%%%%%%%%%%%%%%%%%%%%%%%%%%%%%%%%%%%%%%%%%%%%%%%%%%%%%%%%%%%%%%%%%%%%%%%%%%%

\end{document}